\documentclass[a4paper,11pt]{article}
\usepackage[utf8]{inputenc}
\usepackage{authblk}
\usepackage{algorithmic}
\usepackage{amstext}
\usepackage{graphicx}
\usepackage{amsmath}
\usepackage{amssymb}
\usepackage{amsthm}
\usepackage[caption=false]{subfig}
\usepackage{tikz}
\usepackage{multirow}
\usepackage{ulem}
\usetikzlibrary{fit}
\usetikzlibrary{shapes.geometric}
\usetikzlibrary{arrows.meta,arrows}

\usepackage{pifont}

\usepackage{pdflscape}

\usepackage{tabularx}
\usepackage{arydshln}
\usepackage{bm}
\usepackage[hidelinks]{hyperref}
\usepackage{threeparttable}
\usepackage{comment}
\usepackage{float}

\usepackage{natbib}
\setcitestyle{authoryear,open={(},close={)}}

\usepackage[margin=1in,footskip=0.25in]{geometry}
\usepackage{array}
\usepackage{arydshln}
\usepackage{xcolor}

\newcolumntype{P}[1]{>{\centering\arraybackslash}p{#1}}
\newcolumntype{M}[1]{>{\centering\arraybackslash}m{#1}}

\usepackage{totcount}
\newtotcounter{citnum}
\def\oldbibitem{} \let\oldbibitem=\bibitem
\def\bibitem{\stepcounter{citnum}\oldbibitem}

\usepackage{booktabs}

\title{A comprehensive review of automatic text summarization techniques: method, data, evaluation and coding}

\author[1,6,7]{Daniel O. Cajueiro}
\author[1,7]{Arthur G. Nery}
\author[2]{Igor Tavares}
\author[3,7]{Ma\'{i}sa K. De Melo}
\author[4]{Silvia A. dos Reis}
\author[5]{Li Weigang}
\author[3,7]{Victor R. R. Celestino}

\affil[1]{Department of Economics, FACE, Universidade de Bras\'{i}lia (UnB), Campus Universit\'{a}rio Darcy Ribeiro, 70910-900, Bras\'{i}lia, Brazil. Email: danielcajueiro@gmail.com}

\affil[2]{Mechanic Engineering Department. Universidade de Bras\'{i}lia (UnB), Campus Universit\'{a}rio Darcy Ribeiro, 70910-900, Bras\'{i}lia, Brazil.}

\affil[3]{Department of Mathematics, 
Instituto Federal de Minas Gerais, Campus Formiga, 35577-020, Belo Horizonte, Brazil.
}

\affil[4]{Business Department, FACE, Universidade de Bras\'{i}lia (UnB), Campus Universit\'{a}rio Darcy Ribeiro, 70910-900, Bras\'{i}lia, Brazil.}

\affil[5]{Computer Science Department. Universidade de Bras\'{i}lia (UnB), Campus Universit\'{a}rio Darcy Ribeiro, 70910-900, Bras\'{i}lia, Brazil.}

\affil[6]{Nacional Institute of Science and Technology for Complex Systems (INCT-SC). Universidade de Bras\'{i}lia, Bras\'{i}lia, Brazil.}

\affil[7]{Machine Learning Laboratory in Finance and Organizations, FACE - Universidade de Bras\'{i}lia (UnB), Campus Universit\'{a}rio Darcy Ribeiro, 70910-900, Bras\'{i}lia, Brazil.}

\date{\today}
\begin{document}

\maketitle

\begin{abstract}
We provide a literature review about Automatic Text Summarization (ATS) systems. We consider a citation-based approach. We start with some popular and well-known papers that we have in hand about each topic we want to cover and we have tracked the
``backward citations" (papers that are cited by the set of papers we knew beforehand) and the ``forward citations" (newer papers that cite the set of papers we knew beforehand). In order to organize the different methods, we present the diverse approaches to ATS guided by the mechanisms they use to generate a summary. Besides presenting the methods, we also present an extensive review of the datasets available for summarization tasks and the methods used to evaluate the quality of the summaries.  Finally, we present an empirical exploration of these methods using the CNN Corpus dataset that provides golden summaries for extractive and abstractive methods.

\noindent {\bf Keywords}: Deep Learning, Machine Learning, Natural Language Processing, Summarization.
\end{abstract}


\clearpage

\tableofcontents

\clearpage

\section{Introduction}

Automatic Text Summarization (ATS) is the automatic process of transforming an original text document into a shorter piece of text, using techniques of Natural Language Processing (NLP), that highlights the most important information within it, according to a given criterion. 

There is no doubt that one of the main uses of ATS systems is that they directly address the information overload problem \citep{edmunds2000problem}. They allow a possible reader to understand the content of the document without having to read it entirely. Other ATS applications are keyphrase extraction \citep{hasan2014automatic}, document categorization \citep{brandow1995automatic}, information retrieval \citep{tombros1998advantages} and question answering \citep{morris1992effects}.

The seminal work in ATS systems field is due to \cite{Luhn1958} that used an approach that mixes information about the frequency of words with some heuristics to summarize the text of scientific papers. There are several different approaches to designing ATS systems today. In this paper, we intend to present a comprehensive literature review on this topic. This is not an easy task \citep{bullers2018takes}. First, there are thousands of papers, and we have to face the obvious question that is ``Which set of the works should we include in this review?". Second, the papers use very different approaches. Thus, the second important question is ``How do we present these papers in a comprehensive way?". We address the first question by adopting a citation-based approach. That means we start with a few popular\footnote{The popular papers are the ones more cited in the field.} and well-known papers about each topic we want to cover and we track the ``backward citations" (papers that are cited by the set of papers we knew beforehand) and the ``forward citations" (newer papers that cite the set of papers we knew beforehand). One clear challenge of this approach is to avoid the popularity bias so common in recommendation systems \citep{park2008long,hervas2008word,fleder2009blockbuster}. We deal with this challenge by trying to consider papers that cover different dimensions of the approach we are reviewing. In order to answer the second question, we have tried to present the diverse approaches to ATS guided by the mechanisms they use to generate a summary.

Our paper naturally relates to other reviews about this theme. We may classify these reviews in terms of classical such as \cite{edmundson1961automatic} and \cite{paice1990constructing}, topic-specific such as \cite{rahman2015survey} (query-based summarization), \cite{pouriyeh2018comprehensive} (ontology-based summarization), \cite{jalil2021extractive} (extractive multi-document summarization)  and \cite{alomari2022deep} (deep learning approaches to summarization), and general reviews like ours such as \cite{mridha2021survey} and \cite{el2021automatic}. Although these latter works are very related to ours in terms of general content, the presentation of our work is very different. The models and mechanisms used to build such summaries drive our presentation. Thus, our focus on models and mechanisms used in automatic text summarization aims to provide practical guidance for researchers or practitioners who are developing such systems. By emphasizing these aspects of summarization, our review has the potential to offer unique insights that are not covered by other works in the field, and may help to bridge the gap between the technique used to build the model and the practical application in summarization. Furthermore, besides presenting the models used to generate the summaries, we also present the most popular datasets, a compendium of evaluation techniques, and an exploration of the public python libraries that one can use to implement the task of ATS\footnote{The interested reader may find the complete code used to explore these libraries in the Zenodo: \url{https://zenodo.org/record/7500273}.}.

We organize the manuscript as follows: Section \ref{sec:classification} presents a taxonomy used to classify ATS systems. Section \ref{sec:overview_survey} summarizes the content of other surveys about ATS systems. Section \ref{sec:datasets} describes the datasets used to explore ATS systems. Section \ref{sec:general} illustrates the basic topology of an ATS system. In Section \ref{sec:extractive}, we present the approaches to extractive summarization. We split this section into the following subsections: Subsection \ref{sec:extractive_frequency_based} presents the frequency-based methods. Subsection \ref{sec:extractive_heuristics} presents the heuristic-based methods. Subsection \ref{sec:extractive_linguistic} presents the linguistic-based methods. Subsection \ref{sec:extractive_ml} presents the methods based on supervised machine learning models. Subsection \ref{sec:extractive_reinforcement} presents the reinforcement-learning-based approaches. Section \ref{sec:abstractive} presents the approaches to abstractive summarization.
We divide this section into two subsections. While Subsection \ref{sec:linguistic_abstractive} introduces the linguistic approaches, Subsection \ref{sec:abstractive_sequence_seq2seq} describes the deep learning sequence-to-sequence approaches. Section \ref{sec:compressive_extractive} introduces the compressive extractive hybrid approaches. Section \ref{sec:evaluation} describes the methods used to evaluate ATS systems. Section \ref{sec:libraries} presents the public libraries available in Python and Section \ref{sec:comparison_exercises} explores these libraries in the CNN Corpus dataset \citep{lins2019cnn}\footnote{Our rationale for selecting the CNN dataset is that it stands out as one of the few datasets that provides both extractive and abstractive reference summaries.}, which presents both extractive and abstractive golden summaries for every document\footnote{It is common in this literature to call the reference human-made summaries as the {\it gold-standard summaries}.}.  Finally, Section \ref{sec:final_remarks} presents the main conclusions of this work. 

\section{Classification of ATS systems}
\label{sec:classification}

This section presents some of the different criteria used to build a taxonomy for ATS systems \citep{Jones1998, Hovy1999}: 

\begin{enumerate}

    \item The type of output summary: We may classify a summary into {\it extractive}, {\it abstractive} and {\it hybrid}. While an extractive approach extracts from the text the most important sentences and joins them in order to form the final summary, an abstractive method extracts the main information from the text and rewrites it in new sentences to form the summary. Although humans usually summarize pieces of text in an abstractive way, this approach is more difficult for machines since it depends on a language model to rewrite the sentences. On the other hand, a hybrid approach combines ingredients of both approaches. A compressive extractive approach extracts the most relevant sentences in the first step and requires a language model in order to compress the sentences using only essential words in the second step. We begin our paper by categorizing the methods based on the type of output summary they generate. There are two important reasons for this. Firstly, these three categories represent distinct and well-established approaches to summarization, each with its own set of advantages and limitations. Secondly, datasets with golden summaries often follow a similar division into abstractive, extractive, and compressive abstractive approaches. Thus,  Section \ref{sec:extractive} presents the extractive approaches, Section \ref{sec:abstractive} presents the abstractive approaches and Section \ref{sec:compressive_extractive} presents the hybrid compressive extractive approaches.

    \item The type of available information: We may classify a summary into {\it indicative} or {\it informative}. While the former case calls the attention of the reader to the content we may find in a text document, the objective of the latter case is to present the main findings of the text. Thus, while in the first case the summary intends to be an advertisement of the content of the text, in the second case the reader only reads the main text if he/she wants to learn more about a given result. While most approaches reviewed here and available in the literature are typically indicative, there are some examples of structured approaches that allow for retrieval of the main findings of a text. We may find an example of the informative approach in Section \ref{sec:linguistic_abstractive}. For instance, \cite{genest2012fully} use handcrafted information extraction rules to extract the information they need to build the summary. In particular, they ask questions about the nature of the event, the time, the location and other relevant information in their context.  
    
    \item The type of content: We may classify a summary into {\it generic} and {\it query-based}. While a query-based system intends to present a summary that focuses on keywords previously fed by the user, the generic summary is the opposite.  Most query-based systems are minor modifications of generic ATS systems. For instance, \cite{darling2010multi}, reviewed in Section \ref{sec:extractive_vector_space}, in a generic summarization setup, extracts the most important sentences of a document using information about the distribution of terms in the text. In order to provide a query-based approach, it adds more probability mass to the bins of the terms that arise in the query.  In our work, we call the attention of the reader when the ATS intends to be a query-based system.
 
    \item The number of input documents: We may classify the summary in terms of being a {\it single}-document or a {\it multi}-document summary. The first case happens when the summary is built from only one document and the second case happens when the summary comes from the merging of many documents. It is worth mentioning that multi-document summarization has received growing attention due to the need to automatically summarize the exponential growth of online material that presents many documents with similar tenor. Thus, many sentences in different documents overlap with each other, increasing the need to recognize and remove redundancy.  
    
    In general, multi-document cases present a more serious problem of avoiding redundant sentences and additional difficulties in the concatenation of the sentences in the final summary. It is worth mentioning that many approaches presented in this review present an instance of the multi-document summarization task, and we call the reader's attention when it happens. 
    
    A more recent approach to multi-document summarization is known as {\it update summarization}. The idea of update-based summarization is to generate short multi-document summaries of recent documents under the assumption that the earlier documents were previously considered. Thus, the objective of an update summary is to update the reader with new information about a particular topic and the ATS system has to decide which piece of information in the set of new documents is novel and which is redundant.  
    
    We may find another kind of multi-document summarization if we consider jointly to summarize the original document and the content generated by the users (such as comments or other online network contents) after the publication of the original document. This approach of summarization is known as {\it social context summarization}. We may find examples of social context summarization in Sections \ref{sec:extractive_matrix_factorization} and \ref{sec:extractive_ml}.
    
    \item The type of method used to choose the sentences to be included in the summary: We may classify it in terms of being a {\it supervised} or an {\it unsupervised} method. This is particularly important because while in the former case we need data to train the model, in the latter case that is not necessary. Supervised methods arise in Sections \ref{sec:extractive_ml}, \ref{sec:abstractive_sequence_seq2seq} and \ref{sec:compressive_extractive}.

\end{enumerate}

\section{An overview of other ATS surveys}
\label{sec:overview_survey}

Table \ref{tab:digest_survey_summarization} presents an overview of other surveys about ATS systems. The first column presents the source document. The second column presents a summary of its content. The third column presents the date range of papers cited in the survey. The last column presents the number of papers cited in the review. The intention of the last two columns is to provide an indication of the coverage of the work.  

The most complete surveys to date are the ones presented in \cite{el2021automatic} and \cite{mridha2021survey}. Like ours, they intend to cover most aspects of ATS systems. However, we may find differences among them in terms of content and presentation. Although there is no doubt that most classical papers are present in our work and also in these two works, the presentation of our work is naturally model-guided. In terms of content, our work also presents additional sections that are not available elsewhere, namely Section \ref{sec:libraries} presents the main libraries available for coding ATS systems and Section \ref{sec:comparison_exercises} presents a comparison of the most popular methods of ATS systems, using a subset of the libraries presented in Section  \ref{sec:libraries}, and using the most popular methods to evaluate these methods presented in Section \ref{sec:evaluation}.

In terms of coverage, our work cites \total{citnum}\ references from 1958 to 2022 and it is one of the most complete surveys of the field. It is important to emphasize that the number of citations and the date range are just indicators of the coverage. Table \ref{tab:digest_survey_summarization} presents some very influential surveys with a much smaller number of references. We make a special reference to the amazing classical works \cite{edmundson1961automatic}, \cite{paice1990constructing}, \cite{Jones1998} and the more recent works \cite{nenkova2011automatic} and \cite{Llret2012}.

\begingroup
\renewcommand{\arraystretch}{1.2}
\begin{landscape}
\begin{table}[t]
\tiny
    \centering
    \begin{tabular}{p{3.8cm}p{10cm}ll}
     Source    & Focus & Date range & References \\\hline
    \cite{edmundson1961automatic} & It is an amazing survey of the early ATS systems.  & 1953-1960 & 7 \\\cdashline{1-4}
    \cite{paice1990constructing}     & It presents the classical extractive ATS systems.  & 1958-1989 & 52  \\\cdashline{1-4}
    \cite{Jones1998} & It explores the context, input, purpose, and output factors necessary to develop effective approaches to ATS. & 1972-1997 & 26 \\\cdashline{1-4}
    \cite{das2007survey}     & It presents approaches to extractive and abstractive summarization. It also presents an overview of the evaluation methods.    &  1958-2007 & 48 \\\cdashline{1-4}  
    \cite{Gholamrezazadeh2009} & It reviews techniques of extractive summarization. & 1989-2008 & 24 \\\cdashline{1-4}
    \cite{damova2010query}  & It reviews techniques for query-based extractive summarization. & 2005-2009 & 11 \\\cdashline{1-4}
    \cite{gupta2010survey} & It presents a survey of extractive summarization techniques. & 1958-2010 & 47 \\\cdashline{1-4}
    \cite{nenkova2011automatic} & It presents a fantastic general survey about ATS. &1958-2010& 236\\\cdashline{1-4}
    \cite{Llret2012}     & It presents a great overview of the theme that includes both abstractive and also extractive ATS techniques. It discusses the taxonomy of ATS systems. It  combines ATS systems with intelligent systems such as information retrieval systems, question-answering systems and text classification systems. It also presents an overview of the techniques used to evaluate summaries.& 1958-2012 &  197\\\cdashline{1-4}
    \cite{kumar2012automatic} & It presents a survey of multi-document summarization. & 1998-2012 & 36 \\\cdashline{1-4}
    \cite{dalal2013survey} & It presents a very short overview of the bio-inspired methods of text summarization. & 1997-2011 & 7 \\\cdashline{1-4}
    \cite{ferreira2013assessing} & It presents an overview of sentence scoring techniques for extractive text summarization. & 1958-2013 & 35 \\\cdashline{1-4}
    \cite{munot2014comparative} & It presents the methods of extractive and abstractive summarization. & 1958-2014 & 19 \\\cdashline{1-4}
    \cite{mishra2014text} & It reviews the works  of ATS in the biomedical domain. & 1969-2014 & 53 \\\cdashline{1-4}
    \cite{saranyamol2014survey}     & It describes different approaches to the automatic text summarization process including both extractive and abstractive methods. & 2007-2014 & 7 \\\cdashline{1-4}
    \cite{rahman2015survey}     & It reviews techniques for query-based extractive summarization. & 1994-2014 & 34 \\\cdashline{1-4}
    \cite{meena2015evolutionary} & It presents a survey of extractive ATS systems evolutionary-based approaches. & 2001-2012 & 16 \\\cdashline{1-4}  
    \cite{andhale2016overview} & It presents a survey of extractive and abstractive ATS approaches. & 1998-2015 & 66 \\\cdashline{1-4}
    \cite{mohan2016study} & It presents a survey on ontology-based abstractive summarization.& 1997-2014 & 22 \\\cdashline{1-4}
    \cite{moratanch2016survey} & It presents a survey of abstractive text summarization. & 1999-2016 & 21 \\\cdashline{1-4}
    \cite{jalil2021extractive} & It presents a survey of multi-document summarization. & 1989-2016 &18  \\\cdashline{1-4}
    \cite{gambhir2017recent} & It presents a very general survey of extractive and abstractive methods. It also presents a survey of  evaluation methods and the results found in DUC datasets. & 1958-2016 & 186\\\cdashline{1-4}
    \cite{allahyari2017text} & It presents a survey of extractive ATS approaches. & 1958-2017 & 81 \\\cdashline{1-4}
    \cite{bharti2017automatic} & It presents a very general survey of extractive and abstractive methods. It also presents a survey of available datasets used to investigate ATS systems and evaluation methods.   & 1957-2016 & 132 \\\cdashline{1-4}
    \cite{pouriyeh2018comprehensive}    & It presents an overview of the ontology-based summarization methods. & 1966-2017 & 43  \\\cdashline{1-4}
    \cite{dernoncourt2018repository}& It presents an overview of the available corpora for summarization. & 1958-2018 & 75  \\\cdashline{1-4}
    \cite{gupta2019abstractive}     & It presents the methods of abstractive summarization. & 2000-2018 & 109 \\\cdashline{1-4}
    \cite{tandel2019review} & It surveys the neural network-based abstractive text summarization approaches. & 2014-2018 & 8 \\\cdashline{1-4}
    \cite{klymenko2020automatic} & It presents a general overview of summarization methods, including recent trends. & 1958-2020 & 54\\\cdashline{1-4}
    \cite{awasthi2021natural} & It presents a general overview of summarization methods including very recent works.& 2001-2021 & 37\\\cdashline{1-4}
    \cite{sheik2021deep} & It presents an overview of deep learning for legal text summarization. & 2004-2021 & 23 \\\cdashline{1-4}
    \cite{mridha2021survey}     & It presents a very general survey of extractive and full abstractive methods. It also presents a survey of available datasets used to investigate ATS systems and evaluation methods.   & 1954-2021 & 353 \\\cdashline{1-4}
    \cite{el2021automatic}     & It presents a very general survey of extractive and abstractive methods. It also presents a survey of available datasets used to investigate ATS systems and evaluation methods.  & 1954-2020 & 225 \\\cdashline{1-4}
    \cite{jalil2021extractive} & It presents a survey of extractive multi-document summarization. & 1998-2020 &81  \\\cdashline{1-4}
     \cite{alomari2022deep} & It presents a survey of the approaches based on deep learning, reinforcement learning and transfer learning used for abstractive summarization. It also presents a survey of datasets used in this field, evaluation techniques and results. & 1953-2022 & 205
    \\\hline
    \end{tabular}
    \caption{A representative compilation of other ATS surveys.}
    \label{tab:digest_survey_summarization}
\end{table}
\end{landscape}
\endgroup

\section{Datasets}
\label{sec:datasets}

There is today a large number of datasets that we may use to explore the task of ATS. The datasets may belong to a variaty of domains,  they may be suitable to evaluate  different tasks of summarization, they are in different sizes and they may present a different number of gold-summaries.
For each dataset discussed in the following lines, Table \ref{tab:digest_datasets} presents detailed information about them. This table has a total of nine columns: (1) name of the dataset; (2) language; (3) domain (e.g. news, scientific papers, reviews, etc.); (4)  number of single-documents; (5) number of multi-documents; (6) number of gold-standard summaries per document in the case of single-documents; (7) number of gold-standard summaries per document in the case of multi-documents; (8) URL where we may find the dataset; and (9) the work that presents the dataset. Our primary focus is on datasets containing summaries for texts written in the English language. However, if a method referenced in this review is evaluated using a dataset with texts written in other languages, we also include this dataset in our discussion.

\paragraph{BIGPATENT} \cite{sharma2019bigpatent} introduce the BIGPATENT dataset that provides good examples for the task of abstractive summarization. They build the dataset using Google Patents Public Datasets, where for each document there is one gold-standard summary which is the patent's original abstract. One advantage of this dataset is that it does not present difficulties inherent to news summarization datasets, where summaries have a flattened discourse structure and the summary content arises at the beginning of the document.

\paragraph{BillSum} \cite{kornilova2019billsum}, in order to fill the gap that there is a lack of datasets that deal specifically with legislation, introduce the BillSum dataset. Their documents are bills collected from the United States Publishing Office's Govinfo. Although the dataset focuses on the task of single-document extractive summarization, the fact that each bill is divided into multiple sections makes the problem akin to that of multi-document summarization. Each document is accompanied by a gold-standard summary and by its title.

\paragraph{Blog Summarization Dataset} \cite{ku2006opinion} deal with three NLP tasks related to opinions in news and blog corpora, namely opinion extraction, tracking, and summarization. Concerning summarization, they tackle the problem from the perspective of sentiment analysis at the levels of word, sentence, and document. The authors gather blog posts that express opinions regarding the genetic cloning of the Dolly sheep and they give the task to tag the texts in each one of these levels to three annotators. A comparison of their opinions generates gold-standard words, sentences, and documents that expressed positive or negative opinions. From the categorization made by the annotators, two kinds of gold-standard summaries are generated for the set of positive/negative documents: one is simply the headline of the article with the largest amount of positive/negative sentences (brief summary) and the other is the listing of the sentences with the highest sentiment degree (detailed summary).

\paragraph{CAST} \cite{hasler-CL-03} built the CAST corpus with the intention of having a more detailed dataset to be used in the task of extractive ATS. For that purpose, they provide annotations for each document signaling three types of sentences. The crucial sentences labeled as essential are those without which the text can not be fully understood. The important sentences provide important details of the text, even if they are not absolutely necessary for its understanding. The third group of sentences is comprised of the ones that are not important or essential.  Another advantage of the dataset is that it also contains extra pieces of information about the essential and important sentences. It presents annotations for ``removable parts'' within the essential and important sentences and it indicates linked sentences, which are two sentences labeled as essential or important that need to be paired together for understanding. It is worth mentioning that three graduate students (who were native English speakers and one post-graduate student -- who had advanced knowledge of the English language annotation) were responsible for providing the annotations for this dataset. The number of summaries per document depends on the number of annotators for each document. 

\paragraph{CNN Corpus} \cite{lins2019cnn} introduce the CNN Corpus dataset, comprised of 3,000 Single-Documents with two gold-standard summaries each: one extractive and one abstractive. The encompassing of extractive gold-standard summaries is also an advantage of this particular dataset over others, which usually only contain abstractive ones.

\paragraph{CNN/Daily Mail} \cite{hermann2015teaching} intend to develop a consistent method for what they called ``teaching machines how to read", i.e., making the machine able to comprehend a text via Natural Language Processing techniques. In order to perform that task, they collect around 400k news from CNN and Daily Mail and evaluate what they consider to be the key aspect in understanding a text, namely the answering of somewhat complex questions about it. Even though ATS is not the main focus of the authors, they took inspiration from it to develop their model and include the human-made summary for each news article in their dataset.

\paragraph{CWS Enron Email} \cite{carenini2007summarizing} finds that email ATS systems are becoming quite necessary in the current scenario: users who receive lots of emails do not have time to read them entirely, and reading emails is an especially difficult task to be done in mobile devices. The authors, then, develop an annotated version of the very large Enron Email dataset -- which is described in more detail by \cite{shetty2004enron} -- in which they select 20 email conversations from the original dataset and hired 25 human summarizers (who were either undergraduate or graduate students of different fields of study) to write gold-standard summaries of them. 

\paragraph{DUC} \cite{over2007duc} present an overview of the datasets provided by the Document Understanding Conferences (DUC) until 2006 and \cite{dernoncourt2018repository} provides useful information concerning DUC 2007. If we take a look at Tables \ref{tab:digest_vector_space_models}, \ref{tab:digest_latent}, \ref{tab:digest_graph_methods}, \ref{tab:digest_topic_model}, \ref{tab:digest_word_embeddings}, \ref{tab:digest_heuristics}, \ref{tab:digest_extractive_linguistic}, \ref{tab:digest_suppervised}, \ref{tab:digest_reinforcement}, \ref{tab:digest_abstractive_linguistic}, \ref{tab:digest_transformers_abstractive} and \ref{tab:digest_compressive_extractive}, we can note that DUC datasets are the ones most commonly used by researchers in the task of text summarization. The DUC datasets from 2001 to 2004 contain examples of both single-document and multi-document summarization with each document in each cluster having gold-standard summaries associated and each cluster having its own set of gold-standard summaries. The DUC datasets from 2005 to 2007 focus only on multi-document summarization.

\paragraph{Email} \cite{zhang2019email} note that the subject line is an important feature for going through emails in an efficient manner while surfing through received messages, and an ATS method that performs the task of generating such lines -- what the authors called Subject Line Generation (SLG) -- is of much help. The authors decide, therefore, to compile the dataset by annotating the existing single-document Enron dataset and providing each of its documents with three gold-standard human written summaries. Abstractive summarization fits the desired purposes, especially because of the high compression ratio required by the fact that subject lines are indeed very small.

\paragraph{Gigaword 5} \cite{napoles-etal-2012-annotated} provide a useful description of the dataset -- introduced for the first time by \cite{graff2003english} -- which contains almost 10 million news as its single-documents, each with its own abstractive gold-standard summary. The Gigaword Dataset is very commonly used by researchers for the purpose of training ATS neural networks due to the astronomical amount of documents it contains. However, this dataset only works for extreme summarization exercises since the summaries provided by the dataset are the headlines associated with each document.

\paragraph{GOVREPORT} \cite{huang2021efficient}, in order to study encoder-decoder attention deep-learning-based ATS systems, built the GOVREPORT dataset. 
This dataset contains about 19k large documents with an average of 9,000 words of government reports published by the U.S. Government Accountability Office (GAO), each accompanied by a gold-standard summary written by an expert.

\paragraph{Idebate} \cite{wang2016neural} develop the dataset to go along with Movie Review in order to perform their desired task of multi-documents abstractive ATS of opinions. This dataset contains data retrieved from the argumentation website idebate.com. The Idebate dataset is composed of 2,259 clusters of arguments, each with its respective gold-standard summary written manually by an editor.

\paragraph{Multi-News} \cite{fabbri-etal-2019-multi} decide to come up with the dataset after considering the fact that, while there are very large single-document datasets that deal with news summarization, when it comes to the task of multi-document ATS the number of documents available in the most used datasets is very scarce. Multi-News focuses on abstractive summarization and draws its data from the newser.com website, with each cluster of documents having its own human-written gold-standard dataset. There are about 56k clusters with varying numbers of source documents in the dataset.

\paragraph{NEWSROOM} \cite{grusky2018newsroom} aim at achieving the goal of producing a wide and diverse enough dataset that can be used in order to evaluate the level of extractiveness/abstractiveness of ATS summaries. The NEWSROOM dataset, then, consists of about 1.3M news articles on many topics such as day-to-day life, movies, games, and so on, and each document is accompanied by a human-written gold-standard summary extracted from its HTML metadata, besides its original title.

\paragraph{Opinosis} \cite{ganesan2010opinosis} introduce the Opinosis summarization framework and with it the dataset of the same name. The authors have the goal of summarization of redundant opinions about a certain topic -- usually product reviews -- from different users. A particular characteristic of this dataset is that it focuses on abstractive summarization and aims at generating relatively small summaries that could easily be read on mobile devices. 

\paragraph{Reddit TIFU} \cite{kim2018abstractive} use the Today I F*** Up (TIFU) subreddit to build a single-document-based dataset, which rules that each story must include one short summary and one long summary at the beginning and end of the post. Thus, this dataset is especially convenient for retrieving gold-standard short and long summaries.

\paragraph{Rotten Tomatoes} \cite{wang2016neural} gather data from the RottenTomatoes.com movie review website with the goal of building a robust dataset to perform the task of multi-document opinion ATS. The main difference between the Rotten Tomatoes dataset and other multi-document opinion datasets is its focus being largely on abstractive summarization. It contains clusters of reviews for 3,731 movies and each of them is associated with a gold-standard human-written summary by an editor.

\paragraph{SAMSum Corpus} \cite{gliwa2019samsum} introduce the SAMSum Corpus dataset, which consists of single-document text message-like dialogues. The documents are fabricated by linguists and encompass a wide range of levels of formality and topics of discussion. The dataset contains examples of both formal and informal dialogues in a wide range of contexts, such as a usual conversation between friends or a political discussion. 

\paragraph{Scientific papers (arXiv \& Pubmed)} \cite{cohan2018discourse} use scientific papers as a source for a dataset with large documents with abstractive summaries. Thus, the authors compile a new dataset with arXiv and PubMed databases. Scientific papers are especially convenient due to their large length and the fact that each one contains an abstractive summary made by its author. The union of both the arXiv and PubMed datasets is one of the available ATS TensorFlow datasets \citep{Abadi2015TensorFlow}.

\paragraph{Scisummnet} \cite{yasunaga2019scisummnet} tackle the task of scientific papers ATS because they found the existing literature lacking in a number of respects. The two major ones are the scarcity of documents contained in the most used datasets for that purpose and the fact that the generated summaries do not contain crucial information such as the paper's impact on the field. They then built a dataset gathering the 1,000 most cited papers in the ACL Anthology Network (AAN) and each one was given a gold-standard summary written by an expert in the field. The difference between such summaries and those of other similar datasets is that they take into account the context in which the paper was cited elsewhere so that it is possible to obtain information on what exactly is its relevance in the field.

\paragraph{SoLSCSum} \cite{nguyen2016solscsum} introduce this dataset  for the task of social context summarization. This dataset includes articles and user comments collected from Yahoo News.  Each sentence or comment has a label indicating whether the piece of text is important or not. 

\paragraph{SummBank 1.0} \cite{radev2003evaluation} aim at evaluating eight single and multi-document summarizers. Having the Hong Kong News Corpus (LDC\footnote{Linguistic Data Consortium.} number LDC2000T46) as a basis, the authors build their own corpus which consists of 20 clusters of documents removed from the above-mentioned dataset, ranging from a variety of topics. For evaluation, they collect a total of 100 Million gold-standard automatic summaries at ten different lengths -- generated by human-annotated sentence-relevance analysis. The authors also provide more than 10,000 human-written extractive and abstractive summaries and 200 Million automatic document and summary retrievals using 20 queries.

\paragraph{TAC} After the DUC events ceased to happen, the Text Analysis Conference (TAC) was founded as a direct continuation of it. Its organization is similar to DUC's and the 2008-2011 editions focused on multi-document summarization, following the later DUC editions trend. The perhaps most interesting aspect of the datasets provided by TAC 2008-2011 is that they focus on guided summarization, which aims at generating an ``update summary'' after a multi-document summary is already available. We may find a good overview of the contents of each of the cited TAC datasets in \cite{dernoncourt2018repository}.

\paragraph{TeMário} The TeMário \citep{pardo2003temario} dataset contains 100 news articles -- which covers a variety of topics, ranging from editorials to world politics -- from the newspapers \textit{Jornal do Brasil} e \textit{Folha de São Paulo}, each accompanied by its own gold-standard summary written by an expert in the Brazilian Portuguese language. It is one of the datasets used by \cite{cabral2014platform}, who wanted to address the known problem of multilingual automatic text summarization and the fact that most summarization datasets and methods focus almost exclusively on the summarization of texts in the English language. Taking that into account, they propose a language-independent method for ATS developed using multiple datasets in languages other than English.

\paragraph{TIPSTER SUMMAC} \cite{mani1999tipster} aims at developing a method for evaluating ATS-generated summaries of texts. In order to do that, they apply the so-called TIPSTER Text Summarization Evaluation (SUMMAC), which was completed by the U.S. Government in May 1998. For the performance of the desired task, the authors select a range of topics in the news domain (for the most part, since there are some letters to the editor included as well) and chose 50 from the 200 most relevant articles published in that topic. For each document in the dataset, there are two gold-standard summaries: one of fixed length (S1) and one which was not limited by that parameter (S2).

\paragraph{2013 TREC} \cite{aslam2013trec}, \cite{liu2013ictnet} and \cite{yang2013bjut} provide useful overviews of the Temporal Summarization Track which occurred for the first time at TREC 2013. The task consists in generating an updated summarization summary of multiple timestamped documents from news and social media sources extracted from the TREC KBA 2013 Stream Corpus. Update summarization is convenient, for example, when dealing with so-called crisis events, such as hurricanes, earthquakes, shootings, etc. that require useful information to be quickly available to those involved in them. The gold-standard updates -- which are called nuggets -- are extracted from the event's Wikipedia page and are timestamped according to its revision history since facts regarding the event are included as they happen. With the nuggets in hand, human experts assigned to them a relevance grade -- to make possible proper evaluation -- ranging from 0-3 (no importance to high importance), and an annotated dataset could be generated.

\paragraph{2014 TREC} \cite{zhao2014bjut} present TREC 2014's Temporal Summarization Track in a useful manner, highlighting the differences in comparison to the previous year's edition. The task focused on the Sequential Updates Summarization task and participants have to perform the update summarization of multiple documents contained in the TREC-TS-2014F Corpora, which is a filtered -- and therefore reduced -- version of the track's full Corpora. The data size is also reduced in comparison to the previous year's, going from a size of 4.5 Tb to 559 Gb. An annotated dataset is produced as a byproduct of the track.

\paragraph{2015 TREC} \cite{aliannejadi2015university} give an overview of TREC 2015's Temporal Summarization Track and their participation in it. Participants are given two datasets, namely the TREC-TS-2015F and the TREC-TS-2015F-RelOnly which have smaller sizes when compared to the KBA 2014 corpus. TREC-TS-2015F-RelOnly is a filtered version of the TREC-TS-2015F, which contains many irrelevant documents. The assembly of the annotated dataset is similar to that of the previous years.

\paragraph{USAToday-CNN} \cite{nguyen2017exploiting} create this dataset for the task of social context summarization. This dataset includes events retrieved from USAToday and CNN and tweets associated with the events. Each sentence and each tweet have a label indicating whether the piece of text is important or not.

\paragraph{VSoLSCSum} \cite{nguyen2016vsolscsum}, in order to validate their models of social context summarization, create this non-English language dataset. This dataset includes news articles and their relevant comments collected from several Vietnamese web pages. Each sentence and comment have a label indicating whether the piece of text is important or not. 

\paragraph{XSum} \cite{narayan2018don} introduce the single-document dataset, which focuses on abstractive extreme summarization, such as in \citep{napoles-etal-2012-annotated},  that intends to answer the question ``What is the document about?". They build the dataset with BBC articles and each summary is accompanied by a short gold-standard summary often written by the author of the article. 

\paragraph{XLSum} \cite{hasan2021xl} aim at solving the problem of a lack of sources dealing with the problem of abstractive multi-lingual ATS. To perform that task, they build the XLSum dataset: a dataset composed of more than one million single documents in 44 different languages. The documents are news articles extracted from the BBC database, which is a convenient source since BBC produces articles in a multitude of countries -- and therefore languages -- with a consistent editorial style. Each document is accompanied by a small abstractive summary in every language, written by the text's author, which is used as the gold-standard summary for the purpose of evaluation.

\paragraph{WikiHow} \cite{koupaee2018wikihow} explore a common theme in the development of new datasets and ATS methods, namely that most datasets are limited by the fact that they deal entirely with the news domain. The authors, then, developed the WikiHow dataset with the desire that it would be used in a generalized manner and in a multitude of ATS applications. The dataset consists of about 200k Single-Documents extracted from the WikiHow.com website, which is a platform for posting step-by-step guides to performing day-to-day tasks. Each article from the website consists of a number of steps and each step starts with a summary in bold of its particular content. The gold-standard summary for each document is the concatenation of such bold statements.

\begingroup
\renewcommand{\arraystretch}{1.3}
\begin{landscape}
\begin{table}[p]
    \centering
    \tiny
    \label{tab:digest_datasets}
    \resizebox{24cm}{!}{
    \begin{tabular}{p{2.4cm}p{1.1cm}p{2.41cm}p{1.8cm}p{1.8cm}p{1.5cm}p{1.5cm}p{6.52cm}p{3.5cm}}
     \vtop{\hbox{}\hbox{}\hbox{Dataset}} & 
     \vtop{\hbox{}\hbox{}\hbox{Language}} & 
     \vtop{\hbox{}\hbox{}\hbox{Domain}} & 
     \vtop{\hbox{Summarized}\hbox{single-}\hbox{documents}} & 
     \vtop{\hbox{Summarized}\hbox{multi-}\hbox{documents}} & 
     \vtop{\hbox{Gold-standard}\hbox{summaries per}\hbox{document}} & 
     \vtop{\hbox{Gold-standard}\hbox{summaries per}\hbox{cluster}} &  
     \vtop{\hbox{}\hbox{}\hbox{URL}} & 
     \vtop{\hbox{}\hbox{}\hbox{Source article}} \\\hline
       arXiv & English & Scientific papers & 215,000 &  & 1 &  & \url{https://arxiv.org/help/bulk\_data} & \cite{cohan2018discourse}\\ \cdashline{1-9}
       BIGPATENT & English & Patent documents & 1,341,362 &  & 1 &  & \url{https://evasharma.github.io/bigpatent} & \cite{sharma2019bigpatent}\\ \cdashline{1-9}
       BillSum & English & State bills & 19,400 &  & 1 &  & \url{https://github.com/FiscalNote/BillSum} & \cite{kornilova2019billsum}\\ \cdashline{1-9}
       Blog Summarization & Chinese & Opinion blog posts &  & 1x20 &  & 4 & \url{http://cosmicvariance.com}\& \url{http://blogs.msdn.com/ie} & \cite{ku2006opinion}\\\cdashline{1-9}
       CAST & English & News, Science texts & 163 &  & Varies  &  & \url{http://clg.wlv.ac.uk/projects/CAST/corpus/index.php} & \cite{hasler-CL-03}\\ \cdashline{1-9}
       CNN Corpus & English & News & 3,000 &  & 2 &  & Available upon email request to the authors & \cite{lins2019cnn} \\ \cdashline{1-9}
       CNN/Daily Mail & English & News & 312,085 &  & 1 &  & \url{https://github.com/deepmind/rc-data} & \cite{hermann2015teaching}\\ \cdashline{1-9}
       CWS Enron Email & English & E-mails &  & 20 &  & 5 & \url{https://github.com/deepmind/rc-data} & \cite{carenini2007summarizing}\\ \cdashline{1-9}
       DUC 2001 & English & News & 600 & 60x10 & 1 & 4 & \url{https://www-nlpir.nist.gov/projects/duc/data.html} & \cite{over2007duc}\\ \cdashline{1-9}
       DUC 2002 & English & News & 600 & 60x10 & 1 & 6 & \url{https://www-nlpir.nist.gov/projects/duc/data.html}  & \cite{over2007duc}        \\ \cdashline{1-9}
       DUC 2003 & English & News & 1,350 & 60x10, 30x25 & 1 & 3 & \url{https://www-nlpir.nist.gov/projects/duc/data.html}  & \cite{over2007duc} \\ \cdashline{1-9}
       DUC 2004 & English & News & 1,000 & 100x10 & 1 & 2 & \url{https://www-nlpir.nist.gov/projects/duc/data.html}  & \cite{over2007duc} \\ \cdashline{1-9}
       DUC 2005 & English & News &  & 50x32 &  & 1 & \url{https://www-nlpir.nist.gov/projects/duc/data.html}  & \cite{over2007duc} \\ \cdashline{1-9}
       DUC 2006 & English & News &  & 50x25 &  & 1 & \url{https://www-nlpir.nist.gov/projects/duc/data.html}  & \cite{over2007duc} \\ \cdashline{1-9}
       DUC 2007 & English & News &  & 25x10 &  & 1 & \url{https://www-nlpir.nist.gov/projects/duc/data.html}  & \cite{dernoncourt2018repository}\\\cdashline{1-9}
       Email & English & Emails & 18,302 &  & 3 &  & \url{https://github.com/ryanzhumich/AESLC} & \cite{zhang2019email}\\ \cdashline{1-9}
       Gigaword 5 & English & News & 9,876,086 &  & 1 &  & \url{https://catalog.ldc.upenn.edu/LDC2011T07} & \cite{graff2003english} \\ \cdashline{1-9}
       GOVREPORT & English & Documents & 19,466 &  & 1 &  & \url{https://gov-report-data.github.io} & \cite{huang2021efficient}\\ \cdashline{1-9}
       Idebate & English & Debate threads & Varies  &  & 1 &  & \url{https://web.eecs.umich.edu/~wangluxy/data.html} & \cite{wang2016neural}\\ \cdashline{1-9}
       Multi-News & English & News &  & Varies  &  & 1 & \url{https://github.com/Alex-Fabbri/Multi-News} & \cite{fabbri-etal-2019-multi}\\ \cdashline{1-9}
       NEWSROOM & English & News & 1,321,995 &  & 1 &  & \url{https://github.com/lil-lab/newsroom} & \cite{grusky2018newsroom}\\ \cdashline{1-9}
       Opinosis & English & Reviews &  & 51x100 &  & 5 & \url{http://kavita-ganesan.com/opinosis-opinion-dataset} & \cite{ganesan2010opinosis}\\ \cdashline{1-9}
       PubMed & English & Scientific papers & 133,000 &  & 1 &  & \url{https://pubmed.ncbi.nlm.nih.gov/download} & \cite{cohan2018discourse}\\ \cdashline{1-9}
       Reddit TIFU & English & Blog posts & 122,933 &  & 2 &  & \url{https://github.com/ctr4si/MMN} & \cite{kim2018abstractive}\\ \cdashline{1-9}
       Rotten Tomatoes & English & Movie reviews & Varies  &  & 1 &  & \url{https://web.eecs.umich.edu/~wangluxy/data.html} & \cite{wang2016neural}\\ \cdashline{1-9}
       SAMSum Corpus & English & News & 16,369 &  & 1 &   & \url{https://github.com/Alex-Fabbri/Multi-News} & \cite{gliwa2019samsum}\\ \cdashline{1-9}
       Scisummnet & English & Scientific papers &  & 1,000 &  & 1 & \url{https://cs.stanford.edu/~myasu/projects/scisumm\_net} & \cite{yasunaga2019scisummnet}\\ \cdashline{1-9}
        SoLSCSum & English & News &  &    157 &  &  & \url{http://150.65.242.101:9292/yahoo-news.zip} &\cite{nguyen2016solscsum}\\ \cdashline{1-9}
       SummBank 1.0 & English, Chinese & News & 400 (English) \; \; \; 400 (Chinese) & 40x10 (English) \; \; 10 (Chinese) & Varies  & Varies  & \url{https://catalog.ldc.upenn.edu/LDC2003T16} & \cite{radev2003evaluation}\\ \cdashline{1-9}
       TAC 2008 & English & News &  & 48x20 &   & 1 & \url{https://tac.nist.gov/data/index.html} & \cite{dang2008overview}\\ \cdashline{1-9}
       TAC 2009 & English & News &  & 44x20 &   & 1  & \url{https://tac.nist.gov/data/index.html} & \cite{tac2009}\\  \cdashline{1-9}     
       TAC 2010 & English & News &  & 46x20 &   & 1 & \url{https://tac.nist.gov/data/index.html} & \cite{tac2010}\\  \cdashline{1-9}     
       TAC 2011 & English & News &  & 44x20 &   & 1 & \url{https://tac.nist.gov/data/index.html} & \cite{tac2011}\\\cdashline{1-9}
       TeMário & Portuguese & News articles & 100 &  & 1 &  & \url{https://www.linguateca.pt/Repositorio/TeMario} & \cite{pardo2003temario}\\ \cdashline{1-9}
       TIPSTER SUMMAC & English & Electronic documents & 1,000 &  & 2 &  & \url{https://www-nlpir.nist.gov/related\_projects/tipster\_summac} & \cite{mani1999tipster}\\ \cdashline{1-9}
       TREC 2013 & English & News/Social Media &  & 4.5 Tb &  & Varies  & \url{https://trec.nist.gov/data.html} & \cite{yang2013bjut}\\ \cdashline{1-9}
       TREC 2014 & English & News/Social Media &  & 559 Gb &  & Varies  &  \url{https://trec.nist.gov/data.html} & \cite{zhao2014bjut}\\ \cdashline{1-9}
       TREC 2015 & English & News/Social Media &  & 38 Gb &  & Varies  & \url{https://trec.nist.gov/data.html}  & \cite{aliannejadi2015university}\\ \cdashline{1-9}
        USAToday-CNN & English & News &  & 121 &   &  & \url{https://github.com/nguyenlab/SocialContextSummarization} & \cite{nguyen2017exploiting}\\ \cdashline{1-9}
        VSoLSCSum & Vietnamese & News &  & 141 &   &  & \url{https://github.com/nguyenlab/VSoLSCSum-Dataset} & \cite{nguyen2016vsolscsum}\\  \cdashline{1-9}
       XSum & English & News & 226,711 &  & 1 &  & \url{https://www.tensorflow.org/datasets/catalog/xsum} & \cite{narayan2018don}\\ \cdashline{1-9}
       XLSum & Varies & News & 1,005,292 &  & 1 &  & \url{https://github.com/csebuetnlp/xl-sum} & \cite{hasan2021xl}\\ \cdashline{1-9}
       WikiHow & English & Instructions & 204,004 &  & 1 &  & \url{https://github.com/mahnazkoupaee/WikiHow-Dataset} & \cite{koupaee2018wikihow}
       \\\hline
    \end{tabular}
    }
\caption{A compilation with the main kinds of information that concern the most commonly utilized summarization datasets.}
\end{table}
\end{landscape}
\endgroup

\section{Basic topology of an ATS system}
\label{sec:general}

All ATS systems depend on a basic sequence of steps: {\it pre-processing}, {\it identification of the most important pieces of information} and {\it concatenation of the pieces of information for summary generation}.

While the pre-processing step may vary from one solution to the other, it usually contains some of the steps also very common in other applications of NLP \citep{Denny2018,Gentzkow2019}:

\begin{enumerate}
    \item Sentencization: The process of splitting the text into sentences.
    \item Tokenization: The process of removing undesired information from the text (such as commas, hyphens, periods, HTML tags, etc), standardizing terms (i.e. putting all words in lowercase and removing accents), and splitting the text so that it becomes a list of terms.
    \item Removal of stopwords: The process of removing the most common words in any language, such as articles, prepositions, pronouns, and conjunctions, that do not add much information to the text.
    \item Removal of low-frequency words: This is the process of removing rare words or misspelled words.   
    \item Stemming or Lemmatization: While stemming cuts off the end or beginning of the word, taking into account a list of common prefixes and suffixes that can be found in an inflected word, lemmatization takes the root of the word taking into consideration the morphological analysis of words. The idea of the application of one of these methods is to increase the word statistics. 
    
\end{enumerate}

The identification and selection of the most important pieces of information are two of the most important steps of an ATS system. The details of these steps depend on the used approach. It may depend on the attributes used to characterize the sentences (for instance, the frequency of the words), the method used to value the attributes, and the approach to avoid redundant sentences. We detail these steps in the next sections.

The concatenation step depends also on the used approach. In extractive ATS systems discussed in Section \ref{sec:extractive}, this step is simply a concatenation of the chosen sentences in the last step. In abstractive ATS systems explored in Section \ref{sec:abstractive}, we need a language model to rewrite the sentences that arise in the summary. Finally, In hybrid systems, presented in Section \ref{sec:compressive_extractive}, we usually revise the content of the extracted sentences. 

\section{Extractive summarization}
\label{sec:extractive}

Since the idea behind extractive summarization is to build a summary by joining important sentences of the original text, the two essential steps are (1) to find the important sentences and (2) to join the important sentences. In this section, we show that we can use different methods to implement these tasks. In Subsection \ref{sec:extractive_frequency_based}, we present the frequency-based methods. In Subsection \ref{sec:extractive_heuristics}, we present the heuristic-based methods. In Subsection \ref{sec:extractive_linguistic}, we present the linguistic-based methods. In Subsection \ref{sec:extractive_ml}, we present the methods based on supervised machine learning models. Finally, in Subsection \ref{sec:extractive_reinforcement}, we present the methods based on reinforcement learning approaches. 


\subsection{Frequency-based methods}
\label{sec:extractive_frequency_based}

We may use different models to implement an extractive frequency-based method. Thus, guided by the models used to implement the ATS systems, we split this section into five sections. In Subsection \ref{sec:extractive_vector_space} we present vector-space-based methods. In Subsection \ref{sec:extractive_matrix_factorization}, we present matrix factorization-based methods. In Subsection \ref{sec:extractive_graph}, we present the graph-based methods. In Subsection \ref{sec:extractive_topic}, we present the topic-based methods. Finally, in Section \ref{sec:extractive_embedding}, we present the neural word embedding-based methods.

\subsubsection{Vector-space-based methods}
\label{sec:extractive_vector_space}

The vector space model provides a numerical representation of sentences using vectors, facilitating the measurement of semantic similarity and relevance.
It is a model that represents each document of a collection of $N_S$ sentences by a vector of dimension $N_{V}$, where $N_V$ is the number of words (terms) in the vocabulary. The idea here is to use the vector space model to select the most relevant sentences of the document.  

In order to define precisely the vector space model, we start by defining the sentence-term matrix $\mathbf{M}_{\text{tfisf}}$. It is a $N_S \times N_{V}$ matrix that establishes a relation between a term and a sentence:
\begin{equation}
\mathbf{M}_{\text{tfisf}}=
\begin{array}{rl}
 ~ & ~~~~ w_1 ~~~~~~ w_2 ~~~~~~~~~~~~~ w_{N_V} \\
 \begin{array}{c} s_1 \\ s_2 \\ \vdots \\ s_{N_{S}} \end{array} & \begin{bmatrix} \omega_{1,1} & \omega_{1,2} & \cdots & \omega_{1, N_V} \\ \omega_{2,1} & \omega_{2,2} & \cdots & \omega_{2, N_V} \\ \vdots & \vdots & \cdots & \vdots \\ \omega_{N_{S},1} & \omega_{N_{S},2} & \cdots & \omega_{N_{S},N_V} \end{bmatrix} 
\end{array}
\label{eq:term_document}
\end{equation}
\noindent where each row is a sentence and each column is a term. The weight $\omega_{j,i}$ quantifies the importance of term $i$ in sentence $j$.
It depends on three factors. The first factor ({\it local factor})  relates to the term frequency and captures the significance of a term within a specific sentence. The second factor ({\it global factor}) relates to the sentence frequency and gauges the importance of a term throughout the entire document. The third factor (normalization) adjusts the weight to account for varying sentence lengths, ensuring comparability across sentences. 

 Thus, we may write the weight as
\begin{equation}
 \omega_{j,i} = \frac{\widetilde{\omega}_{j,i}}{\text{norm}_j},
 \label{eq:TFIDF_weight1}
\end{equation}

where 

\begin{equation}
 \widetilde{\omega}_{j,i} = \left\{
 \begin{array}{ll}
  f_{\text{tf}}(\text{tf}_{i,j}) \times f_{\text{isf}}(\text{sf}_{i}) & \text{if} ~ \text{tf}_{i,j} > 0 \\
  0 & \text{if} ~ \text{tf}_{i,j} = 0 \; 
 \end{array}
 \right..
 \label{eq:TFIDF_weight2}
\end{equation}

\noindent In Eq. (\ref{eq:TFIDF_weight1}), $\text{norm}_j$ is a sentence length normalization factor to compensate undesired effects of long sentences. In Eq. (\ref{eq:TFIDF_weight2}),  $f_{\text{tf}}(\text{tf}_{i,j})$ is the weight associated with the term frequency and $f_{\text{isf}}(\text{sf}_{i})$ is the weight associated with the sentence frequency. Table \ref{tab:TfIdfWeights}
 presents the most common choices for $f_{\text{tf}}$, $f_{\text{isf}}$ and $\text{norm}_j$ extracted from \cite{Baeza-Yates:2011:MIR:1796408}, \cite{Manning:2008:IIR:1394399} and \cite{Dumais1991}.
The term frequency (TF) and inverse sentence frequency (ISF) weighting scheme, called \index{Information!TF-ISF}TF-ISF, are the most popular weights in information retrieval. 
\begin{table}[H]
\begin{center}

{\small
\begin{tabular}{ll} 
 \hline
\textbf{Term frequency} & $f_{\text{tf}}(\text{tf}_{i,j})$ \\
\hline
Binary & $\displaystyle \min{\{\text{tf}_{i,j}, 1\}} $ \\[4mm] 
Natural (raw frequency) & $\displaystyle \text{tf}_{i,j}$ \\[4mm]
Augmented & $\displaystyle 0.5 + 0.5 \frac{\text{tf}_{i,j}}{\max_{i'}{\text{tf}_{i',j}}}$ \\[4mm]
Logarithm & $\displaystyle 1 + \log_2{(\text{tf}_{i,j})}$ \\[4mm]
Log average & $\displaystyle \frac{ 1 + \log_2{(\text{tf}_{i,j})} }{ 1 + \log_2{(\text{avg}_{w_{i'} \in d_j}\text{tf}_{i',j})} }$ \\[4mm]
 ~ & ~ \\
\hline
\textbf{Sentence frequency} & $f_{\text{isf}}(\text{df}_{i})$ \\
\hline
  None & $1$ \\[4mm]
  Inverse frequency & $\displaystyle \log_2{\left( \frac{N_S}{\text{sf}_{i}} \right)}$ \\[4mm]
  Entropy & $\displaystyle 1 - \sum_{j}{ \frac{p_{i,j} \log{(p_{i,j})}}{\log{(N_S)}} }$ \\[2mm]
  ~       & $\displaystyle p_{i,j} = \frac{\text{tf}_{i,j}}{\sum_{j}{\text{tf}_{i,j}}} $ \\[4mm]
\hline
\textbf{Normalization} & $\text{norm}_j$ \\
\hline
  None & $1$ \\[4mm]
  Cosine & $\displaystyle \sqrt{\sum_{i}^{N_V}{\widetilde{\omega}_{i,j}^2}}$ \\[4mm]
  Word count & $\displaystyle \sum_{i}^{N_V}{\text{tf}_{i,j}} $ \\[4mm]
\hline
\end{tabular}
}
\end{center}
\caption{The most common variants of TF-ISF weights.}
\label{tab:TfIdfWeights}

\end{table}

Before presenting the methods, it is worth mentioning the study presented by \cite{nenkova2006compositional} that stresses the roles of three different important dimensions that arise in frequency-based attempts to summarization, namely (1) word frequency, (2) composition functions for estimating sentence importance from word frequency estimates, and (3) adjustment of frequency weights based on context:

\begin{enumerate}
    \item Word frequencies (or some variation based on TF-ISF) are the starting points to identify the keywords in any document (or clusters of documents); 
    \item The composition function is necessary to estimate the importance of the sentences as a function of the importance of the words that appear in the sentence.
    \item The adjustment of frequency weights based on context is fundamental since the notion of importance is not static. A sentence with important keywords must be included in the summary as long as there is no other very related sentence with similar keywords in the summary.

\end{enumerate}

Therefore, Eqs. (\ref{eq:TFIDF_weight1}) and (\ref{eq:TFIDF_weight2}) with Table \ref{tab:TfIdfWeights} present different options to select the most important terms in the complete text. In order to identify the most relevant sentences of the text, as above-mentioned, we need to aggregate these measures of importance associated with the terms and also avoid the chosen sentences having the same terms. A simple way to aggregate these measures for each term is to evaluate the average of each term in a sentence. However, in order to avoid the repetition of terms in different selected sentences, after selecting a given sentence, we may penalize the choice of sentences with terms that arise in sentences that had been previously selected. 

The SumBasic method \citep{Nenkova2005} uses the raw document frequency to identify the most important terms. The relevance of each sentence is given by the average of its terms. The idea is to select the most important sentences according to that criterion. However, in order to avoid the selection of highly correlated sentences, after selecting a given sentence, the frequencies of all the terms that arise in the selected sentence are squared and with these new values, the relevance of the sentences is re-evaluated. We may find an extension of \cite{Nenkova2005} in \cite{nenkova2006compositional}. In this paper, using also the frequency of the terms as an input, the authors consider different possibilities for the evaluation of the score associated with each sentence such as (1) multiplication of the frequency of the sentence's terms; (2) addition of these frequencies and the division by the number of terms in the sentence; (3) addition of these frequencies. Note that while (1) favors short sentences, (3) favors longer sentences, and (2) is a combination of both. As in SumBasic, they also consider an additional step in the algorithm to reduce redundancy. After a sentence has been selected for inclusion, the frequencies of the terms for the words in the selected sentences are reduced to 0.0001 (a number close to zero) to discourage sentences with similar information from being chosen again. 

SumBasic+ due to \cite{darling2010multi} is also a direct extension of the work of \cite{nenkova2006compositional}, in which a linear combination of the unigram and bigram frequencies is explored. They choose the parameters of the linear combination in order to maximize the ROUGE score. This work also explores query-based summarization and update-based summarization. While the setup used for update-based summarization is essentially the same, in order to attend to the task of query-based summarization, the authors suggest adding more probability mass on terms that arise in the query vector.

Using the same principle described in SumBasic \citep{Nenkova2005}, we may use any weight given by the combination of the terms in Table \ref{tab:TfIdfWeights} to identify the most relevant terms and sentences and consider different methods to avoid redundancy. For instance, in order to avoid the selection of highly correlated sentences, \cite{Carbonell1998} suggest that we may evaluate the relevance of each sentence by the convex combination between the relevance of the sentence given by TF-ISF terms and the maximal correlation of the sentence and the sentences already included in the summary.

An interesting way to reduce the redundancy of sentences in the final summary is to separate similar sentences into clusters. A simple way to do that is to characterize the sentences with TF-ISF vectors, use a clustering method to split the document into groups of similar sentences, and choose the most relevant sentence of each group as the one that is the closest to the centroid of each cluster \citep{Zhang2009}. These sentences are the candidate sentences to be included in the summary. We select these sentences in order of relevance based on TF-ISF.

Another interesting approach called KL Sum is to choose sentences that minimize the Kullback–Leibler divergence (relative entropy) between the frequency of words in the summary and the frequency of words in the text \citep{haghighi-vanderwende-2009-exploring}, where the sentences are greedily chosen. 

A very interesting multi-document extractive approach is the submodular approach due to \cite{lin2011class}. They formulate the problem as an optimization problem using monotone nondecreasing submodular set functions. A submodular function $f$ on a set of sentences $\mathcal{S}$ satisfies the following property: for any $A \subset B \subset \mathcal{S}\backslash {s} $, we have $f (A + {s}) - f (A) \ge f(B + {s}) - f (B)$, where $s \in \mathcal{S}$. Note that $f$ satisfies the so-called diminishing returns property and it captures the intuition that adding a sentence to a small set of sentences, like the summary, makes a greater contribution than adding a sentence to a larger set. The objective is then to find a summary that maximizes the diversity of the sentences and the coverage of the input text. The authors formulate this problem as the problem of maximizing the objective function given by $F(S) = L(S) + \lambda R(S)$, where $S$ is the summary, $L(S)$ measures the coverage of summary set $S$ to the document, $R(S)$ measures (rewards) diversity in $S$, and $\lambda \ge 0$ is a trade-off between coverage and diversity. The authors also call attention to the fact that $L(S)$ should be monotonic, as coverage improves with a larger summary, and it should also be submodular since the effect of adding a new sentence to a smaller summary has a large effect. On the other hand, assuming that the sentences were previously split into clusters $P_i$ for $i=1,\cdots,K$, in order to reward diversity, they set $R(S)=\sum_{k=1}^{K} g(\sum_{j \in P_i\cap S}r_j)$, where $g$ is a concave function and $r$ is the sentence individual reward. The authors emphasize the fact that $R$ is also submodular. With this objective, they show that an approximate greedy algorithm can be used for the task. In order to deal with a query-based task, they change the function $R$ to be a linear combination of the reward associated with the individual sentences and a reward associated with the relevance of the sentence to the query. In this context, it is interesting to mention a value overview presented in \cite{bilmes2022submodularity} of the use of submodularity in machine learning and artificial intelligence.   

There are many methods for ATS using the vector space model. We present a representative compilation of these methods in Table \ref{tab:digest_vector_space_models}.

\begingroup
\renewcommand{\arraystretch}{1.2}
\begin{landscape}
\begin{table}[t]
    \centering
    \scriptsize
    \begin{tabular}{lp{10cm}p{4cm}p{3cm}}
     Source     & Main contribution & Dataset & Evaluation \\\hline
    \cite{Carbonell1998}  & It uses the TF-ISF to select the sentences and in an iterative fashion, it uses a convex combination of the TF-ISF and the correlation with the previously chosen sentences. & TIPSTER topic \citep{miller1998algorithms}& F-scores of the extracted sentences.  \\\cdashline{1-4}
    \cite{radev2004centroid} & In a multi-document setup, it creates clusters of documents by topics, it represents both sentences and topics using TF-IDF and it chooses the sentences that should be extracted based on scores that weights the proximity of the sentence to the topics using cosine similarity. & A corpus consisting of a total of 558 sentences in 27 documents, organized in 6 clusters extracted by CIDR \citep{radev1999description}.& Human experts\\\cdashline{1-4} 
    \cite{Nenkova2005} & It uses the raw document frequency to determine the relevance of each sentence and in an iterative fashion it penalizes the sentences with words of previously chosen sentences. & DUC 2004, 2005 &  ROUGE-1, ROUGE-2, ROUGE-SU-4, manual Pyramid and repetition.                            \\\cdashline{1-4}
    \cite{nenkova2006compositional} & It is an extension of \cite{Nenkova2005} that considers different options to evaluate the score of a sentence.  & DUC 2004, 2005 & ROUGE-1, ROUGE-2, ROUGE-SU-4, manual Pyramid \\\cdashline{1-4}
    \cite{mcdonald2007study} & It formulates the problem of multi-document summarization as a very general optimization problem where the objective is to choose parts of a text (for instance, sentences) that maximize a given score that increases with the relevance of the parts of the text and decreases the redundancy. It solves both with a greedy algorithm and a dynamic programming approach based on a solution to the 0-1 knapsack problem \citep{cormen2022introduction}. & DUC 2002 & ROUGE-1 and ROUGE-2 \\\cdashline{1-4} 
    \cite{gillick2008icsi} & It evaluates the importance of the sentences in an integer programming framework whose objective is to build a summary that maximizes the concept coverage (bigram frequency). & TAC 2008 & ROUGE-1, ROUGE-2 and ROUGE-SU-4\\\cdashline{1-4}
    \cite{Zhang2009}  & It uses the TF-ISF to characterize the sentences, forms clusters of sentences using this information, and chooses the most relevant sentences of these clusters. & DUC 2003 & ROUGE-1, ROUGE-2 and F-1 score of the extracted sentences. \\\cdashline{1-4}
    \cite{haghighi-vanderwende-2009-exploring}  & It chooses sentences that minimize the Kullback–Leibler divergence between the frequency of words in the summary and the frequency of words in the text. & DUC 2006 & ROUGE-1, ROUGE-2 and ROUGE-SU-4 \\\cdashline{1-4}
    \cite{darling2010multi} & It is an extension of \cite{nenkova2006compositional}. It explores a linear combination of the unigram  and bigram frequencies and it chooses the parameters of the linear combination in order to maximize the ROUGE score. In order to attend to the task of query-based summarization, it adds more probability mass on terms that arise in the query vector. & DUC 2004 and TAC 2010& ROUGE-2,  ROUGE-SU-4, basic elements, linguistic quality and manual Pyramid \\\cdashline{1-4} 
    \cite{lin2011class} & It sets the problem of multi-document extractive summarization as a greedy optimization of a submodular function that trades off between coverage and diversity.  & DUC 2003, DUC 2004, DUC 2005, DUC 2006 and DUC 2007 & ROUGE-1 and ROUGE-2  
    \\\hline
    \end{tabular}
    \caption{A representative compilation of the ATS methods that use vector space models.}
    \label{tab:digest_vector_space_models}
\end{table}
\end{landscape}
\endgroup

\subsubsection{Matrix factorization based methods}
\label{sec:extractive_matrix_factorization}

The idea of the matrix factorization methods of extractive summarization is to decompose the sentence-term matrix presented in Section \ref{sec:extractive_vector_space} into a dense representation, where each term is represented by a feature (or concept). The point is that many different terms present very similar concepts. So, instead of dealing individually with the terms, we may deal directly with the concepts. In particular, in the case of ATS, we can select sentences that are good representations of different concepts that arise in the text.  
 
The starting point for this kind of method is the vector space model reviewed in Section \ref{sec:extractive_vector_space}.  Suppose that we have a text that we want to summarize with $N_S$ sentences. We may represent this text by the sentence-term matrix $\mathbf{M}_{\text{}}^{S}$  in Eq. (\ref{eq:term_document}) with $N_S$ rows and $N_V$ columns. 

Using for instance, Singular Value Decomposition (SVD) \citep{stewart1993early}, \cite{Golub2013} decompose this matrix in the following way:
\begin{equation}
 \mathbf{M}_{\text{tfisf}}^{S} = \mathbf{U} \boldsymbol{\Sigma} \mathbf{V}^T
 \label{eq:LatentSemanticIndexModel}
\end{equation}
\noindent where
$\mathbf{U}$ and $\mathbf{V}^T$ are respectively orthogonal matrices of eigenvectors derived from sentence-sentence and term-term covariance matrices\footnote{This means that the columns of $\mathbf{U}$ are eigenvectors of $\mathbf{M}_{\text{tfisf}}\mathbf{M}_{\text{tfisf}}^T$ and the columns of $\mathbf{V}$ are eigenvectors of $\mathbf{M}_{\text{tfisf}}^T\mathbf{M}_{\text{tfisf}}$.},
and $\boldsymbol{\Sigma}$ is an $r \times r$ diagonal matrix of singular values where $r = \min{(N_S, N_{V})}$ is the rank of $\mathbf{M}_{\text{tfisf}}$. Note that, in this representation, the rows of the matrix $\mathbf{U} \boldsymbol{\Sigma}$ contain the $r$-dimensional representation of the $N_S$ sentences, where each column of $\mathbf{V}$ is a base vector where each sentence is represented. Therefore, each column of $\mathbf{U} \boldsymbol{\Sigma}$ is associated with a concept. We may find a tutorial introduction to SVD in \cite{klema1980singular} and a survey of SVD for intelligent information retrieval in \cite{berry1995using}.

If we want that the summary finds the most important sentences of each concept, the basic idea is to select the sentences that, for each column, have the maximal absolute entry as described in \cite{Gong2001}. 

\cite{steinberger2004using} introduce an interesting modification to \cite{Gong2001}'s method. It calls our attention that the latter presents two significant disadvantages: First, it is necessary to use the same number of dimensions as the number of sentences we want to choose for a summary. However, we know that the higher the number of dimensions of the concept space, less significant topics are introduced in the summary. Second, the sentences that have a higher entry in a given concept are chosen, but they are not necessarily the most important sentences (since some of the concepts may not be that important). Therefore, the idea here is to extract the most relevant sentences $s$ in terms of the weights $\sqrt{\sum_{k=1}^{r}u_{sk}^{2} \sigma_{k}^{2}}$. In order to deal with a multi-document update summarization task, \cite{steinberger2009update} create sets of topics for both previous documents and new documents. Sentences containing novel and significant topics may be extracted for building the update. Novelty is measured by the average of the internal product between the topics
of the previously known documents and the topics of the new documents.

Other interesting approaches use the Non-Negative Matrix Factorization technique (NMF) \citep{paatero1994positive,paatero1997least,lee1999learning,lee2000algorithms,Gillis2020}. The idea of NMF is to decompose $\mathbf{M}_{\text{tfisf}}^{S}$ in a product of other two matrices $W$ and $H$, where we may interpret the columns of the product matrix as linear combinations of the column vectors in $W$ using the coefficients provided by the columns of $H$. In general, we assume that the number of columns of $W$ (or the number of rows of $H$) is lower than those of the product matrix we are decomposing. One simple algorithm to find this decomposition is based on the non-negative least squares, where we minimize the distance using the Frobenius norm between the product matrix and the actual matrices $W$ and $H$. We may find a comprehensive review of the NMF including properties and algorithms in \cite{wang2012nonnegative}. Thus, we may extract the sentences that maximize each of the topics of the documents that are the ones that for each column has the maximal absolute entry as in \cite{Gong2001}. This is the algorithm considered in \cite{lee2009automatic}.

In order to deal with a query-based task, \cite{park2006query} provide an algorithm that follows the same steps of \cite{lee2009automatic} and extracts the sentences that maximize the topics of the document that have higher similarity with the provided query. 

There are many methods for ATS using matrix factorization representations. We present a representative compilation of these methods in Table \ref{tab:digest_latent}. Furthermore, although we have considered here only the two common methods used in summarization, it is worth knowing that there are many other kinds of matrix factorization techniques. We may find a survey of these techniques in \cite{Lyche2020} and \cite{edelman2021fifty}. 

\begingroup
\renewcommand{\arraystretch}{1.2}
\begin{landscape}
\begin{table}[t]
    \centering
    \scriptsize

    \begin{tabular}{lp{10cm}p{4cm}p{3cm}}
     Source    & Main contribution & Dataset & Evaluation \\\hline
    \cite{Gong2001}           &  It applies the SVD decomposition to the TF-ISF representation of the text and it selects the sentences that are the best representative of each concept. & Two months of the CNN Worldview news programs & Precision, recall, and F1 score of the extracted sentences. \\\cdashline{1-4}
    \cite{steinberger2004using} & It applies the SVD decomposition to the TF-ISF representation of the text and it selects the sentences with large weights that jointly represent the importance of the concept and the importance of the sentence as a representative of that concept. & Reuters collection & Cosine similarity and latent-semantic\\\cdashline{1-4}
    \cite{park2006query} & It uses NMF to select the most relevant sentences of the topics that are similar to the given query. & Yahoo Korea News & Precision \\\cdashline{1-4} 
    \cite{steinberger2009update} & It uses latent semantic analysis for creating sets of topics for both previous documents and new documents and it selects sentences containing novel and significant topics. & DUC 2007 and TAC 2008 & Pyramid, ROUGE-2, ROUGE-SU-4, Basic elements and human experts \\\cdashline{1-4}
    \cite{lee2009automatic} & It applies the NMF decomposition to the TF-ISF representation of the text and it selects the sentences that are the best representative of each concept. & DUC 2006 & ROUGE-1, ROUGE-L, ROUGE-W and ROUGE-SU-2 \\\cdashline{1-4}
    \cite{yogatama2015extractive} & It provides a greedy algorithm to create a summary that maximizes the volume (coverage) of selected sentences in the semantic space, where the sentences are represented in the semantic space by the SUV decomposition of the bigrams that form the sentences.& TAC 2008 and TAC 2008 & ROUGE-1 and ROUGE-2 \\\cdashline{1-4}
    \cite{nguyen2019web} & This is an approach to social context summarization, wherein the mathematical formulation of the NMF, the web documents, and the user's content share the same topic matrix. & SoLSCSum, USAToday-CNN, VSoLSCSum and DUC 2004 & ROUGE-1, ROUGE-2 and ROUGE-W\\\cdashline{1-4}
    \cite{khurana2022investigating} & It employs NMF to reveal probability distributions for computing entropy of terms, topics, and sentences in latent space. It uses the classical Knapsack optimization algorithm to select entropic highly informative sentences. & DUC 2001, DUC 2002, CNN/DailyMail & ROUGE-1, ROUGE-2 and ROUGE-L \\\hline
    \end{tabular}
    \caption{A representative compilation of the ATS methods that use matrix factorization methods.}
    \label{tab:digest_latent}
\end{table}
\end{landscape}
\endgroup


\subsubsection{Graph based methods}
\label{sec:extractive_graph}

A graph (or a network) is a pair $G = (V, E)$, where $V$ is a set where each element is called a vertex (node) and $E$ is a set of paired vertices, where each element is called an edge. Networks are used worldwide to model systems whose components interact with each other, such as genetic systems (gene coexpression or gene regulatory systems), protein networks, reaction networks, anatomic networks (intercellular and brain networks), ecological networks, technological networks (electric power networks) and social networks \citep{newman2003structure,estrada2012structure}. Over the years, several metrics were introduced to characterize the networks and also the components of these systems \citep{costa2007characterization}. Among them, we may cite centrality, which we use in this section, that measures the importance of each component in the network. For instance, if we consider Instagram, the social media platform, the most important individuals are the ones that have the largest number of followers. 

In the graph-based methods approach, we represent the text as a network. In this network, each sentence is a node of the network. There is a link between two nodes if the sentences are similar. Although we may think in different ways to weigh the links of these networks, simple ideas are: (1) A link exists when two sentences share a word; (2) A link exists when the similarity between two sentences exceed a given threshold. 

With these definitions in mind, the most relevant sentences are those with the highest centrality and are the ones that should be included in the summary. This is the basic idea of Text Rank \citep{mihalcea2004textrank} and Lex Rank \citep{Erkan2004} that use respectively the page rank \citep{page1999pagerank} and the usual eigenvector approach \citep{bonacich1972factoring,RUHNAU2000} to evaluate the centrality of the sentences in a multi-document setup.

In order to deal with a query-based approach, \cite{otterbacher2005using} basically uses the approach of the Lex Rank method \citep{Erkan2004}. However, instead of selecting the sentences with the highest centrality, they select sentences based on a mixture model that also considers the relevance of the sentences according to the provided query. 

In \cite{wan2006improved}, in a multi-document setup, the sentence similarities are built using the cosine similarity of the TF-IDF vectors of the sentences in an approach that differentiates sentences of the same document from sentences of different documents. These similarities are normalized to define a Markov chain in the network and, based on it, they evaluate the centrality of each sentence (node). In order to choose the sentences to be extracted, they penalize sentences that are very connected with the previously selected sentences. 

There are many graph methods for ATS. We present a significant compilation of these methods in Table \ref{tab:digest_graph_methods}.

\begingroup
\renewcommand{\arraystretch}{1.1}
\begin{landscape}
\begin{table}[t]
    \centering
    \scriptsize
    \begin{tabular}{lp{12cm}p{4cm}p{3cm}}
     Source              & Main contribution & Dataset & Evaluation \\\hline
    \cite{mihalcea2004textrank}  &  It represents the document as a network, where each sentence is a node of the network and there is a link between two nodes if the sentences share a word. It weights the edges by the normalized (by the length of the sentence) number of shared words. It uses the Page Rank to identify the most central sentences. & Inspec database \citep{hulth2003improved} & Precision, recall and F-measure. \\\cdashline{1-4}
    \cite{Erkan2004}  & It uses the same network representation of \cite{mihalcea2004textrank}. It uses the eigenvalue centrality to identify the most central sentences. & DUC 2003, 2004 & ROUGE-1\\ \cdashline{1-4}
    \cite{mihalcea2005language} & It uses the same network representation of \cite{mihalcea2004textrank}. It weights the edges considering three different situations: (a) a simple undirected graph; (b) a directed weighted graph with the orientation of edges set from a sentence to sentences that follow in the text (directed forward), or (c) a directed weighted graph with the orientation of edges set from a sentence to previous sentences in the text (directed backward). It uses the HITS and Page Rank algorithms to identify the most central sentences. & DUC 2002 and TeMário & ROUGE-1\\\cdashline{1-4}
\cite{otterbacher2005using} & It is a query-based approach. It uses the Lex Rank method \citep{Erkan2004} to find out the most relevant sentences and it selects the sentences based on a mixture model that also considers the relevance of the sentences according to the provided query. & A corpus of 20 multi-document clusters of complex news stories & Mean Reciprocal Rank (MRR)
and Total Reciprocal Document Rank (TRDR)
 \\\cdashline{1-4}    
    \cite{wan2006improved} & In a multi-document approach, it uses the same network representation of \cite{mihalcea2004textrank}. It evaluates the sentence similarities using the cosine similarity of the TF-IDF vectors of the sentences. It normalizes the similarities to define a Markov chain and to evaluate the centrality of each sentence (node). In order to choose the sentences to be extracted, it penalizes sentences that are very connected with the previously selected sentences. & DUC 2002 and DUC 2004 & ROUGE-1  \\ \cdashline{1-4}
    \cite{lin2009graph}  & It uses the same network representation of \cite{mihalcea2004textrank} with edge weights given by the similarity
between the sentences. It evaluates the similarity using the cosine between the TF-IDF vectors of the sentences or the ROUGE-1 (F measure) score. In order to extract the sentences, it maximizes a submodular set function defined on the graph using a greedy algorithm. & ICSI meeting corpus \citep{janin2003icsi} & ROUGE-1 and F-measure.\\\cdashline{1-4}
    \cite{thakkar2010graph} & It uses the same network representation of \cite{mihalcea2004textrank}. It associates each edge with a cost that is proportional to the physical distance of the sentences and inversely proportional to the similarity between the sentences and the similarity between the sentence and the title of the document. It creates the summary by taking the shortest path that starts with the first sentence of the original text and ends with the last sentence. & -- & -- \\\cdashline{1-4}
    \cite{barrios2016variations}   & It presents new alternatives to the similarity function for the Text Rank algorithm \citep{mihalcea2004textrank}& DUC 2002 & ROUGE-1, ROUGE-2 and ROUGE-SU-4.
    \\\cdashline{1-4}
\cite{mallick2019graph} & It uses the same network representation of \cite{mihalcea2004textrank} with the edge weights given by a  modified cosine similarity. It uses a modified Page Rank algorithm to score the sentences. & BBC news articles & ROUGE-1, ROUGE-2 and ROUGE-L. \\\cdashline{1-4}
    \cite{van2019query} & This is a query-based approach where it represents sentences as nodes, as usual. However, it adds hyperedges between the sentences that consider the similarity of the themes of each sentence, where the themes of the sentences are determined by a clustering algorithm. It extracts the sentences that cover all themes of the corpus. & DUC 2005, DUC 2006 and DUC 2007 & ROUGE-2 and ROUGE-SU-4.    \\\cdashline{1-4} 
\cite{ucckan2020extractive} &  This is a multi-document summarization method. It uses the same network representation of \cite{mihalcea2004textrank}. It removes the maximum independent set from the original graph and it selects the sentences in the modified graph as the ones with the highest eigenvalue centralities. & DUC 2002 and DUC 2004 & ROUGE-1, ROUGE-2, ROUGE-L and  ROUGE-W. 
    
    \\\hline
    \end{tabular}
    \caption{A significant compilation of the graph methods used for ATS.}
    \label{tab:digest_graph_methods}
\end{table}
\end{landscape}
\endgroup

\subsubsection{Topic-based methods}
\label{sec:extractive_topic}

The topic-based summarization methods rely on topic representations such as the Latent Dirichlet Allocation (LDA) \citep{blei2003latent}. LDA is a generative model\footnote{A generative model is a model that describes the distribution of the data and tells how likely a given example is.} that represents a document by a collection of topics and each topic, in its turn, by a collection of words. 

In order to develop the so-called TopicSum, \cite{haghighi-vanderwende-2009-exploring} assume a fixed vocabulary $V$ and propose a LDA-like generative model. For the sake of organization, although this approach was originally developed for multi-document summarization, we present here this approach in a setup for single-document summarization:

\begin{enumerate}
    \item Draw a ``background" vocabulary distribution $\phi_B$ from $\mathrm{Dirichlet}(V ,\lambda_B )$ shared across the document collection representing the background distribution over vocabulary words.
    \item For each document $d$, we draw a ``content" distribution $\phi_C$ from $\mathrm{Dirichlet}(V ,\lambda_C )$ representing the significant content of $d$ that we wish to summarize. 
    \item For each sentence $s$ of each document $d$, draw a distribution $\psi_T$ over topics (content, background) from a Dirichlet prior with pseudo-counts ($n_C$,  $n_B$), where $n_C<n_B$ reflects the intuition that most of the words in a document come from the background. 
\end{enumerate}

Using this generative model, the authors \cite{haghighi-vanderwende-2009-exploring} estimate $\phi_C$ for each document and select the most important sentences using the same criterion used by the KLSum discussed in Section \ref{sec:extractive_frequency_based}, replacing the frequency of the words in the text, which is a unigram distribution, by $\phi_C$. An extension of this model, called HIERSum and provided by the same work, considers that a document may be formed by different topics as in \cite{chang2009latent}. 

There are other ideas very similar to the approach proposed by \cite{haghighi-vanderwende-2009-exploring} that we present in Table \ref{tab:digest_topic_model}.

\begingroup
\renewcommand{\arraystretch}{1.2}
\begin{landscape}
\begin{table}[t]
    \centering
    \scriptsize
    \begin{tabular}{lp{10cm}p{2cm}p{3cm}}
     Source     & Main contribution & Dataset & Evaluation \\\hline
    \cite{haghighi-vanderwende-2009-exploring}  & It presents an approach for multi-document summarization that chooses sentences that minimize the Kullback–Leibler divergence between the frequency of words in the summary and the frequency of content estimated by an LDA-like approach. It also extends this model to consider the possibility that a document is formed by many topics. & DUC 2006 & ROUGE-1, ROUGE-2 and ROUGE-SU-4\\\cdashline{1-4}
    \cite{chang2009latent} & It explores two variations of the LDA-like approach for extractive summarization. &  DUC 2005 & ROUGE-1, ROUGE-2 and ROUGE-L\\\cdashline{1-4}
    \cite{wang2009multi} & In a multi-document summarization approach, it proposes a unigram model as a mixture of several topic unigram models and, in its turn, it assumes that topic unigram models are mixtures of sentences unigram models. Thus, each topic is represented by a set of sentences and the sentences to be extracted are the most representative of each topic. & DUC 2002 and DUC 2004. &  ROUGE-1, ROUGE-2, ROUGE-L and ROUGE-SU-4\\\cdashline{1-4}
    \cite{delort2012dualsum} & It presents a variation of LDA that aims to learn to distinguish between common information and novel information. & TAC 2008 and TAC 2009 & ROUGE-1, ROUGE-2 and ROUGE-SU-4 \\\cdashline{1-4}      
    \cite{belwal2021text} & It mixes ingredients of topic modeling using LDA with vector space models representation. It selects sentences represented by the vector space models that are the most similar to the topics previously selected. & CNN/DailyMail &  ROUGE-1, ROUGE-2, ROUGE-L and ROUGE-SU-4 \\\cdashline{1-4}
    \cite{srivastava2022topic} & It uses LDA for topic modeling and the $K$-medoids clustering method for summary generation. & Wikihow, CNN/DailyMail and DUC 2002 &  ROUGE-1, ROUGE-2 and ROUGE-L \\
    \hline
    \end{tabular}
    \caption{A representative compilation of the ATS methods that use topic-based methods.}
    \label{tab:digest_topic_model}
\end{table}
\end{landscape}
\endgroup

\subsubsection{Neural word embedding based methods}
\label{sec:extractive_embedding}

Word embeddings represent words as vectors in a high-dimensional space, capturing their semantic meaning \citep{bengio2000neural}. While the vectors aggregating words by concepts discussed in Section \ref{sec:extractive_matrix_factorization} could be seen as a form of word embeddings, traditional definitions align more closely with representations derived from neural network models that extend classical models of language.

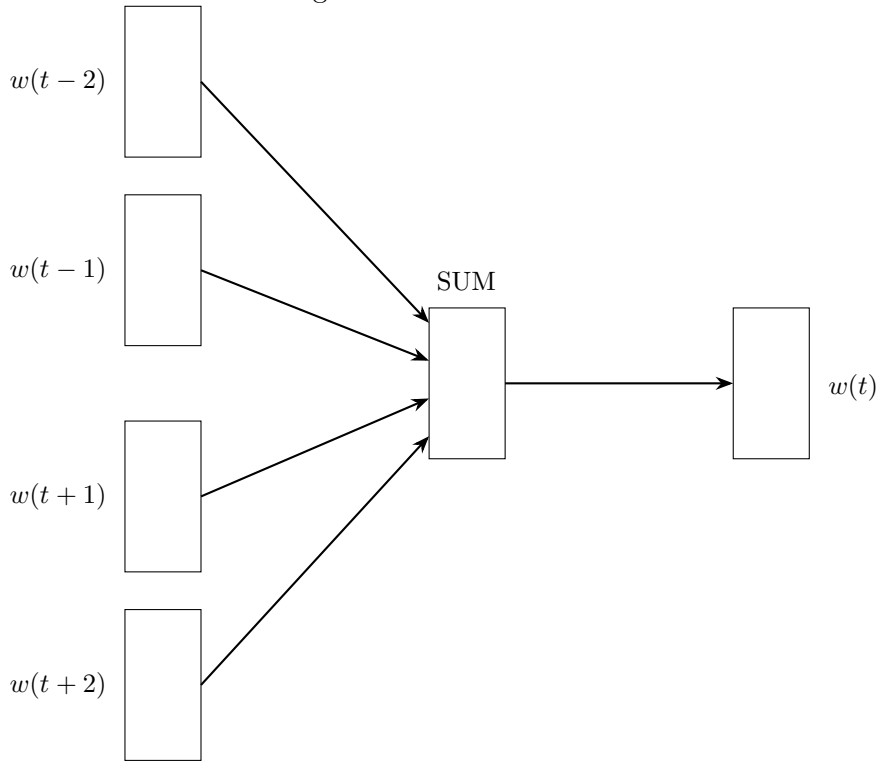
\begin{figure}[htb]
\begin{center}
\caption{CBOW model.}
\label{fig:cbow}
\begin{tikzpicture}

\draw [fill=none,draw=black] (0,8) rectangle ++(1,2); 
\draw [fill=none,draw=black] (0,5.5) rectangle ++(1,2); 
\draw [fill=none,draw=black] (0,2.5) rectangle ++(1,2); 
\draw [fill=none,draw=black] (0,0) rectangle ++(1,2); 
\draw [fill=none,draw=black] (4,4) rectangle ++(1,2); 
\draw [fill=none,draw=black] (8,4) rectangle ++(1,2); 

\draw [-Stealth, thick](1,9) -- (4,5.8); 
\draw [-Stealth, thick](1,6.5) -- (4,5.3); 
\draw [-Stealth, thick](1,3.5) -- (4,4.8); 
\draw [-Stealth, thick](1,1) -- (4,4.3); 
\draw [-Stealth, thick](5,5) -- (8,5); 

\node [xshift=-25] at (0,9) {\small $w(t-2)$}; 
\node [xshift=-25] at (0,6.5) {\small $w(t-1)$}; 
\node [xshift=-25] at (0,3.5) {\small $w(t+1)$}; 
\node [xshift=-25] at (0,1) {\small $w(t+2)$}; 
\node [xshift=14,yshift=67] at (4,4) {\small SUM}; 
\node [xshift=45,yshift=27] at (8,4) {\small $w(t)$}; 

\end{tikzpicture}

\end{center}
\end{figure}

\begin{figure}[htb]
\begin{center}
\caption{Skip-gram model.}
\label{fig:skip-gram}
\begin{tikzpicture}

\begin{scope}[xscale=-1]
\draw [fill=none,draw=black] (0,8) rectangle ++(1,2); 
\draw [fill=none,draw=black] (0,5.5) rectangle ++(1,2); 
\draw [fill=none,draw=black] (0,2.5) rectangle ++(1,2); 
\draw [fill=none,draw=black] (0,0) rectangle ++(1,2); 
\draw [fill=none,draw=black] (4,4) rectangle ++(1,2); 
\draw [fill=none,draw=black] (8,4) rectangle ++(1,2); 

\draw [Stealth-, thick](1,9) -- (4,5.8); 
\draw [Stealth-, thick](1,6.5) -- (4,5.3); 
\draw [Stealth-, thick](1,3.5) -- (4,4.8); 
\draw [Stealth-, thick](1,1) -- (4,4.3); 
\draw [Stealth-, thick](5,5) -- (8,5); 
\end{scope}

\node [xshift=25] at (0,9) {\small $w(t-2)$}; 
\node [xshift=25] at (0,6.5) {\small $w(t-1)$}; 
\node [xshift=25] at (0,3.5) {\small $w(t+1)$}; 
\node [xshift=25] at (0,1) {\small $w(t+2)$}; 
\node [xshift=-44,yshift=27] at (-8,4) {\small $w(t)$}; 

\end{tikzpicture}

\end{center}
\end{figure}
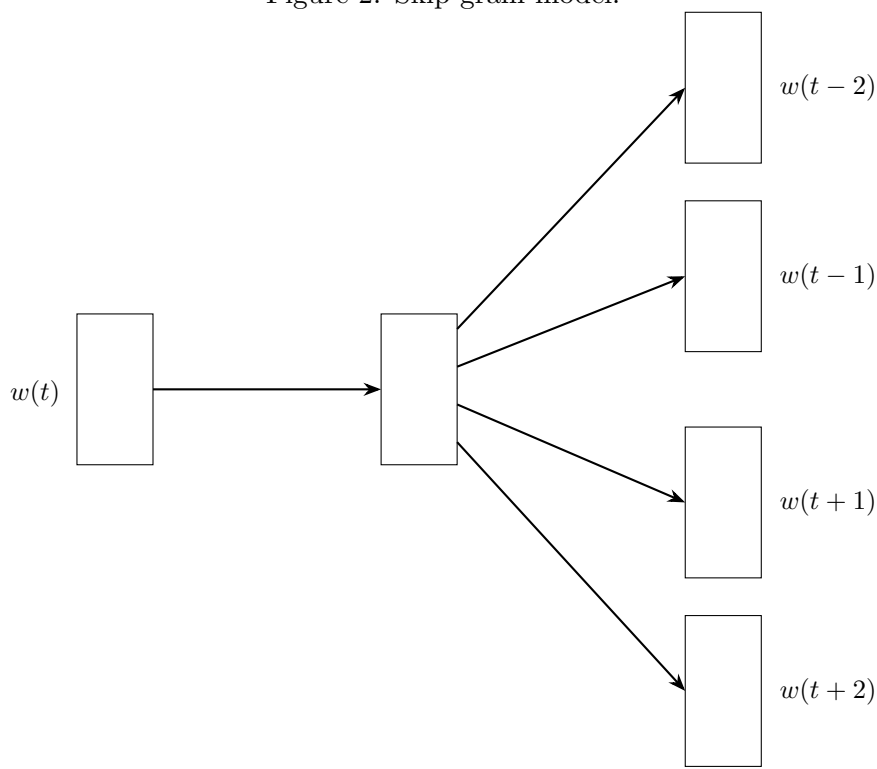

Numerous methods have been proposed for generating word embeddings. Among the most notable are Word2vec \citep{Mikolov2013}, GloVe \citep{Pennington2014}, and fastText \citep{joulin2016}. A comprehensive review of these methods can be found in \cite{gutierrez2018systematic}. Although these models are all rooted in neural probabilistic language frameworks and are trained semi-supervised\footnote{This means that we build inputs and outputs for the neural networks using all the sentences and all the words in these sentences, that are available in the training documents without the need for annotated texts.}, they differ in performance metrics, input/output types, and their balance between local and global word information.
 While some of these models use performance indexes that are variations of the likelihood functions associated with multinomial logit models (a kind of cross-correlation entropy), others use variations of the mean square error. 
 For instance, the CBOW, presented in Figure \ref{fig:cbow}, and skip-gram, presented in Figure \ref{fig:skip-gram},  models, both introduced by \cite{Mikolov2013}, approach the problem differently. Using a sentence like ``You get a shiver in the dark" from the song ``Sultans of Swing" by Dire Straits, CBOW tries to predict the word "shiver" from its surrounding words, while skip-gram does the opposite.
 The objective of the CBOW approach, is to maximize the average log probability
\begin{equation}
 \frac{1}{L} \sum_{k=\eta}^{L-\eta}{ \log{p(w_{i_k} | C_{\eta}(w_{i_k}) ) } },
 \label{eq:Word2vec_objective_cbow}
\end{equation}
\noindent where $w_{i_k}$ is the central word and  $C_{\eta}(w_{i_k}) = [w_{i_{k-\eta}}, \ldots, w_{i_{k-1}}, w_{i_{k+1}}, \ldots, w_{i_{k+\eta}}]$ is the training context of size $\eta$. We may define similarly the performance index associated with the skip-gram model. Another interesting approach is due to \cite{collobert2008unified} (CW) who train a neural network to differentiate between a valid $n$-gram and a corrupted one. Furthermore, in these examples, we may note that we are only using local information. However, as we mentioned before, we may also use global information given by, for instance, the term-document matrix (similar to the term-sentence matrix considered in Section \ref{sec:extractive_vector_space}) to weight the performance index. All these algorithms have some design and hyperparameter choices and we may find very different results depending on them \citep{levy2015,Liu2017}.

In these models, the input and output layers have their sizes given by the vocabulary that arises in the collection of documents. In the CBOW model, the input layer indicates the words that arise in the context and the output layer indicates the desired output. In this model, they maximize the probability given by Eq. (\ref{eq:Word2vec_objective_cbow}) using the representation of the words given by vectors such as the ones presented in 
\begin{equation}p(w_{i_k} \mid C_{\eta}(w_{i_k}))=\frac{\exp{(v_{i_k}^{\prime} u_{i_k})}}{\sum_{l\in I_V} \exp{(v_{i_k}^{\prime} u_{i_l})}}, \label{eq:probcbow}\end{equation}
\noindent where the vector $v_{i_k}$ arises when  $w_{i_k}$ is the central word and the vector 
$u_{i_l}$ arises when $w_{i_l}$ belongs to the context $C_{\eta}(w_{i_k})$. Thus, in this model, we represent each word by two vectors. 
Neural word embeddings, denoted as $\mathbf{v}_{w_i}$,  are the average of these vectors and reside in  $ \mathbb{R}^{N_W}$, where $N_W$ represents the embedding dimension, a hyperparameter of the model. Analogous definitions can be applied to the skip-gram model.
A recent extension of these models are the so-called contextualized word embeddings \citep{liu2020survey} such as CoVe \citep{mccann2017learned} and ELMo \citep{peters2018}. In these models, each token has a representation that is a function of the entire text sequence. They are trained using sequence-to-sequence models discussed in our text, for the sake of organization, in Section \ref{sec:abstractive_sequence_seq2seq}.

There are several methods that we can use to consider the information encapsulated in word embeddings. The main motivation behind the use of word embeddings is to deal with the main drawbacks of the space vector models approach associated with the fact that similar words are treated separately: (1) Similar words may have very different rankings. Therefore, it fails to assign appropriate scores to the sentences; (2) The summary may be redundant, since the sentences of the summary may come from different words that have similar use and meaning.

One of the first ideas of using word embeddings in extractive summarization is due to \cite{kaageback2014extractive}. They use the setup of greedy submodular optimization due to \cite{lin2011class}, reviewed in Section \ref{sec:extractive_vector_space}, and different word embeddings (Word2Vec and CW) to extract sentences.

One of the simplest ideas is to use a kind of centroid method such as in \cite{Rossiello2017}. We may review this method using the following steps: (1) Create a representation of the documents in the dataset using the vector space models; (2) For each document, identify the most relevant words, i.e., the words that have a weight (provided by the space vector model) larger than a given threshold; (3) Evaluate the centroid of each document averaging the word embeddings of each word selected in the last step; (4) Evaluate the word embedding of each sentence in a document averaging the word embeddings of each word that arises in the sentence; (5) Identify the most relevant sentences that are the sentences that are the most similar to the centroid of the document.

\cite{mohd2020text} use word embeddings to find the $m$ most similar words to each word in a given sentence. Then each sentence is represented by a large vector of words, where each word in the original sentence is replaced by these $m$ most similar words previously found using the word embedding representation. With this new representation of each sentence, it applies TF-ISF to this new representation of the sentences and any algorithm presented in Section \ref{sec:extractive_vector_space} may be used to extract the most important sentences. In particular, they use a clustering method similar to \cite{Zhang2009}.

The idea behind the work of \cite{hailu2020framework} is to build a list of important words that they call keywords (first sentence words and high-frequency words) and to rank the sentences in the document according to the cosine similarity between the embeddings of the keywords and the embeddings of the words that form the sentences.   

There are many methods for ATS that uses word embeddings. We present a significant compilation of these methods in Table \ref{tab:digest_word_embeddings}.

\begingroup
\renewcommand{\arraystretch}{1.2}
\begin{landscape}
\begin{table}[t]
    \centering
    \scriptsize
    \begin{tabular}{lp{10cm}p{4cm}p{3cm}} 
     Source          & Main contribution & Dataset & Evaluation \\\hline
    \cite{kaageback2014extractive} & It uses the approach of optimization of submodular functions \citep{lin2011class} and the Word2Vec and CW representations in different setups. & Opinosis & ROUGE-1, ROUGE-2 and ROUGE-SU-4  \\\cdashline{1-4}
    \cite{yin2015optimizing}& It creates a sentence representation based on a kind of word embedding using a convolutional neural network (CNN) and, in a setup close to the ones used in graph-based methods presented in Section \ref{sec:extractive_graph}, it chooses the sentences that are nearly optimizing a function that trades off prestige (importance) and dissimilarity. & DUC 2002 and DUC 2004 & ROUGE-1, ROUGE-2 and ROUGE-SU-4   \\\cdashline{1-4}
    \cite{Rossiello2017}& It creates a representation of the documents in the dataset using the vector space model, it identifies the most relevant words using TF-IDF, and it uses the average word embedding of these words to represent the sentences and the document. It identifies the most relevant sentences, those that are the most similar to the centroid of the document. & DUC 2004& ROUGE-1 and ROUGE-2 \\\cdashline{1-4}
    \cite{mohd2020text}           & Using the Word2Vec representation, it represents each sentence by a large vector of words, where each word in the original sentence is replaced by these $m$ most similar words. Thus, it applies TF-ISF to account for the most important words and selects the most important sentences from the clusters formed using the TF-ISF representation. & DUC 2007 & ROUGE-1, ROUGE-2, ROUGE-L, ROUGE-SU-4 \\\cdashline{1-4}
    \cite{hailu2020framework}  & It builds a list of keywords based on first sentence words and high-frequency words and ranks the sentences according to the cosine similarity between the embeddings of the keywords and the embeddings of the words that form the sentences. & NEWSROOM summarization dataset & ROUGE-1, ROUGE-2, ROUGE-L, BLEU-1, BLEU-2, BLEU-3, BLEU-4, F-measure (of 1, 2 and 3 grams), WEEM4TSw, WEEM4TSg and WEEM4TSf.\\\cdashline{1-4}
    \cite{barman2021graph} & It represents each word by GLOVE vectors and each sentence by the average of the words it contains. In order to build the network, each sentence represents a node and the weights of the edges are evaluated by the cosine similarity. In order to rank the sentence, it uses the text ranking algorithm.  & BBC news & ROUGE-1 and ROUGE-2.
    
    \\\hline 
    \end{tabular}
    \caption{A significant compilation of ATS systems that use word embeddings.}
    \label{tab:digest_word_embeddings}
\end{table}
\end{landscape}
\endgroup

\subsection{Heuristic-based methods}
\label{sec:extractive_heuristics}
As above-mentioned, the field of ATS started with the heuristic approach of \cite{Luhn1958}.  As stated in this work, we should evaluate the importance of a sentence according to the frequency of the words that form it. In particular, in Luhn's work, the significance of a word depends on the frequency within the document, according to the following rules: (1) It does not consider pronouns, prepositions, and articles\footnote{We refer to these words today as ``stop words" and they are usually filtered out before any natural language processing analysis.}; (2) It does not consider least frequent words; (3) Words lying in the frequency range above least frequent are the significant ones. However, in order to select the sentences that should be extracted, this work also considers a heuristic evaluation of the physical distance among the significant words in a sentence that is used to determine the clusters of significant words. Using these definitions, relevant sentences are the ones that have clusters with a large number of significant words. \cite{edmundson1961automatic} make an important contribution since it is probably the first work that calls attention to the use of more informative measures of word frequency (such as the ones presented in Section \ref{sec:extractive_vector_space}) than the simple frequency. 

It is worth citing \cite{baxendale1958machine}, who compares three methods to extract the essential information of a text in order to mimic the way an average reader scans the content of a paper. The methods are based on scanning of topic sentences (the first sentence in 85\% of the cases), a syntactical deleting process (stop word removal), and an automatic selection of prepositional phrases (units of expression, composed of a preposition, a noun, or pronoun, together with appropriate modifiers). It  shows that the three methods provide equivalent results. Furthermore, one particular contribution of this paper is emphasizing the importance of the feature of sentence position in scanning the content of texts. However, this feature was further investigated in \cite{lin1997identifying} showing that we cannot define the position of the topic sentences a priori as in \cite{baxendale1958machine} since the discourse
structure significantly varies over domains. Thus, \cite{lin1997identifying} find the optimal position of the sentences by comparing them with the keywords associated with the text. They also evaluate the quality of the method by comparing the topic sentences determined by this method with the sentences found in abstracts provided by humans. 

\cite{Edmundson1969} includes additional heuristics to evaluate the importance of a sentence. This work particularly considers four methods: (1) {\it Cue method} (the presence of cue words): It considers that the relevance of a sentence is affected by the presence of pragmatic words such as ``significant", ``impossible" and ``hardly". (2) {\it Key method} (the presence of Keywords): It considers like \cite{Luhn1958} topic words based on frequency; (3) {\it Title method}: It considers the presence of the words in specific parts of the document such as the title or the headings; (4) {\it Location method} (position of the sentences): It assumes that important sentences may arise in specific parts of the document such as the beginning or the end. 

There are many methods for ATS that use heuristics. We present a significant compilation of these methods in Table \ref{tab:digest_heuristics}.

\begingroup
\renewcommand{\arraystretch}{1.2}
\begin{landscape}
\begin{table}[t]
    \centering
    \scriptsize
    \begin{tabular}{lp{10cm}p{4cm}p{5cm}}
     Source              & Main contribution & Dataset & Evaluation \\\hline
    \cite{Luhn1958}           & Seminal work. It uses the raw frequency of the words and a heuristic method to evaluate the distance of significant words in sentences.& 50 articles ranging from 300 to 4,500 words each. & Human experts                  \\\cdashline{1-4}
    \cite{baxendale1958machine}  & It compares three heuristics (scanning of topic sentences, a syntactical deleting process and selection of prepositional phrases).& 6 technical articles & Comparative frequencies between the abstracts generated by the author, automatically and by human experts. \\\cdashline{1-4}
    \cite{Edmundson1969}&   Besides the frequency of the words in the document, it also considers the position of the sentences, the presence of certain specific words, and the information provided by the title. & 200 Technical documents with approximate lengths ranging from 100 to 3900 words, with an average of 2500. & Human experts and a kind of precision and recall of the extracted sentences.\\\cdashline{1-4}
    \cite{white2003task} & In a query-based setup, it extracts sentences based on four attributes: title, location, sentence position and relation to the query. & Experiment-based task. & Human experts \\\cdashline{1-4}    
    \cite{yih2007multi} & In a multi-document setup, it introduces heuristics to evaluate the average position of the word in a cluster of documents and it assumes that good summaries should favor words with low averages. It builds a score for the sentences based on the average position and the frequency of the words. It selects the best sentences using a stack decoder\footnote{A decoder is the jargon term for an algorithm that performs the translation task. This term is very common in the sequence-to-sequence models presented in Section \ref{sec:abstractive_sequence_seq2seq}.  The term "stack decode" refers to the fact that hypotheses for the summary are maintained in stacks, which in this case are priority queues in order to avoid an exponential explosion in the number of solutions.} algorithm. & DUC 2004 and MSE 2005 & ROUGE-1, ROUGE-2 and ROUGE-SU-4 \\\hline
    \end{tabular}
    \caption{A representative compilation of the ATS methods that use sentence heuristics.}
    \label{tab:digest_heuristics}
\end{table}
\end{landscape}
\endgroup

\subsection{Linguistic-based methods}
\label{sec:extractive_linguistic}

There are different attempts to extract sentences to form abstracts based on linguistic methods. We may use the same classification that \cite{saranyamol2014survey} suggested to  classify full abstractive methods to be mentioned in Section \ref{sec:linguistic_abstractive}, namely {\it structured-based} and {\it semantic based}. The {\it structured-based} approach uses cognitive schema such as a set of rules, a template, a tree parser and a domain ontology. On the other hand, the {\it semantic-based} approach starts with a semantic representation of the document and uses this representation to identify the most important sentences of the document.   

\cite{rush1971automatic} is one of the first works to include linguistic information to decide whether a sentence should belong to the summary. Although this work shares several ingredients with \cite{Edmundson1969}, it has a clear concern about making inferences about the contextual importance of the sentence in a structured-rule-based approach. Thus, it provides a set of rules to select or not select a given sentence based on the extensions of the {\it location method} and {\it cue method}\, introduced by \cite{Edmundson1969}, and reviewed in Section \ref{sec:extractive_heuristics}. The authors assume that the location method is based on the physical arrangement of the linguistic elements of an article. This arrangement can be described in terms of the location of a sentence in the document, the location of words, or the punctuation in a sentence. They suggest that while the location of the sentence in a document is very subjective and it depends on the authors' choices, it is possible to get some pieces of relevant information using the location of words or punctuation in a sentence. Additionally, they claim that question marks should never be included in the summary for the following reasons: (1) they never provide a complete description of the facts; (2) if a question is selected, then the context behind the question should also be selected. It is worth mentioning that this discussion about whether a piece of text should be included provides the basic ideas for abstractive summarization discussed in Section \ref{sec:abstractive}. On the other hand, it defends that the cue method provides a powerful approach to sentence selection or rejection. For instance, words such as ``Our work," ``This paper," and ``Present research”, which are used to state the purpose of a paper, serve to indicate that such sentences should be selected for the abstract. On the other hand, opinions, or references to figures or tables such as ``obvious", ``believe", ``Fig.” ``Figure 1", and ``Table IV" should not be included in an abstract. Furthermore, cue words may also indicate the presence of inter-sentence references that are fundamental to the creation of an abstract. 

We may find an interesting example of a structured-ontology-based approach in \cite{wu2003ontology}. A domain ontology is a set of concepts and categories in a given subject area that shows their properties and the relations between them. \cite{wu2003ontology} encode an ontology with a tree structure, and each node includes the concepts represented by the node's children. When the count of any node increases, the counts associated with their ancestors also increase. They use this principle to score paragraphs. After marking the counts of the nodes in the ontology, the authors select second-level nodes that have higher counts as the main subtopics of the article. In order to implement the method, the authors only consider the top $n$ subtopics. Their system uses the obtained subtopics to select paragraphs to form the summary. They rank the paragraphs based on their ``closeness" to the selected subtopics by counting the words in common between the paragraph and each selected subtopic. Since we select $n$ possible subtopics, there are also $n$ scores associated with each paragraph, and these $n$ scores represent the relevance of the $n$ paragraphs for each selected topic. They assume that the score of each paragraph is the sum of its weighted relevance to the topics. 

We may find a kind of informative semantic-graph-based approach for multi-document summarization in \cite{canhasi2011semantic}. The authors use a {\it semantic role labeling} parser to extract each argument of the sentence. Semantic role labeling assigns labels to words or phrases indicating their semantic role. It is often described as a technique to answer ``Who did what to whom". Thus, the authors calculate the composite similarity between all semantic frames based on the event-indexing model \citep{zwaan1995construction} to keep track of five indices, namely temporality, spatiality, causality, and intention. Then, they generate a semantic graph where nodes are semantic frames and edges are the composite similarity values. In order to choose the most important sentences, they modify the Page Rank, used in Section \ref{sec:extractive_graph}, in order to identify the most significant edges in the graph. 

\cite{reeve2006biochain} present an interesting semantic-lexical-chain approach that uses semantically related concepts to identify the sentences useful for extraction. A lexical chain is a sequence of semantic-related ordered words \citep{morris1991lexical}. WordNet\footnote{WordNet is a large lexical database of English. It groups nouns, verbs, adjectives, and adverbs into sets of cognitive synonyms, the so-called synsets, each expressing a distinct concept \citep{Miller1995, Fellbaum2000, Wordnet}. Although WordNet superficially resembles a thesaurus, there are two main differences: (1) WordNet interlinks specific senses of words; (2) WordNet labels the semantic relations among words, whereas the groupings of words in a thesaurus does not follow any explicit pattern other than meaning similarity.} is an excellent source to extract lexical chains. For example, this lexical chain was extracted from WordNet: ``device $\rightarrow $ musical instrument $\rightarrow $ string instrument $\rightarrow $ guitar  $\rightarrow $ electrical guitar". \cite{reeve2006biochain} use a ``part-of-speech" tagger and a comprehensive thesaurus to map words in concepts. The sentences to be extracted are the ones that present the most important concepts. 

Table \ref{tab:digest_extractive_linguistic} significant compilation of ATS systems based on linguistic-extractive summarization.

\begingroup
\renewcommand{\arraystretch}{1.2}
\begin{landscape}
\begin{table}[t]
    \centering
    \scriptsize
    \begin{tabular}{llp{7cm}p{4cm}p{3cm}}
     Source          & Type    & Main contribution & Dataset & Evaluation\\\hline
    \cite{rush1971automatic} &   Structured-rule       & Seminal work. It explores the application of the Position and Cue methods to make inferences about the context importance of a sentence. Its implementation is based on a word list and a set of rules implementing certain functions specified for each word entry.& 5 academic papers & Human experts \\\cdashline{1-5}
    \cite{wu2003ontology} &  Structured-ontology       & This work encodes the ontology with a tree structure. A paragraph is said to be relevant if its words arise in the topics more common in the piece of text. & Collected articles from The New York Times and Wall Street Journal with summaries from the ProQuest database.   & Precision, Recall and F-measure of the selected paragraphs. \\\cdashline{1-5}
    \cite{chen2006} & Structured-ontology & In a query-based setup, it uses an ontology to create a summary with extracted sentences that are close to the provided query. It works in five steps: (1) The user provides a query; (2) The query is revised by an ontology; (3) It calculates the distance of each sentence of the document to the query; (4) It calculates the pairwise distance between the sentences in order to avoid redundancy; (5) It divides the sentences into groups and selects the highest one from each group. & The paper \cite{tashimo2004aqueous}. & Precision and recall of extracted sentences.\\\cdashline{1-5}
    \cite{reeve2006biochain}& Semantic-lexical-chain & It maps words in concepts and extracts the sentences with the most important concepts.  & Drexel University dataset with approximately 1,200 oncology clinical trial documents that have been manually selected, evaluated, and summarized. & Human experts \\\cdashline{1-5}
    \cite{canhasi2011semantic} & Semantic-graph & This work builds an informative semantic-graph-based approach on the event-indexing model \citep{zwaan1995construction} for multi-document summarization to keep track of four dimensions, namely temporality, spatiality, causality, and intention. It uses a modified Page Rank to select the most important sentences.& DUC 2004 & ROUGE-1
    \\\cdashline{1-5}
    \cite{saleh2017language} & Structured-rule & It proposes a structured-rule approach that does not depend on specific characteristics of the language. It extracts sentences that have higher scores that depend on sentence shapes (for instance, the number of words with capital letters) and $n$-grams statistics. & DUC 2002, TeMario, and French and Spanish documents collected from news websites. & ROUGE-1 and ROUGE-2  \\\cdashline{1-5}
    \cite{mohamed2019srl} & Semantic-graph & This work presents a semantic-graph-based approach  for single- and multi-document summarization. It uses semantic role labeling to build a semantic representation of documents and it pairs matching roles for any two sentences. Then, it maps the sentences onto their corresponding concepts in Wikipedia. It builds a weighted semantic graph where each sentence is modeled as a multi-node vertex containing the Wikipedia concepts of its semantic arguments. It scores the sentences using the Page Rank algorithm. & DUC 2002 & ROUGE-1, ROUGE-2 and ROUGE-SU-4 
    \\\hline 
    \end{tabular}
    \caption{A significant compilation of ATS systems based on linguistic-extractive summarization.}
    \label{tab:digest_extractive_linguistic}
\end{table}
\end{landscape}
\endgroup

\subsection{Supervised machine learning-based methods}
\label{sec:extractive_ml}

The objective of using supervised machine learning methods in ATS is to explore the problem of ATS as a binary classification problem in which we use supervised methods of machine learning to select the sentences that should arise in the summary using their features. 

This classical and conventional approach follows the steps: 

\begin{enumerate}
    \item Tag the sentences manually in the documents as positive ones (the ones that should be extracted to build the summary) and the negative ones (the opposite).
    \item Select the features associated with each sentence that are used as inputs in the classification algorithm. These features may be word-based features and sentence-based features. The word-based features may be, for instance, weights that come from the space vector representation or the word embeddings. On the other hand, the sentence-based features may be the sentence position or the sentence length. 
    \item Estimate the model. These models usually have hyperparameters that have to be set before the estimation of the parameters. Since there is no way to choose these hyperparameters without data experimentation, a cross-validation\footnote{ Cross-validation is a resampling method that splits the data into different sets in order to test and train a model on different iterations.} \citep{stone1978cross} procedure is necessary.
    \item Apply the classification algorithm to find out the sentences that should be included in the summary.
\end{enumerate}

There are several supervised machine learning classifiers in the literature and, in essence, we can use any of them. A machine learning classifier is a mathematical model that, given a set of attributes, provides a label associated with a class. The most popular are the {\it logistic regression} \citep{berkson1944application,berkson1951prefer} and their regularized versions {\it LASSO-logistic}, {\it Ridge-logistic} and {\it elastic-net logistic} \citep{friedman2010regularization,simon2011regularization}, maximum entropy classifier \citep{nigam1999using}, {\it naive Bayes} \citep{duda1973pattern,domingos1997optimality}, {\it hidden Markov model} \citep{rabiner1986introduction}, the Tree-based approaches such as {\it decision tree} \citep{breiman1984classification}, {\it random forest} \citep{breiman2001random} and {\it gradient boosting} \citep{friedman2001greedy}, the Support Vector Machine (SVM) \citep{cortes1995support} and its rank version \citep{joachims2006training}, and the  multilayer perceptron \citep{rumelhart1985learning}.   
We may find a review of the supervised machine learning methods in \cite{bishop2006pattern} and \cite{izenman2008modern} and a review of  classical neural network models in \cite{haykin1994neural}.

On ther other hand, modern approachs based on neural networks are able to replace the second and third step above by the automatical selection of a ``composition" of features that are able to solve the binary classification problem of extractive summarization.
There are different neural network models that we can use such as multilayer perceptron \citep{rumelhart1985learning}, convolution neural networks \citep{zhang1988shift,lecun1989backpropagation,li2021survey} and sequence-to-sequence neural network models as reviewed in Section \ref{sec:abstractive_sequence_seq2seq}. We may find a review of the  deep neural network models in \cite{goodfellow2016deep}.

It is worth mentioning that one difficulty that this approach faces is that human-made summaries are usually not extractive. Thus, there are few datasets available for this purpose, such as the DUC 2002 dataset \citep{over2007duc} and CNN \citep{lins2019cnn}. However, these datasets are small compared with other human-made datasets available (for instance, datasets formed by summaries of scientific papers).  Thus,  one relevant issue is how  to create a dataset with a large number of documents with sentences labeled as to whether they should be extracted or not. The general approach to executing this task is to label the sentences of a document based on a human-made summary. Therefore, we have to solve the inverse problem of finding the sentences that we have to extract to meet the human-made summary. In practice, this is done by a function that associates some statistics of the sentences of the document we need to summarize to the labels 0 or 1. We may find an example of this approach in  \cite{cheng2016neural} where they use the highlights created by editors to label the sentences of news articles. We may find another example in \cite{nallapati2017summarunner}. The authors assign the label 1 to the sentences that maximize the ROUGE score of the human-made summaries. We should note that the problem of adjusting datasets for natural language processing is not limited to the field of ATS. Several approaches, for instance, were introduced to augment datasets \citep{chen2021empirical,feng2021survey,shorten2021text}.

The first paper that used a supervised approach to extractive summarization is \cite{Kupiec95}. This work uses several sentence features such as whether the sentences include the most frequent words of the document, whether the sentences include words presented in the title or in the list of keywords associated with the document, the position of the sentences in the document (important sentences usually arise at the beginning or the end of the document) and if the sentences contain indicator phrases such as "This report (...)". 

We may find an interesting contribution to summarization in \cite{conroy2001text}. With the motivation to model the local dependencies between sentences, the authors use a Hidden Markov Model (HMM) that models the transitions between sentences that should or should not belong to the summary. They use only three features, namely the position of the sentence in the document, the number of terms in the sentence, and how likely sentence terms are in the document.

\cite{osborne2002using} calls attention to the fact that the assumption of independence of features of the Naive Bayes is a strong assumption and the maximum entropy classifier outperforms this model. In particular, this work uses the following features: {\it word pairs}, which simply tells whether a particular word pair is present, {\it sentence position}, which indicates whether the sentence belongs to the beginning, the middle or the end of the document and {\it discourse features}, which informs if the sentence belongs to the introduction or conclusion or is located in the start of the paragraph.

An interesting contribution comes from \cite{svore2007enhancing} that uses a feedforward neural network to find the most important sentences of a piece of text. The neural network is trained using the RankNet algorithm \citep{burges2005learning}\footnote{RankNet is a pair-based gradient descent algorithm used to rank a set of inputs, in this case, the set of sentences in a given document.}. For each sentence, the work considers several features such as the position of the sentence, $n$-grams of words in the sentence, terms common with the title and the presence of some specific words such as in \cite{Edmundson1969}. A particular innovation in this work is to use features from third-party sources such as query logs from Microsoft’s news search engine\footnote{http://search.live.com/news or http://www.bing.com/news} and Wikipedia\footnote{http://www.wikipedia.org} entries.

\cite{leskovec2005extracting} and a sequence of papers \citep{leskovec2004learning,leskovec2004learningb,rusu2009semantic} extract semantic information such as subject–object–predicate triples, co-reference resolution and anaphora resolution and use these pieces of information as additional inputs of a classifier. The other inputs are the location of the sentence within the document, the triplet location within the sentence, the frequency of the triplet element, the number of named entities in the sentence and the similarity of the sentence with the centroid (the central words of the document). 

\cite{jain2017}, besides using several of these previously discussed attributes to characterize the sentences of a document, they also use the mean of word embeddings associated with each word of a sentence as an additional attribute. 

In order to deal with a query-based task, AttSum \citep{cao2016attsum} builds a convolutional neural network composed essentially of three major layers: (1) A convolutional neural network layer to project the sentences and queries onto the embeddings; (2) A pooling layer to combine the sentence embeddings to form the document embedding in the same latent space and to indicate the query relevance of a sentence; (3) A ranking layer that ranks sentences according to the similarity between its embedding and the embedding of the document cluster.

The sequence-to-sequence neural networks models \citep{nallapati2017summarunner,zhou2018neural,liu2019text}, discussed in Section \ref{sec:abstractive_sequence_seq2seq}, as we mention in the beginning of the section, different from the classical and conventional approaches, are able to automatically extract features of the sentences and represent them mathematically by internal weights of the neural network that can be used to classify the sentences as the ones that should belong to the summary. 

Finally, it is worth mentioning that these works use attributes that come from different models previously discussed, such as frequency metrics presented in Section \ref{sec:extractive_vector_space}, word embeddings presented in Section \ref{sec:extractive_embedding} and heuristic choices as in Section \ref{sec:extractive_heuristics}.

There are many supervised machine learning methods for ATS systems. We present a significant compilation of these methods in Table \ref{tab:digest_suppervised}.

\begingroup
\renewcommand{\arraystretch}{1.2}
\begin{landscape}
\begin{table}[t]
    \centering
    \scriptsize
    \begin{tabular}{lp{2.4cm}p{9cm}p{3.5cm}p{3.5cm}}
     Source          & Model    & Features & Dataset & Evaluation\\\hline
    \cite{Kupiec95}  & Naive Bayes  &  Sentence attributes: the presence of frequent or special words, the position and the presence of indicator phrases.& 188 document-summary pairs, sampled from 21 publications in the scientific and technical domains. & Precision and Recall of the sentences \\\cdashline{1-5}
    \cite{conroy2001text} & HMM & Sentence attributes: the position, number of terms, and how likely sentence terms are in the document. & TREC Conference data set & A kind of Recall of the sentences \\\cdashline{1-5}
    \cite{osborne2002using}& Maximum entropy classifier & Sentence attributes: word pairs and different measures of sentence position (position in the document, if it belongs to introduction or conclusion, position in a paragraph). & 80 conference papers \citep{teufel2001task} & Precision, Recall and F-measure. \\\cdashline{1-5}
    \cite{leskovec2005extracting} & Support vector machine & Sentence attributes: subject, predicate, object triples,  part of speech tags, and about 70 semantic tags (such as gender, location name, person name), location of the sentence in the document and the triple in the sentence, frequency, location of the word inside the sentence and number of different senses of the word.  & DUC 2002  & Precision, recall and F1 measures of the extracted sentences and ROUGE-1\\\cdashline{1-5}
    \cite{svore2007enhancing} & Feedforward neural network & Sentence attributes: position, $n$-grams of words, terms common with the title, the presence of some specific words, query logs from Microsoft’s news search engine and Wikipedia entries. & 1365 news documents gathered from CNN.com. & ROUGE-1 and ROUGE-2\\\cdashline{1-5}
    \cite{fattah2009ga} & Linear regression, neural network, gaussian mixture model & Sentence attributes: position, positive and negative keywords, sentence centrality, Sentence resemblance to the title, sentence inclusion of name entity or numerical data and sentence relative length. & 200 Arabic articles about politics and 150 English articles about religion & Precision\\\cdashline{1-5}
    \cite{nguyen2016learning} & SVM rank & This is a social context summarization approach. It uses a lot of features and it splits into local features (such as sentence attributes and different measures of similarities) and social features (that measure the similarities between a sentence and the comments associated with it). & Solscsum and USAToday-CNN & ROUGE-1 and ROUGE-2\\\cdashline{1-5}
    \cite{cao2016attsum} & Convolution neural network & In order to deal with a query-based task, It introduces a neural network model with an additional layer that aims to learn the query relevance of the sentence. & DUC 2005 and DUC 2007 & ROUGE-1 and ROUGE-2\\\cdashline{1-5}
    \cite{jain2017} & MLP & Sentence attributes: mean TF-ISF, length, position, correlation with other sentences, correlation with the document centroid, depth of tree in an agglomerative cluster, presence of keywords, presence of proper names, presence of non-essential information and mean GLOVE word embedding.& 10K documents from CNN news article corpus having 90K documents \citep{hermann2015teaching} & ROUGE-1, ROUGE-2 and ROUGE-L .\\\cdashline{1-5}
    \cite{nallapati2017summarunner} & GRU Recurrent Neural Network & The GRU \citep{cho2014learning}is able to recover sentence characteristics such as content, saliency and novelty. & CNN/DailyMail & ROUGE-1, ROUGE-2 and ROUGE-L \\\cdashline{1-5}
    \cite{zhou2018neural} & Bidirectional GRU and RNN & It builds a model based on an encoder and a sentence extractor. The encoder uses a bidirectional GRU \citep{cho2014learning} to create a representation of the sentences. The sentence extractor is a recurrent neural network that remembers the partial output summary and provides a sentence extraction state that can be used to score sentences with their representations. & CNN/Daily Mail & ROUGE-1, ROUGE-2 and ROUGE-L\\\cdashline{1-5}
    \cite{liu2019text}  & BERT & The modified BERT transformer \citep{devlin2018bert} for extractive summarization is capable of extracting automatically the features in the internal layers. & CNN/DailyMail & ROUGE-1, ROUGE-2 and ROUGE-L\\\hline 
    \end{tabular}
    \caption{A significant compilation of ATS systems that use supervised machine learning.}
    \label{tab:digest_suppervised}
\end{table}
\end{landscape}
\endgroup

\subsection{Reinforcement learning based methods}
\label{sec:extractive_reinforcement}

A reinforcement learning problem explores a situation where the objective is to map states to actions in order to maximize a numerical reward signal. We usually define a reinforcement learning solution with the following ingredients \citep{puterman2014markov,bertsekas2012dynamic,bertsekas2011dynamic,bertsekas1996neuro,sutton2018reinforcement}:

\begin{enumerate}
    \item An agent (the learner or the decision-maker) that interacts with the environment in a sequence of instants $t=0,1,2,\cdots.$
    \item At each instant $t$, the agent faces a state $s_t\in \mathcal{S}$, where $\mathcal{S}$ is a finite set of possible states, and selects an action $a_t\in \mathcal{A}(s_t)$, where $\mathcal{A}(s_t)$ is the set of admissible actions contingent to the state $s_t$.
    \item At each state the agent receives a reward that is contingent to the chosen action $r:\mathcal{S}_t\times \mathcal{A}(s_t)\rightarrow \Re$. We usually assume that $\max_{s\in S} \max_{a\in \mathcal{A}(s) }r(s,a)<\infty$.
    \item In each state, the agent maps states to probabilities of selecting a possible action. This map is called agent policy and it is denoted by $\Pi_t$, where $\Pi_t(s,a)$ is the probability that $a_t=a$ when $s_t=s$.
    \item In each state $s_t$, the agent intends to maximize the expected value of the return given by
    \begin{equation}
        v_{\gamma}^{\pi}(s)=E^\pi\left[R_t\Big/s_t=s
        \right] =E^\pi\left[\sum_{k=0}^{\infty}\gamma^{k}r_{t+k+1}\Big/s_t=s
        \right] \label{eq:RLindex}
    \end{equation}
    \noindent where the return is given by $R_t=\sum_{k=0}^{\infty}\gamma^{k}r_{t+k+1}$ and $E^\pi\left[\sum_{k=0}^{\infty}\gamma^{k}r_{t+k+1}\Big/s_t=s\right]$ is the expected value of the discounted sequence of rewards received by the agent assuming that in the beginning of the trajectory it was in state $s$ and, from this state, it followed the policy $\pi$. A common assumption is that the rewards are bounded and the distributions of rewards and states are stationary. The discount factor $0<\gamma<1$ has two functions. From the economic point of view, it makes explicit the value of money over time. On the other hand, from the mathematical point of view, it ensures that under mild regularity conditions, there is an optimal policy that maximizes the performance index presented in Eq. (\ref{eq:RLindex}). Another important assumption here is that we may describe the transitions between the states as a Markovian process, i.e., 
    \begin{equation}
        P(s_{t+1}=s',r_{t+1}=r/s_t,a_t,r_t,\cdots,s_{0},a_{0},r_{0})=P(s_{t+1}=s',r_{t+1}=r/s_t,a_t,r_t).
    \end{equation}
    This assumption allows a parsimonious representation that reduces the computational complexity of the problem and simplifies the definition of the transition matrices, the reward structure and the set of admissible actions.
    \item If the rewards and the transition probabilities are known and stationary, we may show that the solution of the problem is given by the Bellman equation:
    \begin{equation} 
        v_{\gamma}^{\pi}=\mathtt{R}_{\pi}\nonumber +\gamma \mathtt{P}_{\pi} v_{\gamma}^{\pi},\label{eq:value_function} (9)
    \end{equation}
    where $\mathtt{P}_{\pi}=[\mathtt{P}_{ss'}]$ is the transition matrix contingent to the policy $\pi$, $\mathtt{R}_{\pi}=[\mathtt{R}_s]$ is the returns vector contingent to the policy $\pi$, $\mathtt{P}_{ss'}=\sum_{a\in \mathcal{A}(s)}q_{s}^{a} P_{ss'}^{a}$, $\mathtt{R}_{ss'}^{a}=E[r_{t+1}/s_t=s,a_t=a,s_{t+1}=s']$ and $q_{s}^{a}$ is the probability of using the action $a$ in state $s$. In order to solve this problem, we need to know the value function given the policy $\pi$, which is a problem known as {\it policy evaluation}, and to improve the policy given another policy, which is a problem known as {\it policy improvement}. Note that the problem of policy evaluation is equivalent to finding the solution of a linear system, i.e., given the policy and assuming that the rewards and probabilities are stationary and known, Eq. (\ref{eq:value_function}) is a linear system. We can prove that the operator $L(v)=\mathtt{R}_{\pi}\nonumber +\gamma \mathtt{P}_{\pi}v$ is a contraction and, according to the Banach fixed point theorem, it can be solved by fixed point iterations\footnote{See, for instance, \cite{puterman2014markov}.}. In order to improve the policy, we use the Q-function \(Q^\pi(s,a)=E_\pi[R_t/s_t, a_t=a]\), that is the expected return for taking action $a$ in state $s$ and thereafter following an optimal policy. Thus, if \(Q^\pi(s,a)>v^\pi(s)\), then $\pi$ can be improved if $\pi(s)$ is replaced by $a$. There are two popular algorithms to solve this problem, namely policy iteration and improvement \citep{watkins1989learning} and value iteration \citep{puterman1978modified}. While in the former approach the  policy improvement and policy evaluation tasks are run in separate loops, in the latter, these are carried out in the same loop.
    \item When the transition probabilities and rewards are not known, we use Monte Carlo methods to sample sequences of states, actions, and rewards \citep{michie1968boxes,barto1993monte,singh1996reinforcement}. The policy evaluation and policy improvement steps are completed using average returns of episodic tasks.  
    \item Temporal difference learning methods combine Monte Carlo and dynamic programming approaches  \citep{sutton1988learning}. They learn from experience like the Monte Carlo methods and they update estimates like the dynamical programming approach. Two algorithms can be used to implement the temporal difference approach, namely SARSA (an Acronym for State-Action-Reward-[next]State-[next]Action)  and Q-learning. While the Q-learning approach updates the value function using a greedy action approach \citep{rummery1994line,sutton1995generalization}, SARSA updates with the actual action used for generating experience by the agent \citep{watkins1989learning}.
\end{enumerate}

\cite{ryang2012framework}, in order to formulate the problem of extractive summarization as a reinforcement learning problem, reduce the document to be summarized in a set of $n$ sentences, define a score function for any subset of sentences of the document $S\subset d$, where $S$ is one of the possible summaries and $d$ is the document, and also define the length function that indicates the size of the summary. They use a temporal difference learning approach to find the summary that maximizes the score function that considers a trade-off between relevance and redundancy subject to the maximal summary length. Thus, in this problem, a state is a summary of a given length. In this work, the actions are deterministic and are basically the decision to include or not a given sentence. The reward is defined in such a way that the agent only receives the rewards when the summary reaches the final size. The score function specifically depends on the coverage of important words (the count of top-100 words in terms of the TF-IDF included), coverage ratio (the count of top-100 elements included), redundancy ratio (the counting of the number of elements that excessively cover the top 100 elements), length ratio (the ratio between the length of the summary and length limitation) and the sum of the inverse position of the sentences. 

We may find an extension of the work of \cite{ryang2012framework} in \cite{rioux2014fear}. In a multi-document setup and also working with a query-based task, they use the SARSA algorithm, explore different types of rewards (not only delayed rewards as in the case of \cite{ryang2012framework}) and also use the ROUGE to evaluate the similarity between the reference summary and the automatic summary. 

There are other reinforcement-based approaches for ATS systems. We present a significant compilation of these methods in Table \ref{tab:digest_reinforcement}.

\begingroup
\renewcommand{\arraystretch}{1.2}
\begin{landscape}
\begin{table}[t]
    \centering
    \scriptsize
    \begin{tabular}{lp{7cm}p{4cm}p{3cm}}
     Source          & Main contribution & Dataset & Evaluation\\\hline
    \cite{ryang2012framework}  & It introduces a reinforcement learning model for extractive summarization. & DUC 2004 & ROUGE-1, ROUGE-2 and ROUGE-3 \\\cdashline{1-4}
    \cite{rioux2014fear}  & In a multi-document approach, it extends the work of \cite{ryang2012framework} using the SARSA algorithm, considering different types of rewards and using the ROUGE as a measure of similarity. & DUC 2004 and DUC 2006 & ROUGE-1, ROUGE-2 and ROUGE-L \\\cdashline{1-4}
    \cite{henss2015reinforcement} & In a multi-document summarization setup, it extends the work of \cite{ryang2012framework} using the Q-learning algorithm and considering the reference summaries in the training phase. & DUC 2001, DUC 2002, DUC 2004, ACL and Wikipedia & ROUGE-2, ROUGE-2 and ROUGE-L\\\cdashline{1-4}
    \cite{molla2017towards} & In order to deal with a query-based task, it extends the work of \cite{ryang2012framework} including the use of the reference summary in the evaluation of the reward.  & BioASQ 5b Phase B & ROUGE-L \\\cdashline{1-4}
    \cite{lee2017automatic} & It explores the advantages of embedding features in a Q-learning model and also designs a deep neural network to approximate the value function. & DUC 2001, DUC 2002, ACL-ARC and Wikipedia & ROUGE-2\\\cdashline{1-4}
    \cite{Narayan2018RankingSF} & It formulates the problem of extractive summarization as a sentence ranking problem using reinforcement learning and it uses a neural network model to encode the features of the sentences. & CNN and Daily-Mail & ROUGE-1, ROUGE-2 and ROUGE-L\\\cdashline{1-4}
    \cite{yao2018deep} & It develops a neural network for both encoding the features of the sentences and also choosing the sentences that must be included in the summary.  & CNN and Daily-Mail & ROUGE-1, ROUGE-2 and ROUGE-L
    \\\hline 
    \end{tabular}
    \caption{A significant compilation of ATS systems that use reinforcement learning techniques.}
    \label{tab:digest_reinforcement}
\end{table}
\end{landscape}
\endgroup

\section{Abstractive summarization}
\label{sec:abstractive}

Different from extractive summarization, which builds a summary from the combination of the important sentences previously extracted from the original text, in abstractive summarization, we need a language model to rewrite the summary from scratch. We split this section into two subsections. While in subsection \ref{sec:linguistic_abstractive} we present the ATS systems based on classical linguistic models, in Subsection \ref{sec:abstractive_sequence_seq2seq} we present the modern methods based on sequence-to-sequence neural network models.

\subsection{Linguistic approaches}
\label{sec:linguistic_abstractive}

As we have mentioned in Section \ref{sec:extractive_linguistic}, we may split the linguistic approaches to abstractive summarization into two large groups, namely {\it structured-based} and {\it semantic-based} \citep{saranyamol2014survey}. 

An example of the {\it structured-based} approach that uses a {\it rule-based} scheme is \cite{genest2012fully}. In this work, in order to provide summaries in the fields of ``Accidents and Natural Disasters", ``Attacks, Health and Safety", ``Endangered Resources", and ``Investigations/Trials", the authors use handcrafted information extraction rules, content selection heuristics and generation patterns. In particular, to extract the information they need to build the summary, they ask the following questions: (1) What: what happened; (2) When: date, time, other temporal placement markers; (3) Where: physical location; (4) Perpetrators: individuals or groups responsible for the attack; (5) Why: reasons for the attack; (6) Who was affected: casualties (death, injury), or individuals otherwise negatively affected; (7) Damages: damages caused by the attack; (8) Countermeasures: countermeasures, rescue efforts, prevention efforts, other reactions. With the answers to these questions in hand, extracted using predefined extraction rules, they use previously generated patterns to build the summaries. The reader may note that this is a particular example of an informative summary defined in Section \ref{sec:classification}.

An example of the structured approach that uses a {\it template} for multi-document summarization is \cite{harabagiu2002generating}. In this work, the authors use templates that represent each topic of the piece of text that need to be summarized and are populated using information extraction rules. In order to generate the summaries, they use a parse tree to identify the Subject-Verb-Object structures\footnote{In linguistic, subject–verb–object (SVO) is a sentence structure where the subject comes first, the verb second, and the object third.} and WordNet to classify the topic of each structure. The topics with a high frequency are included in the summary. 

We may find an example of the {\it structured-approach} that uses a {\it domain ontology} in \cite{lee2005fuzzy}. In this work, the authors extend a domain ontology using concepts of fuzzy logic by embedding a set of membership degrees in each concept of the domain ontology. They use this fuzzy domain ontology to extract the sentences related to the text domain and  to generate the abstract by concatenating concepts with relations.

We may find one of the first examples of the {\it semantic-based} approach in \cite{genest2011framework}. This work proposes the concept of {\it Information Items} (INITs) to help to define the abstract representation, which is the smallest  element of coherent information in a text or a sentence. The goal is to identify all entities in the text, their properties, the predicates between them, and the characteristics of the predicates. In that work, the implementation of INITs is constrained to dated and located subject–verb–object (SVO) triples and the summary generation as above-mentioned is carried out using parse trees. This work is another example of an ATS that generates informative summaries.

Another interesting example of the {\it semantic} based approach is the {\it multimodal model} of \cite{greenbacker2011towards}. We may summarize the implementation of this model in three steps: (1) Building the semantic model: The semantic model, which should consider both images and pieces of text, is built based on a knowledge representation based on a domain ontology; (2) Rating the informational content: In order to rate the content, the authors propose the so-called information density metric (ID) which rates a concept's importance based on factors such as completeness of attributes, the number of connections with other concepts and the number of expressions that put the concept in evidence; (3) Generating a summary: The work uses the concepts of TAG (Tree Adjoining Grammars) Derivation Trees\footnote{A TAG is a formalism that builds grammatical representations through the composition of smaller pieces of syntactic structure \citep{joshi1997tree}.} as in \cite{mcdonald2010if} to express the concepts and relationships found in previous steps.

\cite{moawad2012semantic} implement the {\it semantic-based} approach using a {\it semantic graph}. The method consists of three steps: (1) The creation of the semantic graph called Rich Semantic Graph (RSG) for the original document; (2) The  reduction of the generated semantic graph; (3) The generation of the final abstractive summary from the reduced semantic graph. In RSG, the verbs and nouns of the input document are represented as graph nodes along with edges corresponding to semantic and topological relations between them. The graph nodes are instances of the corresponding verb and noun classes in the domain ontology. The Rich Semantic Graph Reduction Phase aims to reduce the generated rich semantic graph of the source document to a reduced graph. A model of heuristic rules is applied to reduce the graph by replacing, deleting, or consolidating the graph nodes using the WordNet relations. Finally, the Summarized Text Generation Phase aims to generate the abstractive summary from the reduced rich semantic graph. To achieve its task, this phase accesses a domain ontology, which contains the information needed in the same domain of RSG to generate the final texts.

Table \ref{tab:digest_abstractive_linguistic} presents a significant compilation of ATS systems based on linguistic abstractive summarization.

\begingroup
\renewcommand{\arraystretch}{1.2}
\begin{landscape}
\begin{table}[t]
    \centering
    \scriptsize
    \begin{tabular}{llp{7cm}p{4cm}p{3cm}}
     Source          & Type    & Main contribution & Dataset & Evaluation\\\hline
    \cite{harabagiu2002generating} & Structured-template & It presents a multi-document summarization approach that uses a template to identify the important pieces of information, a tree parser to build subject-verb-object structures, the WordNet to identify the topics and generates the abstract based on the high-frequency structures.& DUC 2002 & Human experts \\\cdashline{1-5}
    \cite{lee2005fuzzy}  & Structured-domain ontology  & It extends a domain ontology using concepts of fuzzy logic. It uses this domain ontology to extract the sentences and to generate the abstract  &  Chinese news of three weather events, including “Typhoon event,” “Cold current event,” and “Rain event,” from the Chinatimes website in 2002, 2003, and 2004. \\\cdashline{1-5}
    \cite{genest2011framework}  & Semantic-INITs  & It extracts sentences using a variety of methods and produces the summary with these sentences using a parse-tree.  & Text Analysis Conference (TAC) 2010  & Precision and Recall evaluated by human experts \\\cdashline{1-5}
    \cite{greenbacker2011towards}  & Semantic-multimodal  & It implements the model in three steps: (1) Building the semantic model based on a domain ontology; (2) Rating the information content based on completeness of attributes, connections with other concepts and the number of expressions that put the concept in evidence; (3) Generating the summary using TAGs. & An article from the May 29, 2006 edition of Businessweek magazine entitled, ``Will Medtronic's Pulse Quicken?". & Human experts.  \\\cdashline{1-5}
    \cite{genest2012fully}  & Structured-rule  & It uses handcrafted information extraction rules, content selection heuristics and generation patterns.  & Attack category of Text Analysis Conference (TAC) 2011 \citep{owczarzak2011overview} & Manual Pyramid \\\cdashline{1-5}
    \cite{moawad2012semantic}  & Semantic-graph  & It implements the model in three steps: (1) It creates the semantic graph; (2) It reduces  the generated semantic graph; (3) It generates the final abstractive summary from the reduced semantic graph. & A simulated case study called ``Graduate students". & Text coherence using the original words of the sentences and using the synonyms of the words of the sentences. 
    \\\hline 
    \end{tabular}
    \caption{A significant compilation of ATS systems based on linguistic abstractive summarization.}
    \label{tab:digest_abstractive_linguistic}
\end{table}
\end{landscape}
\endgroup

\subsection{Sequence-to-sequence deep learning methods}
\label{sec:abstractive_sequence_seq2seq}

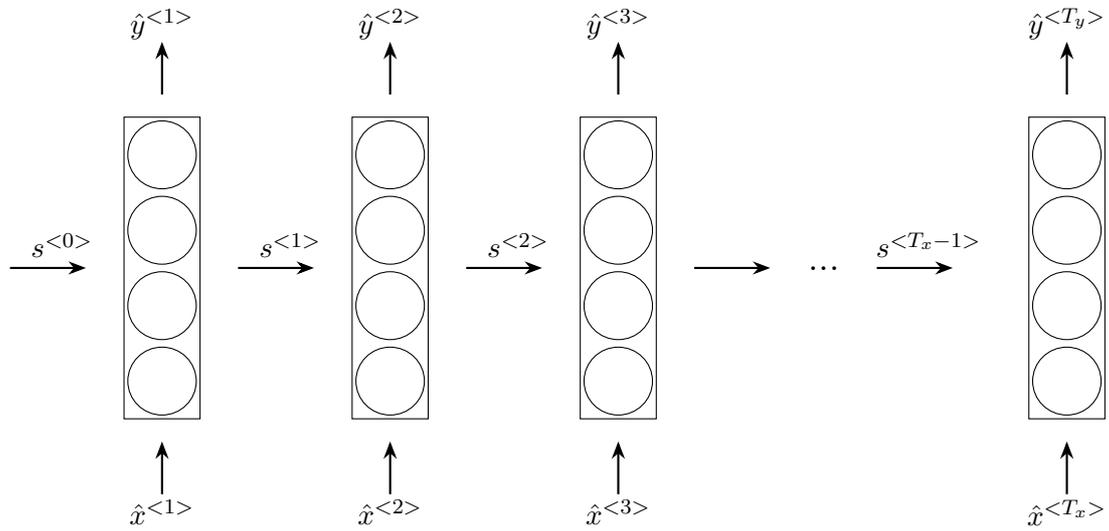
\begin{figure}[b]
    \centering
        \caption{Vanilla RNN.}

 \begin{tikzpicture}

\def\RNN at (#1,#2){\begin{scope}[xshift=#1 cm,yshift=#2 cm]
\draw [fill=none,draw=black] (0,0) rectangle ++(1,4); 
\filldraw[color=black, fill=none](.5,.5) circle (.45);
\filldraw[color=black, fill=none](.5,1.5) circle (.45);
\filldraw[color=black, fill=none](.5,2.5) circle (.45);
\filldraw[color=black, fill=none](.5,3.5) circle (.45);
\end{scope}}

\RNN at (0,0);
\RNN at (3,0);
\RNN at (6,0);
\RNN at (11.9,0);

\draw [-Stealth, thick](-1.5,2) -- (-0.5,2) node [above, xshift=-9pt] {$s^{<0>}$}; 
\draw [-Stealth, thick](1.5,2) -- (2.5,2) node [above, xshift=-9pt] {$s^{<1>}$}; 
\draw [-Stealth, thick](4.5,2) -- (5.5,2) node [above, xshift=-9pt] {$s^{<2>}$}; 
\draw [-Stealth, thick](7.5,2) -- (8.5,2) node [right, xshift=10pt] {\Large ...}; 
\draw [-Stealth, thick](9.9,2) -- (10.9,2) node [above, xshift=-9pt] {$s^{<T_x-1>}$}; 
\draw [-Stealth, thick](0.5,4.3) -- (0.5,5) node [above, xshift=0pt,yshift=-3pt] {$\hat{y}^{<1>}$}; 
\draw [-Stealth, thick](3.5,4.3) -- (3.5,5) node [above, xshift=0pt,yshift=-3pt] {$\hat{y}^{<2>}$}; 
\draw [-Stealth, thick](6.5,4.3) -- (6.5,5) node [above, xshift=0pt,yshift=-3pt] {$\hat{y}^{<3>}$}; 
\draw [-Stealth, thick](12.4,4.3) -- (12.4,5) node [above, xshift=0pt,yshift=-3pt] {$\hat{y}^{<T_y>}$}; 
\draw [-Stealth, thick](0.5,-1) -- (0.5,-0.3) node [below, xshift=0pt,yshift=-18pt] {$\hat{x}^{<1>}$}; 
\draw [-Stealth, thick](3.5,-1) -- (3.5,-0.3) node [below, xshift=0pt,yshift=-18pt] {$\hat{x}^{<2>}$}; 
\draw [-Stealth, thick](6.5,-1) -- (6.5,-0.3) node [below, xshift=0pt,yshift=-18pt] {$\hat{x}^{<3>}$}; 
\draw [-Stealth, thick](12.4,-1) -- (12.4,-0.3) node [below, xshift=0pt,yshift=-18pt] {$\hat{x}^{<T_x>}$}; 

\end{tikzpicture}
    \label{fig:rnn}
\end{figure}

Sequence-to-sequence deep learning models used for tasks of NLP are NN models in which the inputs and  outputs are sequences of tokens of varying sizes. In order to explore the most recent approaches of  sequence-to-sequence models we need to trace back to the first architectures of Recurrent Neural Networks (RNN) \citep{jordan1986serial,elman1990finding}. RNNs are sequence models that deal directly with two modeling constraints of the standard multilayer perceptrons that are essential to model sequences. First, it is assumed that the input sequences have the same length. Second, it's an architecture that does not allow for the same token in different positions of the text to have similar features. The first ideas of RNNs arise in the seminal works of \cite{jordan1986serial} and \cite{elman1990finding}. We may summarize the vanilla RNN by the set of equations:
\begin{equation}
    s^{i} = g(W_{ss} s^{i-1} + W_{sx} x^i + b_s) 
\end{equation}
\noindent and
\begin{equation}
    y^i=g(W_{ys} s^i + b_y),
\end{equation}
where $x^i$ is a token in a piece of text, $y^i$ is the token we want to predict, $s^i$ is the state of the RNN, $W_{ss}$, $W_{ax}$, $W_{ys}$, $b_s$ and $b_y$ are the parameters of the network that we need to learn and $s^0$ is a vector of zeros. Figure \ref{fig:rnn} represents this model. Like the other models of language, we try to predict the probability of the next word. Therefore, we may estimate the parameters of this model in a semi-supervised fashion using the {\it backpropagation through time} algorithm. Suppose, for instance, that your text includes the first sentence of the song Africa by the band Toto ``I hear the drums echoing tonight.". Thus, $x^1=\mathrm{``I"}$, $x^2=\mathrm{``hear"}$ $x^3=\mathrm{``the"}$, $x^4=\mathrm{``drums"}$, $x^5=\mathrm{``echoing"}$,  $y^1=\mathrm{``hear"}$ $y^2=\mathrm{``the"}$, $y^3=\mathrm{``drums"}$, $y^4=\mathrm{``echoing"}$ and $y^5=\mathrm{``tonight"}$. Thus, this algorithm tries to maximize the probability that the token ``hear" arises when the token ``I" is an input, to maximize the probability the token ``the" happens when "hear" is the input and $s_1$ is the state of the system generated by ``I" and so on. Unfortunately, vanilla RNNs are not good at dealing with long sequences. Problems that arise in this context are the so-called {\it exploding} and {\it vanishing} gradients \citep{bengio1993problem,pascanu2013difficulty,ribeiro2020beyond}. This happens naturally due to the algorithm of backpropagation, which is based on the chain rule of calculus, and the consequent multiplication of the same shared matrix of the parameters several times. 

In order to overcome the difficulties of exploding and vanishing gradients, two important models of RNNs were introduced, namely Long Short-Term Memory (LSTM) \citep{hochreiter_1997,gers2000learning} and Gated Recurrent Units (GRU) \citep{cho2014learning} empirically explored in \cite{chung2014empirical}. The basic idea behind these models is to replace the simple units of the vanilla RNN model with complex units which include gates that control the flow of information that passes from one unit to the other. 

\begin{figure}[b]
    \centering
         \caption{Sequence-to-sequence models.}

     \begin{tikzpicture}

\def\RNN at (#1,#2){\begin{scope}[xshift=#1 cm,yshift=#2 cm]
\draw [fill=none,draw=black] (0,0) rectangle ++(1,4); 
\filldraw[color=black, fill=none](.5,.5) circle (.45);
\filldraw[color=black, fill=none](.5,1.5) circle (.45);
\filldraw[color=black, fill=none](.5,2.5) circle (.45);
\filldraw[color=black, fill=none](.5,3.5) circle (.45);
\end{scope}}
\draw [fill=none,draw=black] (0,0) rectangle ++(1,1); 
\draw [fill=none,draw=black] (5,0) rectangle ++(1,1); 
\draw [fill=none,draw=black] (8,0) rectangle ++(1,1); 
\draw [fill=none,draw=black] (13,0) rectangle ++(1,1); 

\draw [-Stealth, thick](1.7,0.5) -- (2.7,0.5) node [right, xshift=1pt] {\Large ...}; 
\draw [-Stealth, thick](3.5,0.5) -- (4.5,0.5); 
\draw [-Stealth, thick](6.5,0.5) -- (7.5,0.5); 
\draw [-Stealth, thick](9.7,0.5) -- (10.7,0.5) node [right, xshift=1pt] {\Large ...}; 
\draw [-Stealth, thick](11.5,0.5) -- (12.5,0.5); 

\draw [-Stealth, thick](8.5,1.3) -- (8.5,2) node [above, xshift=0pt,yshift=-3pt] {$\hat{y}^{<1>}$}; 
\draw [-Stealth, thick](13.5,1.3) -- (13.5,2) node [above, xshift=0pt,yshift=-3pt] {$\hat{y}^{<T_y>}$}; 
\draw [-Stealth, thick](0.5,-1) -- (0.5,-0.3) node [below, xshift=0pt,yshift=-18pt] {$\hat{x}^{<1>}$}; 
\draw [-Stealth, thick](5.5,-1) -- (5.5,-0.3) node [below, xshift=0pt,yshift=-18pt] {$\hat{x}^{<T_x>}$}; 

\end{tikzpicture}
    \label{fig:s2s}
\end{figure}
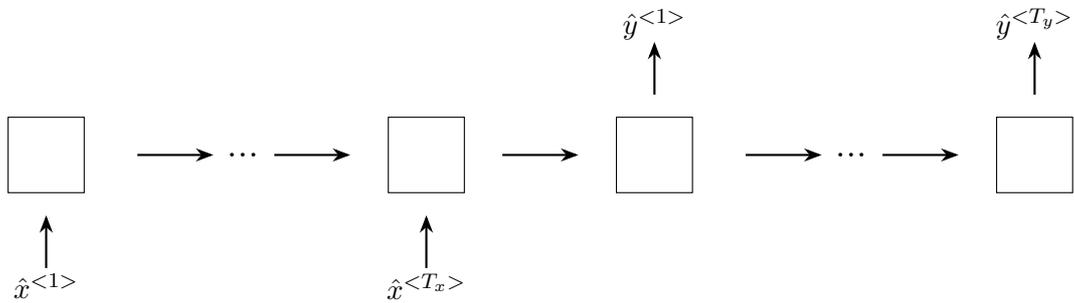

A natural extension of the vanilla RNNs is to consider the case where the input sequence and output sequence have different sizes. These models are called {\it Encoder-Decoder} models and they aim at mapping one sequence to another sequence \citep{sutskever2014sequence,vinyals2015neural}, where the encoder (decoder) is the part of the model that deals with the input (output) sequence. Figure  \ref{fig:s2s} presents an example of this topology, where each unit of this model is the RNN unit (vanilla, LSTM or GRU). In this type of model, the intention is, given a sequence, to predict another sequence not necessarily of the same size. For example, consider that we want to use a sequence-to-sequence model to translate the first sentence of the song ``Smoke on the Water" by the English band Deep Purple to Spanish. We may have  $x^1=$``We", $x^2=$``all", $x^3=$``came", $x^4=$``out", $x^5=$``to",   $x^6=$``Montreux", $y^1=$``Todos", $y^2=$``salimos'", $y^3=$``a" and $y^4=$``Montreux" in Figure  \ref{fig:s2s}, where $T_x=6$ and $T_y=4$.

A fundamental contribution to improving the learning process of sequence-to-sequence models is the attention mechanism \citep{bahdanau2014neural}. The main idea behind this paper is to include a set of weights to inform the model about the context. For instance, consider again the sentence ``We all came out to Montreux". We know that the phrasal verb ``come out" has different meanings. However, in this sentence, it is obvious that the sense of this verb is ``to go somewhere" and this happens because of the word ``Montreux", which is a town in Switzerland. In the work by \cite{bahdanau2014neural}, the authors evaluate the attention mechanisms by normalizing the output of a multilayer perceptron that depends on the state of the decoder model and the activation signal that comes from the encoder.       

The most recent models of NLP are based on transformers. Transformers are networks models that comprise the idea of processing complex information in parallel\footnote{This idea comes from the architecture of the Convolution Neural Network (CNN) that processes complex information in parallel \citep{lecun2015lenet}.} and an extension of the above-mentioned attention mechanism called multi-head attention \citep{vaswani_2017}. Multi-head attention is the idea of evaluating several attention models simultaneously. In a given sentence, we may use several attention mechanisms to explore several different dimensions at once. For instance, the chorus of the song ``America" by Neil Diamond says ``They're coming to America today". We may think of a multi-head mechanism as a way to help us ask and answer questions, i.e., it says where we should pay attention in a sentence to answer a given question\footnote{It has the same role as the filters built automatically by NN models to identify dimensions of images in CNN.}. Thus, the first question could be: ``What is happening?". Then, the answer is ``They are coming somewhere.". The second question could be ``Where?". Then, the answer is ``America". The third question could be ``When?". Then, the answer is ``Today". In the transformer network, this information is encoded in vectors called {\it Queries} ($Q$), {\it Keys} ($K$), and {\it Values} ($V$). Like in the recurrent sequence-to-sequence models, the architecture of Transformer models has an encoder-decoder structure. The encoder model receives the embeddings (discussed in Section \ref{sec:extractive_embedding}) of the inputs with a position encoding. The position encoding serves to inform the position of the token in the sequence, which is necessary here since this is a parallel model where the entire text is simultaneously inputted. The encoder block is formed by a stack of multi-head attention blocks connected with a feedforward network to generate the vectors $Q$, $K$ and $V$ that feed the following encoder blocks. The decoder block is a stack of an additional multi-head attention block and a block similar to the encoder block. The first block receives the output embeddings and generates the vector $Q$ to be used together with the vectors $K$ and $V$ it receives from the encoder block. This model is used to generate the output sequence in a recursive fashion.

There is now a large list of transformers that have been used in successful NLP tasks such as  
Google’s BERT \citep{devlin2018bert}, PEGASUS \citep{zhang2020pegasus}, T5 \citep{raffel2019exploring}, and Switch \citep{fedus2021switch}, Facebook’s BART \citep{lewis2019bart}, and Open AI’s Generative Pre-Training (GPT) \citep{radford2018improving},
GPT-2 \citep{radford2019language}, and GPT-3 \citep{brown2020language}.

An extension of these models is the so-called Longformer \citep{beltagy2020longformer}, which is a modified Transformer architecture with a self-attention operation that scales linearly with the sequence length, making it versatile for processing long documents. Finally, it is worth mentioning that we may find a survey of pre-trained language models for text generation in \cite{li2021pretrained} and a survey of dynamic neural network models for natural language processing in \cite{xu2022survey}.

Table \ref{tab:digest_transformers_abstractive} presents a summary of the most popular transformers used for abstractive summarization with their main characteristics. 

\begingroup
\renewcommand{\arraystretch}{1.2}
\begin{landscape}
\begin{table}[t]
    \centering
    \scriptsize
    \begin{tabular}{llp{7cm}p{4cm}p{3cm}}
     Name &Source              & Main characteristics & Dataset & Evaluation\\\hline
    BERT & \cite{liu2019text} & It is a bidirectional encoder model.   &CNN-DailyMail, XSUM and NYT & ROUGE-1, ROUGE-2 and ROUGE-L\\\cdashline{1-5} 
    BART &\cite{lewis2019bart} & It is a bidirectional encoder model with an autoregressive decoder architecture. &  CNN-DailyMail and XSUM & ROUGE-1, ROUGE-2 and ROUGE-L \\\cdashline{1-5}
    T5 &\cite{raffel2019exploring} & It is essentially the original transformer model equivalent to \cite{vaswani_2017} with  the normalization layer outside the residual path, and using a different position embedding scheme. & CNN/DailyMail & ROUGE-2\\\cdashline{1-5}
    GPT & \cite{radford2019language}& It is essentially the original transformer model equivalent to \cite{vaswani_2017}. & CNN/DailyMail & ROUGE-1, ROUGE-2 and ROUGE-L \\ \cdashline{1-5} 
    UniLM &  \cite{NEURIPS2019_c20bb2d9} &   It is a transformer model shared among three different self-attention masks. & CNN/DailyMail & ROUGE-1, ROUGE-2 and ROUGE-L\\ \cdashline{1-5} 
    MASS    & \cite{Song2019MASSMS} & It is a transformer model designed for encoder-decoder-based language generation tasks. & Gigaword &   ROUGE-1, ROUGE-2 and ROUGE-L           \\ \cdashline{1-5} 
    UniLMv2 & \cite{unilmv2}  &  It is a transformer model shared among three different self-attention masks. & CNN/DailyMail and XSUM &    ROUGE-1, ROUGE-2 and ROUGE-L                       \\ \cdashline{1-5} 
    PEGASUS&\cite{zhang2020pegasus} & It is essentially the original transformer model equivalent to \cite{vaswani_2017}. & XSUM, CNN/DailyMail, NEWSROOM, Multi-News, Gigaword, arXiv, PubMed, BIGPATENT, WikiHow, Reddit TIFU, AESLC and BillSum & ROUGE-1, ROUGE-2 and ROUGE-L  \\\cdashline{1-5}
    Ernie-Gen & \cite{xiao2020ernie} &  It is a transformer model with parameters shared among different tasks. &  Gigaword and CNN/DailyMail &  ROUGE-1, ROUGE-2 and ROUGE-L \\ \cdashline{1-5} 
    ProphetNet  & \cite{qi2020prophetnet} & It is a transformer model in the so-called future n-gram prediction as described in & Gigaword and CNN/DailyMail &  ROUGE-1, ROUGE-2 and ROUGE-L\\ \cdashline{1-5}      
    Longformer &\cite{beltagy2020longformer}  & It is a modified Transformer model with a self-attention operation that scales linearly with the sequence length. & arXiv &  ROUGE-1, ROUGE-2 and ROUGE-L
    \\\hline 
    \end{tabular}
    \caption{A significant compilation of transformers used for abstractive summarization.}
    \label{tab:digest_transformers_abstractive}
{\begin{scriptsize}\raggedright {\bf Notes:}
BERT due to \citep{devlin2018bert} and trained by \cite{liu2019text} is the acronym for Bidirectional Encoder Representations from Transformers. GPT due to \cite{radford2018improving} stands for Generative Pre-Trained model. BART due to \cite{lewis2019bart} is the acronym for Bidirectional and Auto-Regressive. Pegasus due \cite{vaswani_2017} and trained by \cite{zhang2020pegasus} is the acronym for Pre-training with extracted gap-sentences for abstractive summarization. T5 due to \cite{raffel2019exploring} stands for Text-to-Text Transfer Transformer.  UniLM due to \cite{NEURIPS2019_c20bb2d9} stands for Unified Language Model. MASS \citep{Song2019MASSMS} stands for MAsked Sequence-to-Sequence.
     \end{scriptsize}  \par } 
\end{table}
\end{landscape}
\endgroup

\section{Compressive extractive approaches}
\label{sec:compressive_extractive}

Compressive extractive approaches usually depend on two steps: (1) We extract the sentences that should belong to the summary; (2) We compress the chosen sentences that come from the original text in order to present only essential information. 

When selecting the sentences to be extracted, we usually rely on one of the methods already discussed in Section \ref{sec:extractive}. On the other hand, in order to compress the sentences we need a model of language.  

One advantage of the compressive extractive approaches over the pure extractive approaches is that for the case of the summaries that need to have a previously given constant length, the summaries created with the compressive approach usually contain more information than the summaries created with extractive approaches. This happens because by removing insignificant sentence components, we make room for more relevant information in the summary.

We have discussed many methods to extract sentences from the whole text in Section \ref{sec:extractive}. On the other hand, sentence compression is by itself a research problem. It starts with the input which is a sentence with $n$ words. The algorithm for sentence compression may drop any subset of these words and the words that remain with an unchanged order form a compression. Thus, the problem is not trivial since there are $2^n$ possible pieces of text to be chosen \citep{clarke2006models}. In order to solve this problem, we have to develop a method to determine what is the relevant piece of information in a sentence and how to present this information grammatically. 
 
Although our focus here is on text summarization, it is worth mentioning that there are some interesting specific applications of solutions of the sentence compression \citep{knight2000statistics}, such as the generation of TV captions that due to time and space constraints often requires only the most important parts of sentences \citep{zdenek2011sounds} and audio scanning services for the blind \citep{grefenstette1998producing}.
 
The precursors of compressive extractive summarization were the early attempts to provide text summaries in the style of newspapers headlines as in \cite{witbrock1999ultra} and \cite{banko2000headline}, where the summaries consist of a single sentence or even less than a sentence extracted from the text.  

In order to implement the compressive step of compressive extractive summarization, we may use unsupervised, supervised methods or hybrid approaches. 
 
The unsupervised approaches delete words based on part-of-speech tags\footnote{In linguistics, Part-Of-Speech (POS) tags, are tags used in a text (corpus) to associate a given word with its corresponding part of speech, based on both its definition and its context. The list of universal POS tags is: ADJ (adjective), ADP (adposition), ADV (adverb), AUX (auxiliary), CCONJ (coordinating conjunction), DET (determiner), INTJ (interjection), NOUN (noun), NUM (numeral), PART (particle), PRON (pronoun), PROPN (proper noun), PUNCT (punctuation), SCONJ (subordinating conjunction), SYM (symbol), VERB (verb) and X (other).} or the lexical items\footnote{A lexical item  may be a single word, a part of a word, or a sequence of words that forms the basic elements of a language's vocabulary. Examples of lexical items are: words (dog, table), phrasal verbs (get up, get over, get in, get on, idioms (``better late than never'', ``pull yourself together'') and sayings (``An apple a day keeps the doctor away.", ``Actions speak louder than words.").} alone. In particular, we may find a very interesting approach for unsupervised compressive summarization in  \cite{hori2004speech}. In order to generate a summary, their approach focuses on extracting topic words, weighting correct-word concatenations linguistically, and extracting reliable components of speech recognition acoustically as well as linguistically. A set of words maximizing a summarization score, indicating the appropriateness of a summarized sentence, is selected from those using a Dynamic Programming (DP) technique. The summarization score consists of word significance measured by the frequency of each word in the sentence, word confidence measured by the logarithm of the probability of $n$-grams, and the linguistic likelihood of summarized sentences.

Another interesting approach is the graph-based approach due to \cite{filippova2010multi} where a directed word graph is constructed. In this digraph, nodes represent words and edges represent the adjacency between words in a sentence. Thus, the authors can compress sentences by finding the k-shortest paths in the digraph. 

The supervised approaches may depend on a number of resources such as an annotated corpus with the original sentences and their corresponding reduced forms written by humans for training and testing purposes, a lexicon\footnote{A lexicon is the vocabulary of a language or a subject. For instance, the lexicon of computer science must present keys such as ``algorithm", ``big data", ``class", ``design pattern" and so on.} and a syntactic parser to generate a parse tree\footnote{A syntactic parsing converts the sentence into a tree whose leaves hold POS tags, but the rest of the tree tells how exactly these words join together to make the complete sentence. For example, a linking verb and a verb may combine to be a Verb Phrase (VP) such as in ``I \underline{have been studying} English for years.", where the underlined piece of text is a verb phrase.}. 

\cite{jing2000sentence} focuses specifically on the problem of sentence reduction of extracted sentences for summarization. The author assumes that the input of his system is the collection of extracted sentences that we can build using one of the methods of Section \ref{sec:extractive}. On the other hand, his algorithm of sentence reduction has five steps: (1) Syntactic parsing: He parses the input sentence to produce the sentence parse tree; (2) Grammar checking: He determines which components of the sentence must not be deleted to keep the sentence grammatical. To do this, he traverses the parse tree generated in the first step in top-down order and marks, for each node in the parse tree, which of its children are grammatically obligatory; (3) Context information: The system decides which components in the sentence are most related to the main topic being discussed. To measure the importance of a phrase in the local context, the system relies on lexical links between words; (4) Corpus evidence: The program uses the annotated corpus consisting of sentences reduced by human professionals and their corresponding original sentences to compute how likely (measuring the probabilities) are humans to remove a certain phrase; (5) Final decision: The final reduction decisions are based on the results from all the earlier steps. To decide which phrases to remove, the system traverses the annotated sentence parse tree and removes a phrase when it is not grammatically obligatory, not the focus of the local context and has a reasonable probability of being removed by humans.

Another interesting approach to supervised compressive summarization explores the noisy channel framework \citep{knight2000statistics}. In this framework, the authors consider that every sentence was originally shorter and then someone added some additional noisy text to it. Thus, the task of compressive summarization is to find the original sentence. It is worth mentioning that it is not relevant here whether or not the ``original" string is real or hypothetical. In this approach, the authors split the problem of sentence compression into three sub-problems: (1) Source model: They assign to every string (sentence) $s$ a probability $P(s)$, which gives the chance that $s$ is generated as an ``original short string"; (2) Channel model: They assign to every pair of strings (sentences) $s$ and $t$ a probability $P(t|s)$, which gives the chance that the expansion of the short string $s$ results in the long string $t$; (3) Decoder: They search for the short string s that maximizes $P(s|t)= P(s)P (t|s)$. In order to implement this, they assume that the probabilities $P(s)$ and $P (t|s)$ are associated with the representation of these sentences using parse trees. 

In \cite{cheng2016neural} and \cite{zhang2018neural} different from the above-mentioned works deal with both the extractive and compressive steps using neural network models such as the ones presented in Section \ref{sec:abstractive_sequence_seq2seq}.

Although the most common compressive approaches focus on editing sentences using compressive operations, other operations are also possible. \cite{jing2000cut}, based on analysis of human written abstracts, call attention to different types of operations such as {\it sentence combination} (merging material from several sentences), {\it syntactic transformation} (for instance, to change the position of the subject or to transform a piece of text from passive voice to active voice), {\it lexical paraphrasing} (replacing phrases with their paraphrases), {\it generalization or specification} (replacing phrases or clauses with more general or specific descriptions) and {\it reordering} (changing the order of specific sentences). However, besides compressing, they only implement the fusion of sentences when two sentences are close to each other and share the same subject or when a person or an entity is mentioned for the first time in a summary and there is a description of this person or entity in the text. Another approach is due to \cite{ganesan2010opinosis} that, using a graph-based approach where each word is a node in the graph, executes both compressive and fusion tasks of highly redundant sentences.

Table \ref{tab:digest_compressive_extractive} presents a significant compilation of ATS systems based on compressive extractive summarization.

\begingroup
\renewcommand{\arraystretch}{1.2}
\begin{landscape}
\begin{table}[t]
    \centering
    \scriptsize
    \begin{tabular}{llp{7cm}p{4cm}p{3cm}}
     Source          & Type    & Main contribution & Dataset & Evaluation\\\hline
    \cite{jing2000cut} & Unsupervised & It calls attention to other types of editions besides the compressive ones and it implements a limited set of sentence fusions. & 305 sentences from 50 summaries & Human experts \\\cdashline{1-5}
    \cite{jing2000sentence}  & Supervised  & It provides an algorithm based on 5 steps: (1) Synctatic parsing; (2)  Grammar checking; (3) Context information; (4) Corpus evidence; (5) Final decision.  & Free daily news service ``Communications-related headlines", provided by the Benton Foundation\footnote{http://www.benton.org}. & Percentage of system decisions that agree with human decisions\\\cdashline{1-5}
    \cite{knight2000statistics}  & Supervised  & It assumes that every sentence was originally shorter  and then someone added some additional noisy text to it. It splits the problem of sentence compression into three sub-problems: (1) Source model:  it assigns to every sentence probability that it was generated as an ``original shorter string"; (2) Channel model: it assigns to every pair of sentences the probability that one is the expansion of the other; (3) Decoder: They search for the short strings that maximize the product of the two previous probabilities.   & The Ziff-Davis corpus, which is a collection of newspaper articles announcing computer products. & Human experts\\\cdashline{1-5}    
    \cite{hori2004speech}  & Unsupervised  & It uses a dynamic programming approach to maximize a  summarization score that depends on the word significance, word confidence and the linguistic likelihood of summarized sentences.  & Japanese news broadcasts on TV in 1996. & Summarization accuracy \\\cdashline{1-5}
    \cite{filippova2010multi}  & Unsupervised  & It builds a directed graph where nodes represent words and edges represent the adjacency between words in a sentence and compresses the sentences finding the k-shortest paths in this digraph.   & News articles presented in clusters on Google News. & Human experts \\\cdashline{1-5}
    \cite{ganesan2010opinosis} & Unsupervised & It provides a graph based algorithm based on two kinds of operations, namely compression and fusion. & Reviews of hotels, cars and other products collected from Tripadvisor, Amazon and Edmunds. & ROUGE-1, ROUGE-2, ROUGE-SU-4\\\cdashline{1-5}
    \cite{cheng2016neural} & Supervised & It develops neural network models (LSTMs) to extract sentences and words in a supervised fashion as in Section \ref{sec:extractive_ml}. & DUC 2002 and CNN/Dailymail & ROUGE-1, ROUGE-2 and ROUGE-L \\\cdashline{1-5}
    \cite{zhang2018neural} & Supervised & It develops neural network models (LSTMs) to extract sentences and words in a supervised fashion as in Section \ref{sec:extractive_ml}. One particular contribution of this paper is to use the human-generated summaries in the training process.  & CNN/Dailymail & ROUGE-1, ROUGE-2 and ROUGE-L
    \\\hline 
    \end{tabular}
    \caption{A significant compilation of ATS systems based on compressive summarization.}
    \label{tab:digest_compressive_extractive}
\end{table}
\end{landscape}
\endgroup

\section{Evaluation methods}
\label{sec:evaluation}

The literature divides the methods for evaluating the quality of the generated summaries by the so-called {\it intrinsic} and {\it extrinsic} methods \citep{jing1998summarization,steinberger2009evaluation}. While intrinsic methods measure the quality of the summary, extrinsic methods measure a summary's performance when involved in a particular task such as {\it document categorization} and {\it question answering}. We may divide the intrinsic methods into three groups: {\it text quality evaluation}, {\it content-based evaluation} and {\it hybrid}. Text quality evaluation focuses on the readability of the summary.  On the other hand, content-based evaluation considers the performance of the method according to the chosen words. The hybrid methods consider both worlds. We may split the content group into three subgroups: {\it free-reference based}, {\it co-selection} and {\it content-based}. Free-reference based methods evaluate the content of the summaries without the need of a human-made reference. Co-selection evaluators pay attention to the sentences that were selected using metrics that come from the field of information retrieval. Finally, content-based evaluators focus on the selection of words. Considering the methods that need a human-made reference, content-based evaluators are much more popular than co-selection evaluators, since the former can be naturally applied to both extractive and abstractive summarization. The application of the latter makes more sense in extractive summarization.

Table \ref{tab:digest_evaluation} presents a digest of the methods used to evaluate ATS.  DUC 2005 readability arises in this table as a quality-based approach. This entry refers to the DUC 2005 that partially used this technique to evaluate the summaries presented by the competitors \citep{dang2005overview}. This method evaluates the quality of the summary using the following dimensions: grammatical (the summary should not have grammatical errors), non-redundancy (the summary should not have unnecessary repetition), referential clarity (the summary should be easy to identify who or what the pronouns and noun phrases in the summary are referring to), focus (the summary should contain only sentences with information that is related to the rest of the summary), structure and coherence (the summary should be well-structured and well-organized). Based on these dimensions, humans experts evaluate the summaries using a scale that ranges from {\it very Poor}, {\it poor}, {\it barely acceptable}, {\it good} to {\it very good}.  TAC 2008 also uses a quality-based approach and it considers exactly the same dimensions and scale of DUC 2005 readability \citep{dang2008overview}.

Another interesting attempt to provide a quality-based approach is due to \cite{grusky2018newsroom}. The authors select two semantic dimensions, namely informativeness and relevance, and two syntactic dimensions, namely fluency and coherence, for evaluation. They ask Amazon Mechanical Turk crowd workers to answer the following questions about these dimensions: (1) Informativeness: ``How well does the summary capture the key points of the article?" ; (2) Relevance ``Are the details provided by the summary consistent with details in the article? "; (3) Fluency: ``Are the individual sentences of the summary well-written and grammatical?"; (4) Coherence: ``Do phrases and sentences of the summary fit together and make sense collectively?". 

The idea considered in \cite{pitler2008revisiting} is to combine lexical, syntactic, and discourse features to produce a predictive model of human readers’ judgments of text readability. In this context, \cite{xenouleas2019sumqe} propose Sum-QE, a Quality Estimation model that adds a task-specific layer to a pre-trained BERT model, which is fine-tuned on the task of predicting scores for the five linguistic qualities assessed in DUC 2005. The model achieves a high correlation with human scores by addressing linguistic quality aspects that are only indirectly captured by recall-oriented evaluation metrics. 

There is a bunch of free-reference-based methods. The most simple methods are based on the classical distribution divergences. For instance, we may evaluate the divergence between $n$-grams of the candidate summary and the document using the Kullback–Leibler divergence given by:
\begin{equation}
    D_\text{KL}(P \parallel Q) = \sum_{x\in\mathcal{X}} P(x) \log_2\left(\frac{P(x)}{Q(x)}\right),
\end{equation}
\noindent where $P(x)$ is the probability of an event $x$ (the appearance of an $n$-gram) in the document and $Q(x)$ is the probability of an event in the candidate summary \citep{louis2013automatically}. Another option is to use the Jensen–Shannon divergence, as in FRESA \citep{torres2010summary} or \cite{louis2013automatically},  that symmetrizes the evaluation of the KL divergence:
\begin{equation}
    {\rm JSD}(P \parallel Q)= \frac{1}{2}D_{\text{KL}}(P \parallel M)+\frac{1}{2}D_\text{KL}(Q \parallel M).
\end{equation} 
SummTriver \citep{cabrera2018summtriver} considers the possibility of more than one candidate summary and it uses this piece of information to evaluate the trivergence among the distribution of an event in the candidate summary, the distribution of an event in a document formed by the set of candidate summaries and the distribution of an event in the document. It may evaluate the trivergence using different compositions of the divergences between two of these given probabilities.  We may also use the Hellinger distance \citep{pollard2002user,gonzalez2013class} to evaluate the distance between two distributions, given by:
\begin{equation}   
    H(P, Q) = \frac{1}{\sqrt{2}}  \sqrt{\sum_{x\in\mathcal{X}} (\sqrt{P(x)} - \sqrt{Q(x)})^2}, 
\end{equation}
where $0\le H(P, Q)\le 1$.

Furthermore, there are also other possibilities for evaluating the similarity between a candidate summary and the original text. For instance, we may evaluate the similarity by cosine of the representation of the text and candidate summaries using vector space models \citep{louis2013automatically} as discussed in Section \ref{sec:extractive_vector_space}  or word embeddings \citep{sun2019feasibility} as discussed in Section \ref{sec:extractive_embedding}. \cite{louis2013automatically} also suggests splitting the text and candidate summary into topics and to evaluate the quality of the summary by the fraction of the summary composed of text topic signatures or the percentage of topic signatures from the text that also arise in the summary. The summary likelihood approach considers the candidate summary as being generated according to word distributions in the document.  \cite{louis2013automatically} suggests two different ways to evaluate the likelihood of the summaries, namely the unigram probability model and the multinomial probability model. Another approach to free-reference-based evaluation is SUPERT (SUmmarization evaluation with Pseudo references and bERT) presented in \cite{gao2020supert}. In this work, the authors measure the relevance of a candidate summary by comparing it with a pseudo-reference summary.  Finally, \cite{bao2020end} uses the idea of transfer learning \citep{zhuang2020comprehensive} to train a neural network model using data from a dataset that has human-made summaries to provide a score to a summary of a document that does not have human-made summaries.

Two co-selection methods arise in Table \ref{tab:digest_evaluation}. The information retrieval-based approach, which considers the measures of precision, recall and F-score \citep{Baeza-Yates:2011:MIR:1796408} and the relative utility approach \citep{radev2004centroid}. In order to apply the information retrieval approach to evaluate the generated extractive summaries, we suppose that the objective of the method is to recover the sentences that are part of the human-generated summaries. Thus, we call the sentences that appear in the human-generated summaries {\it True} and the ones that do not, {\it False}. With that in mind, we are able to evaluate the {\it precision}, {\it recall} and {\it F-metric}. The precision is the ratio between the number of \textit{True} sentences recovered by the ATS systems and the number total of sentences recovered by it as in Eq. (\ref{eq:precision}):
\begin{equation}
\text{precision}=\frac{|\{\text{sentences in humans' summary}\}\cap\{\text{ATS systems' retrieved sentences}\}|}{|\{\text{ATS systems' retrieved sentences}\}|}.
\label{eq:precision}
\end{equation}

The recall is the ratio between the number of \textit{True} sentences recovered by the automatic summarizer and the total number of sentences that should be recovered according to the human-generated summary as in Eq. (\ref{eq:recall}):
 \begin{equation}
    \text{recall}=\frac{|\{\text{sentences in humans' summary}\}\cap\{\text{ATS systems' retrieved sentences}\}|}{|\{\text{sentences in humans' summary}\}|}.
    \label{eq:recall}
 \end{equation}
 
 The $F$-measure is the harmonic mean of precision and recall \citep{Baeza-Yates:2011:MIR:1796408} as in:\footnote{An extension of the $F$-measure is the \(F_\beta = (1+\beta^2) \frac{\mathrm{precision}\times  \mathrm{recall} }{ \beta^2 \mathrm{precision} + \mathrm{recall}}\), where $\beta$ measures the relative importance between {\it precision} and {\it recall}. If {\it precision} is as important as {\it recall}, we choose $\beta=1$ and the recover the harmonic mean.}
 \begin{equation}
    F = 2 \frac{\mathrm{precision}\times  \mathrm{recall} }{  \mathrm{precision} + \mathrm{recall}}.
    \label{eq:F-measure}
 \end{equation}
 
 The relative utility approach is an extension of the idea of using the information retrieval approach to evaluate summaries \citep{radev2003evaluation}. The starting point of this method is to assume that the human-generated summaries present the level of confidence that a given sentence should belong to the summary. With these weights in hand, we are able to evaluate the {\it relative utility}, which is the ratio between the weighted importance of the sentences recovered by the automatic summarizer and the weighted importance of the sentences that should be recovered by the automatic summarizer. In order to define this measure mathematically, we suppose that there are $N$ human experts, the document to be summarized has $n$ sentences and $e$ is the number of sentences in the extracted summary. Let $\delta_{sj}$ be the characteristic function that equals 1 when sentence $j$ belongs to the extracted summary of the ATS system and that equals 0 otherwise. Let also $\epsilon_{j}$ be the characteristic function that equals 1 when sentence $j$ belongs to the summary built based on the top $e$ sentences according to the average utilities of all judges and equals 0 otherwise. Thus, we may define the relative utility as  
\begin{equation}
    S=\frac{\sum_{j=1}^{n} \delta_{sj}  \sum_{i=1}^{N} u_{ij}}{\sum_{j=1}^{n} \epsilon_{j}  \sum_{i=1}^{N} u_{ij}}.
    \label{eq:relative_utility}     
\end{equation}
We may note this is a kind of weighted {\it recall}.

As we can see in Table \ref{tab:digest_evaluation}, there are several methods to evaluate ATS using the content-based approach. 

Factoid \citep{van2003examining}  and the Pyramid method \citep{nenkova2004evaluating} are the content-based manual approaches in Table \ref{tab:digest_evaluation}. In both cases, the human evaluators assess whether the information available in the human-made summaries is also in the automated summaries by comparing the content units of both texts. A factoid is a pseudo-semantic representation based on atomic information units that can be manually and robustly marked in the text. On the other hand, a pyramid presents the distinct units of text found in several reference summaries and the importance of these distinct units based on how many human-made summaries they occur. 
 
The cosine similarity is a natural approach to compare the tokens generated by the ATS and the tokens of the human-made summaries when we represent the summaries using the vector space model discussed in Section \ref{sec:extractive_vector_space}. In this approach, we evaluate the  cosine between the vectors that represent each summary. 

The unit overlap approach basically evaluates the Jaccard distance \citep{levandowsky1971distance} between the words that occur in the human-made and computer-made summaries \citep{saggion2002developing} as in
\begin{equation}
    \text{unit overlap}=\frac{|A\cap B|}{|A|+|B|-|A\cap B|},    
    \label{eq:unit_overlap}    
\end{equation}
where $A=\{\text{sentences in humans' summary}\}$ and $B=\{\text{ATS systems' retrieved sentences}\}$.

BLEU (BiLingual Evaluation Understudy) is a metric that was originally created for automatically evaluating machine-translated text.  It measures the overlap of $n$-grams between an automatically generated summary and a reference summary \citep{papineni2002bleu} in a {\it precision fashion} as in
\begin{equation}
    \text{BLEU}=\frac{|\{n\text{-grams} \in A\}\cap\{n\text{-grams} \in B\}|}{| n\text{-grams} \in B |},
    \label{eq:bleu}
\end{equation}
where $A=\{\text{humans' summary}\}$ and $B=\{\text{ATS systems' summary}\}$.

The next measure is the size of the Long Common Subsequence (LCS)\footnote{Let $X=\{x_1,x_2,\cdots,x_m\}$ and $Y=\{y_1,y_2,\cdots,y_n\}$. We say the $Y$ is a subsequence of $X$ if there exists a strictly increasing sequence of indexes $\{i_1,i_2,\cdots,i_K\}$ such that for all $j \in \{1,2,\cdots,K\}$ we have $x_{i_j}=y_j$. Given two sequences $X$ and $Y$, the Longest Common Subsequence (LCS) of $X$ and $Y$ is a common subsequence with maximum length.} between the strings formed by the human-made and computer-made summaries \citep{crochemore1994text}. LCS is evaluated using dynamic programming \citep{cormen2022introduction}.  

ROUGE (Recall Oriented Understudy for Gisting Evaluation) is a general approach to evaluating ATS systems. Since it comprises many different approaches, it is one of the most popular. ROUGE-$n$\footnote{\cite{lin2011class} calls attention that one interesting property of the ROUGE-$n$ is submodularity suggesting that if this is an important way to evaluate summaries, then optimal submodular approaches such as in their paper, \cite{lin2009graph} and \cite{kaageback2014extractive} are well justified.   } measures the overlap of $n$-grams between an automatically generated summary and reference summary as in
\begin{equation}
    \text{ROUGE-}n=\frac{|\{n\text{-grams} \in A\}\cap\{n\text{-grams} \in B\}|}{| n\text{-grams} \in A |},
    \label{eq:rouge-n}
\end{equation}
where $A=\{\text{humans' summary}\}$ and $B=\{\text{ATS system' summary}\}$. Although \cite{lin2003automatic} originally created it in a  {\it recall} fashion as in Eq. (\ref{eq:rouge-n}), the available softwares usually also present the precision fashion of it (which is the BLEU in Eq. (\ref{eq:bleu})) and the $F$-measure, which is the harmonic mean between the precision and recall as shown in Eq. (\ref{eq:F-measure}). 

It is worth mentioning that there are other commonly used variations of ROUGE \citep{lin2004rouge}. ROUGE-$L$ intends to consider the length of the LCS such as in \cite{crochemore1994text}. In this case, the LCS is measured in precision and recall fashions
\begin{equation}
    \text{precision}_L=\frac{LCS(\text{humans' summary},\text{ATS system' summary})}{\text{length}(\text{ATS system' summary})},
\end{equation}

\begin{equation}
    \text{recall}_L=\frac{LCS(\text{humans' summary},\text{ATS system' summary})}{\text{length}(\text{humans' summary})},
\end{equation}
and the $F$-measure is evaluated as in Eq. (\ref{eq:F-measure}).

Note that ROUGE-L, as the LCS, does not differentiate whether the match between the sequence and its subsequence is consecutive or not. For instance, let $X=\text{[`a',`b',`c',`d',`e']}$, $Y_1=\text{[`a',`b',`c',`f',`g']}$ and $Y_2=\text{[`a',`f',`b',`g',`c']}$. Both $Y_1$ and $Y_2$ have the same ROUGE-L score. Thus, ROUGE-W measures the WLCS (Weighted Longest Common Subsequence) that weights the LCS by the length of consecutive matches. 

A skip-bigram is any pair of words in their sentence order, allowing for arbitrary gaps. ROUGE-$S$ measures skip-bigram co-occurrence statistics between the human expert's summary and the ATS system's summary. A potential problem for ROUGE-$S$ is that it scores null if the sentence does not have any word pair co-occurring with its reference. For instance, ROUGE-S gives a score of 0 for the sequences $X=\text{[`a',`b',`c',`d']}$ and $Y=\text{[`d',`c',`b',`a']}$. ROUGE-SU-$k$ counts both skip bigrams and unigrams with maximal skip distance $k$.

One critique that we can make of the approaches that count the number of common extracts between the computer-based summary and the human-based summary is that the use of words with the same semantic is not considered. Thus, the next line of Table \ref{tab:digest_evaluation} presents a collection of works that extend the ROUGE methodology to consider semantically related words. Some of them use dictionaries or corpora. METEOR \citep{banerjee2005meteor}, \cite{ganesan2018rouge} and \cite{shafieibavani2018graph} use WordNet. \cite{zhou2006paraeval} build an independent paraphrase collection. On the other hand, other works apply neural network representations, such as \cite{ng2015better} (word embeddings), \cite{zhang2019bertscore} (BERT), \cite{zhao2019moverscore} (BERT), \cite{hailu2020framework} (word embeddings) and \cite{lee2020reference} (BERT),  presented in Sections \ref{sec:extractive_embedding} and \ref{sec:abstractive_sequence_seq2seq}.

The next row considers methods based on relevance analysis that measure the quality of the candidate summary by using a relative decrease in retrieval performance when indexing summaries instead of full documents. They use a search engine to find the most related documents (from a given set) to the candidate summary and to its corresponding human-made summaries forming two lists of retrieved documents. \cite{radev2003evaluation} and \cite{cohan2016revisiting} evaluate respectively the candidate summary score using the Kendall or the Spearman correlation measures and the intersection between the truncated lists of retrieved documents.

The latent-based approach \citep{steinberger2009evaluation} uses singular value decomposition discussed in Section \ref{sec:extractive_matrix_factorization} to capture the main topics of the document.  The computer-based summaries are ranked according to the similarity of the main topics of their summaries and their reference documents.

The semi-automated pyramid methods \citep{harnly2005automation,passonneau2013automated,passonneau2018wise}  use respectively unigram overlap, cosine similarity of latent vector representations and a weighted set cover algorithm to score the summaries given the manual pyramids. On the other hand, the automated pyramid methods \citep{yang2016peak,peyrard2017supervised,gao2018pyreval,gao2019automated} face the herculean task of building the pyramids automatically. In order to do that, \cite{yang2016peak} extract relation tuples using the Stanford Open Information Extraction \citep{angeli2015leveraging} and \cite{peyrard2017supervised} use a phrase structure parse and dependency parse to convert each sentence using the Stanford CoreNLP \citep{manning2014stanford}. In order to find the semantic similarity, they respectively use WordNet and word embeddings. SSAS (Semantic Similarity for Abstractive Summarization) \citep{vadapalli2017ssas} applies the model of \cite{yang2016peak} to extract the SCUs and combine various semantic and lexical similarity measures to score the quality of the candidate summary.

The basic elements approach \citep{hovy2005evaluating,hovy2006automated}, as in the case of BLEU, ROUGE or pyramid methods, compares the content of the reference summaries and generated summaries using small units as defined by the authors as (1) the head of a major syntactic constituent (noun, verb, adjective or adverbial phrases), expressed as a single item; or (2) a relation between a head and a single dependent, expressed as a triple (head | modifier | relation). In order to evaluate the score, the basic elements are evaluated according to lexical identity (the words must match exactly), lemma identity (the root forms of the words must match according to WordNet) and distributional similarity (words are similar according to the cosine distance on mutual information-based distributional similarity scores \citep{lin2002concept}). 

The SEE (Summary Evaluation Environment) \citep{Lin2001SEE,lin2003automatic} is a hybrid approach in Table \ref{tab:digest_evaluation}, in the sense that comprises ingredients of both the quality and content approaches. In this approach, the human evaluators compare the human-made summary with the automatic summary by finding the common pieces of text and specifying if the found pieces of text express all, most, some, hardly any or none of the content of the current model unit. In order to measure quality, the human evaluators rate grammar, cohesion, and coherence at those five levels.

The TAC 2008 overall responsiveness \citep{dang2008overview} is also a manual hybrid approach in Table \ref{tab:digest_evaluation}. It evaluates the degree to which a summary is responding to the information necessary to describe the topic state and the linguistic quality. It uses the five-point scale: (1) very poor; (2) poor; (3) barely acceptable; (4) good; (5) very good.

The extrinsic evaluation methods explore the quality of the summary in the completion of a specific task. The difficulty with matching a computer-made summary against an ideal summary is that the ideal summary is hard to establish.  In Table \ref{tab:digest_evaluation}, some tasks are considered, namely {\it document categorization}, {\it information retrieval}, {\it question answering} and {\it masked token task} \citep{taylor1953cloze}.

\begingroup
\renewcommand{\arraystretch}{1.2}
\begin{landscape}
\begin{table}[t]
    \centering
    \scriptsize
    \begin{tabular}{llllp{12cm}}
Approach                    & Kind                        & Sub-kind                        & Method                          & Source                                                                                           \\ \hline
\multirow{25}{*}{Intrinsic} & \multirow{4}{*}{Quality}                     &                                 & DUC 2005 readability                       & \cite{dang2005overview}                                                                         \\ \cdashline{4-5}

 &                      &                                 & TAC 2008 readability                        & \cite{dang2008overview}                                                                         \\ \cdashline{4-5}

 &                      &                                 & Newsroom human evaluation                        & \cite{grusky2018newsroom}                                                                          \\ \cdashline{4-5}

 &                      &                                 & Sum-QE                        & \cite{xenouleas2019sumqe}                                                                          \\ \cline{2-5}
                            & \multirow{18}{*}{Content}   & \multirow{5}{*}{Free reference based}   & Distributions divergence    &  \cite{torres2010summary} and \cite{cabrera2018summtriver}                                                            \\ \cdashline{4-5}
                            &    &    & Cosine similarity    &  \cite{louis2013automatically,sun2019feasibility}                                                              \\ \cdashline{4-5}
                            &    &    & topic    &  \cite{louis2013automatically}                                                              \\ \cdashline{4-5} &    &    & likelihood    &  \cite{louis2013automatically}                                                              \\ \cdashline{4-5}                           
                            &    &    & Pseudo reference    &  \cite{gao2020supert}                                                              \\ \cdashline{4-5}
                            &    &    & Transfer learning & \cite{bao2020end}\\
                            
                            \cline{3-5}
                            &    & \multirow{2}{*}{Co-selection}   & Information retrieval based     & \cite{Baeza-Yates:2011:MIR:1796408}                                                             \\ \cdashline{4-5} 
                            &                             &                                 & Relative utility                & \cite{radev2004centroid}                                                                        \\ \cline{3-5} 
                            &                             & \multirow{11}{*}{Content-based} & Manual                          & \cite{van2003examining,nenkova2004evaluating}                                                   \\ \cdashline{4-5} 
                            &                             &                                 & Cosine similarity               & \cite{salton1989automatic}                                                                      \\ \cdashline{4-5} 
                            &                             &                                 & Unity overlap                   & \cite{saggion2002developing}                                                                    \\ \cdashline{4-5} 
                            &                             &                                 & BLEU ($n$-grams matching)       & \cite{papineni2002bleu}                                                                         \\ \cdashline{4-5} 
                            &                             &                                 & Longest Common Subsequence      & \cite{saggion2002meta}                                                                          \\ \cdashline{4-5} 
                            &                             &                                 & Extracts matching (ROUGE)       & \cite{lin2003automatic,lin2004rouge}                                                            \\ \cdashline{4-5} 
                            &                             &                                 & Extracts matching with semantic & \cite{banerjee2005meteor}, \cite{zhou2006paraeval}, \cite{ng2015better}, \cite{ganesan2018rouge}, \cite{shafieibavani2018graph}, \cite{zhao2019moverscore} and \cite{hailu2020framework}. \\ \cdashline{4-5}
                            &                             &                                 & Search-based matching & \cite{radev2003evaluation} and \cite{cohan2016revisiting} \\ \cdashline{4-5}
                            &                             &                                 & Latent-based                    & \cite{steinberger2009evaluation}                                                                \\ \cdashline{4-5}

                            &                             &                                 & Semi-automated pyramid          & \cite{harnly2005automation,passonneau2013automated,passonneau2018wise}                          \\ \cdashline{4-5} 
                            &                             &                                 & Automated pyramid               & \cite{yang2016peak,peyrard2017supervised,vadapalli2017ssas}                                                       \\ \cdashline{4-5}
                            &                             &                                 & Basic elements               & \cite{hovy2005evaluating} and \cite{hovy2006automated}                                                       \\ \cline{2-5}                             
                            & \multirow{2}{*}{Hybrid}                      &                                 & SEE                             & \cite{Lin2001SEE,lin2003automatic} \\\cdashline{4-5}     
                            &                       &                                 & Overall responsiveness                             & \citep{dang2008overview}                                  
                            \\ \hline
\multirow{3}{*}{Extrinsic}  & \multirow{3}{*}{Task-based} & \multirow{3}{*}{}               & Document categorization         & \cite{brandow1995automatic,mani1997multi}                                                       \\ \cdashline{4-5} 
                            &                             &                                 & Information retrieval           & \cite{tombros1998advantages,radev2003evaluation}                                                \\ \cdashline{4-5} 
                            &                             &                                 & Question answering              & \cite{morris1992effects}, \cite{chen2018semantic}, \cite{scialom2019answers} and \cite{eyal2019question} \\ \cdashline{4-5}     
                            &                             &                                 & masked token task &
                            \citep{vasilyev2020fill}
                            \\ \hline
\end{tabular}
    \caption{A digest of the methods used to evaluate ATS.}
    \label{tab:digest_evaluation}
\end{table}
\end{landscape}
\endgroup

It is worth mentioning that, as we have shown in Table \ref{tab:digest_evaluation}, due to the inherent complexity of evaluating ATS systems, there are different methods for this task. However, as reported in \cite{blagec2022global}, BLEU and ROUGE dominate the field of evaluation in spite of their low correlation with human evaluation \citep{liu2008correlation,ng2015better}. We may find the same conclusion if we consider carefully the last column of Tables \ref{tab:digest_vector_space_models}, \ref{tab:digest_latent}, \ref{tab:digest_graph_methods}, \ref{tab:digest_topic_model}, \ref{tab:digest_word_embeddings},  \ref{tab:digest_heuristics}, \ref{tab:digest_extractive_linguistic}, \ref{tab:digest_suppervised}, \ref{tab:digest_reinforcement}, \ref{tab:digest_abstractive_linguistic}, \ref{tab:digest_transformers_abstractive} and \ref{tab:digest_compressive_extractive}.

\section{Open libraries}
\label{sec:libraries}

There are now several libraries with implementations of the most popular methods of ATS. Table \ref{tab:libraries_classical_extractive} presents a compendium of the extractive approaches introduced in Section \ref{sec:extractive}. Table \ref{tab:libraries_transformers} presents a compendium of summarization methods based on transformer models discussed in Section \ref{sec:abstractive_sequence_seq2seq}. Finally, Table \ref{tab:libraries_evaluation} presents a digest of open libraries with evaluation methods.

\begingroup
\renewcommand{\arraystretch}{1.2}
\begin{landscape}
\begin{table}[t]
    \centering
    \scriptsize
        \begin{tabular}{lp{6cm}llc}
Name                              & Source code                                                               & Method    & Source                                     & Section                           \\ \hline
\multirow{8}{*}{Sumy}             & \multirow{8}{*}{\url{https://github.com/miso-belica/sumy/}}               & Random    &                                            &                                   \\ \cdashline{3-5} 
                                  &                                                                           & KL-Sum    & \cite{haghighi-vanderwende-2009-exploring} & \ref{sec:extractive_vector_space} \\ \cdashline{3-5} 
                                  &                                                                           & Luhn      & \cite{Luhn1958}                            & \ref{sec:extractive_vector_space} \\ \cdashline{3-5} 
                                  &                                                                           & Edmundson & \cite{Edmundson1969}                       & \ref{sec:extractive_vector_space} \\ \cdashline{3-5} 
                                  &                                                                           & LSA       & \cite{steinberger2004using}                & \ref{sec:extractive_matrix_factorization}       \\ \cdashline{3-5} 
                                  &                                                                           & Textrank  & \cite{mihalcea2004textrank}                & \ref{sec:extractive_graph}        \\ \cdashline{3-5} 
                                  &                                                                           & Lexrank   & \cite{Erkan2004}                           & \ref{sec:extractive_graph}        \\ \cdashline{3-5} 
                                  &                                                                           & Reduction &                                            & \ref{sec:extractive_graph}        \\ \hline
\end{tabular}
    \caption{A compendium of the extractive approaches presented in Section \ref{sec:extractive}. The methods come from the library Sumy \citep{belica2022sumy}.}
    \label{tab:libraries_classical_extractive}
\end{table}
\end{landscape}
\endgroup

\begingroup
\renewcommand{\arraystretch}{1.2}
\begin{landscape}
\begin{table}[]
\scriptsize
\centering
\begin{tabular}{p{3.5cm}p{4.7cm}lllc}

\hline
Name                                                                                                 & Source Code                                                                                                                                                                                                                                                                                             & Kind                         & Method                                                                                       & Source                                        & Section                                                 \\ \hline
\multirow{8}{*}{\begin{tabular}[c]{@{}l@{}}transformers\\      (SummarizationPipeline)\end{tabular}} & \multirow{8}{*}{\href{https://huggingface.co/transformers/v3.0.2/_modules/transformers/pipelines.html\#SummarizationPipeline}{\vtop{\hbox{\texttt{huggingface.co/}}\hbox{\texttt{transformers/v3.0.2/\_modules/}}\hbox{\texttt{transformers/pipelines.html\#}}\hbox{\texttt{SummarizationPipeline}}}}}  & \multirow{8}{*}{Abstractive} & \href{https://huggingface.co/facebook/bart-large-cnn}{bart-large-cnn}                        & \cite{lewis2019bart}                          & \ref{sec:abstractive_sequence_seq2seq}                  \\ \cdashline{4-6} 
  & & & \href{https://huggingface.co/google/pegasus-xsum}{pegasus-xsum}                              & \multirow{2}{*}{\cite{zhang2020pegasus}}      & \multirow{2}{*}{\ref{sec:abstractive_sequence_seq2seq}} \\
  & & & \href{https://huggingface.co/google/pegasus-large}{pegasus-large}                            & & \\ \cdashline{4-6} 
& & & \href{https://huggingface.co/t5-small}{t5-small}                                             & \multirow{5}{*}{\cite{raffel2019exploring}}   & \multirow{5}{*}{\ref{sec:abstractive_sequence_seq2seq}} \\
 & & & \href{https://huggingface.co/t5-base}{t5-base}                                           
 & & \\
 & & & \href{https://huggingface.co/t5-large}{t5-large}                                         
 & & \\
& & & \href{https://huggingface.co/t5-3b}{t5-3b}                                                  & & \\
 & & & \href{https://huggingface.co/t5-11b}{t5-11b}                                                 &  & \\ \hline
\multirow{11}{*}{transformerSum}                                                                     & \multirow{11}{*}{\href{https://github.com/HHousen/TransformerSum}{\vtop{\hbox{\texttt{https://github.com/HHousen/}}\hbox{\texttt{TransformerSum}}}}} & \multirow{8}{*}{Extractive}  & \href{https://huggingface.co/distilroberta-base}{distilroberta-base-ext-sum}                 & \multirow{2}{*}{\cite{sanh2019distil}}        & \multirow{2}{*}{\ref{sec:extractive_ml}}                             \\
                                                                                                     &                                                                                                                                                                                                                                                                                                         &                              & \href{https://huggingface.co/distilbert-base-uncased}{distilbert-base-uncased-ext-sum}       &                                               &                                                         \\ \cdashline{4-6} 
                                                                                                     &                                                                                                                                                                                                                                                                                                         &                              & \href{https://huggingface.co/roberta-large}{roberta-large-ext-sum}                           & \multirow{2}{*}{\cite{liu2019roberta}}        & \multirow{2}{*}{\ref{sec:extractive_ml}}                             \\
                                                                                                     &                                                                                                                                                                                                                                                                                                         &                              & \href{https://huggingface.co/roberta-base}{roberta-base-ext-sum}                             &                                               &                                                         \\ \cdashline{4-6} 
                                                                                                     &                                                                                                                                                                                                                                                                                                         &                              & \href{https://huggingface.co/bert-base-uncased}{bert-base-uncased-ext-sum}                   & \multirow{2}{*}{\cite{devlin2018bert}}        & \multirow{2}{*}{\ref{sec:extractive_ml}} \\
                                                                                                     &                                                                                                                                                                                                                                                                                                         &                              & \href{https://huggingface.co/bert-large-uncased}{bert-large-uncased-ext-sum}                 &                                               &                                                         \\ \cdashline{4-6} 
                                                                                                     &                                                                                                                                                                                                                                                                                                         &                              & \href{https://huggingface.co/allenai/longformer-base-4096}{longformer-base-4096-ext-sum}     & \multirow{2}{*}{\cite{beltagy2020longformer}} & \multirow{2}{*}{\ref{sec:extractive_ml}} \\
                                                                                                     &                                                                                                                                                                                                                                                                                                         &                              & \href{https://huggingface.co/google/mobilebert-uncased}{mobilebert-uncased-ext-sum}          &                                               &                                                         \\ \cline{3-6} 
                                                                                                     &                                                                                                                                                                                                                                                                                                         & \multirow{3}{*}{Abstractive} & \href{https://huggingface.co/allenai/led-base-16384}{led-base-16384}                         & \multirow{3}{*}{\cite{beltagy2020longformer}} & \multirow{3}{*}{\ref{sec:abstractive_sequence_seq2seq}} \\
                                                                                                     &                                                                                                                                                                                                                                                                                                         &                              & \href{https://huggingface.co/allenai/led-large-16384}{led-large-16384}                       &                                               &                                                         \\
                                                                                                     &                                                                                                                                                                                                                                                                                                         &                              & \href{https://huggingface.co/HHousen/distil-led-large-cnn-16384}{distil-led-large-cnn-16384} &                                               &                                                         \\ \hline
                                                 
\end{tabular}
    \caption{A compendium of the transformer approaches presented in Sections \ref{sec:extractive_ml} and \ref{sec:abstractive_sequence_seq2seq}. The methods come from the Transformers \citep{Wolf_Transformers_State-of-the-Art_Natural_2020} and TransformerSum libraries.\citep{housen2022transformersum}.}
\label{tab:libraries_transformers}
\end{table}
\end{landscape}

\begingroup
\renewcommand{\arraystretch}{1.2}
\begin{landscape}
\begin{table}[]
\centering
\scriptsize
\begin{tabular}{llll}
Library                    & Source Code                                                 & Method                 & Source                               \\ \hline
\multirow{6}{*}{NLTK}      & \multirow{6}{*}{\url{https://www.nltk.org/}}                & Cosine simlarity       & \cite{salton1989automatic}          \\ \cdashline{3-4} 
                           &                                                             & Precision              & \cite{Baeza-Yates:2011:MIR:1796408} \\ \cdashline{3-4} 
                           &                                                             & Recall                 & \cite{Baeza-Yates:2011:MIR:1796408} \\ \cdashline{3-4} 
                           &                                                             & F-Measure              & \cite{Baeza-Yates:2011:MIR:1796408} \\ \cdashline{3-4} 
                           &                                                             & Likelihood             & \cite{louis2013automatically}       \\ \cdashline{3-4} 
                           &                                                             & Unit overlap (Jaccard) & \cite{saggion2002developing}        \\ \hline
\multirow{2}{*}{Gensim}    & \multirow{2}{*}{\url{https://radimrehurek.com/gensim/}}     & Kullback-Leibler       & \cite{louis2013automatically}       \\ \cdashline{3-4} 
                           &                                                             & Unit overlap (Jaccard) & \cite{saggion2002developing}        \\ \hline
\multirow{3}{*}{py-ROUGE} & \multirow{3}{*}{\url{https://github.com/Diego999/py-rouge}} & ROUGE-N                & \cite{lin2004rouge}                 \\ \cdashline{3-4} 
                           &                                                             & ROUGE-L                & \cite{lin2004rouge}                 \\ \cdashline{3-4} 
                           &                                                             & ROUGE-W                & \cite{lin2004rouge}                 \\ \hline
ROUGE-metric              & \url{https://github.com/li-plus/rouge-metric}               & ROUGE-SU               & \cite{lin2004rouge}                 \\ \hline
\end{tabular}
\caption{A compendium of the evaluation methods used in summarization and discussed in Section \ref{sec:evaluation}. These methods come from the NLTK \citep{bird2009natural}, Gensim  \citep{rehurek_lrec}, Py-ROUGE, and ROUGE-metric libraries.}
    \label{tab:libraries_evaluation}
\end{table}
\end{landscape}
\endgroup

\section{Empirical exercises}
\label{sec:comparison_exercises}

In this section, we present an empirical exploration of some of the free ATS libraries presented in Section \ref{sec:libraries}. For this exercise, we use CNN Corpus \citep{lins2019cnn} presented in Section \ref{sec:datasets}. We have run all extractive approaches presented in Section \ref{sec:libraries}. In order to run  the Edmundson summarizer we must specify a list of bonus words, that is, words that are positively relevant, a list of stigma words, negatively relevant, and a list of null words, deemed irrelevant to the summary. While the null words were defined as the list of stopwords of the NLTK library for the english language, the lists of bonus and stigma words were obtained by calculating the document frequency of terms in a subset of 200 golden summaries and source texts from each corpus, after stopwords removal. The 10 most frequent words in the subset of golden summaries were used as bonus words for each dataset, and the 10 words with the highest ratio of term frequency in source texts to the frequency in golden summaries were defined as stigma words.

We chose to run BART, mT5 and PEGASUS as the transformer models for abstractive summarization in our experiments. Given our intention of selecting ready-to-use summarization methods, these are some of the most popular transformers on the summarization pipeline of the Huggingface package. Therefore, these models are readily available in a pre-trained state by this implementation. As for which fine-tuning for each model was chosen, we decided upon pegasus-xsum, bart-large-cnn and mT5-multilingual-XLSum, trained with the XSum, CNN/Daily Mail and XL-Sum datasets, respectively.

We also present the typical summaries generated by these methods for a document from the CNN Corpus in Tables \ref{tab:typical_Summaries_extractive} and \ref{tab:typical_Summaries_abstractive}. This dataset has both extractive and abstractive golden summaries for each text. The ones used in our examples, as well as the source text, are presented below:
\clearpage
\vspace*{0.5 cm}
\noindent \fbox{\begin{minipage}{40em}
\small
{\bf Source text:} 
 A fourth infant has been discovered to have been infected with a rare, sometimes fatal form of bacteria that can come from baby formula, but there is no evidence the cases are related, federal health authorities said Friday. “Based on test results to date, there is no need for a recall of infant formula and parents may continue to use powdered infant formula, following the manufacturer's directions on the printed label”, the Centers for Disease Control and Prevention and the Food and Drug Administration said in a joint update. The latest case of Cronobacter infection occurred in Florida, the update said. After cases occurred in Missouri and Illinois, authorities looked for other cases, and found an Oklahoma case and the one in Florida. The Florida and Missouri infants died this month of their infections. The Missouri case prompted retail giant Wal-Mart to pull all cans of the same size and lot number of baby formula from its shelves. But DNA fingerprinting of the bacteria from the Missouri and Illinois cases found the bacteria differed, suggesting they were not related, the agencies said. Bacteria from the other two cases were not available for testing, they said. In the Missouri case, Cronobacter bacteria were found in an opened bottle of nursery water and prepared infant formula, but it was not clear how they became contaminated, the update said. Tests on factory-sealed containers of powdered infant formula and nursery water with the same lot numbers turned up no Cronobacter bacteria, it said, adding, “There is currently no evidence to conclude that the infant formula or nursery water was contaminated during manufacturing or shipping.” Formula maker Mead Johnson Nutrition said the agencies' test results corroborated its own. “We're pleased with the FDA and CDC testing, which should reassure consumers, health care professionals and retailers everywhere about the safety and quality of our products”, Tim Brown, senior vice president and general manager for North America, said in a statement. “These tests also reinforce the rigor of our quality processes throughout our operations.” Cronobacter infection typically occurs during the first days or weeks of life. In a typical year, the CDC said, it learns of four to six such cases. This year, with increased awareness of the infection, it has tallied 12 cases. The bacteria can cause severe infection or inflammation of the membranes that cover the brain. Symptoms can start with fever, poor feeding, crying or listlessness. “Any young infant with these symptoms should be in the care of a physician”, the update says. The bacteria can be found in the environment and can multiply in powdered infant formula once it is mixed with water, said the CDC, which recommends breastfeeding whenever possible. Cronobacter is fatal in nearly 40\% of cases, according to the CDC. Infection survivors can be left with severe neurological problems. 
\end{minipage}}

\vspace*{0.5 cm}

\noindent \fbox{\begin{minipage}{40em}
\small
{\bf Extractive golden summary:} 
“Based on test results to date, there is no need for a recall of infant formula and parents may continue to use powdered infant formula, following the manufacturer's directions on the printed label”, the Centers for Disease Control and Prevention and the Food and Drug Administration said in a joint update. But DNA fingerprinting of the bacteria from the Missouri and Illinois cases found the bacteria differed, suggesting they were not related, the agencies said. The bacteria can be found in the environment and can multiply in powdered infant formula once it is mixed with water, said the CDC, which recommends breastfeeding whenever possible. 

\end{minipage}}
\vspace*{0.5 cm}

\noindent \fbox{\begin{minipage}{40em}
\small
{\bf Abstractive golden summary:} 
“There is no need for a recall of infant formula”, federal health officials say. DNA fingerprinting finds the Missouri and Illinois bacteria are different, suggesting they're not related. CDC recommends breastfeeding whenever possible. 

\end{minipage}}\\

Based on the results of Table \ref{tab:CNNCorpus_exercise}, there is no consensus on the best method. However, we can see that some of the methods stand out. Among the classical extractive methods, we note the one due to \cite{Luhn1958} and Text Rank \citep{mihalcea2004textrank}. Among the transformers, we may cite the BART \citep{lewis2019bart}. This result makes sense since the training data of BART comprises a set of news. Thus, in some sense, this exercise is an exercise of learning transference \citep{zhuang2020comprehensive}. 

\clearpage

\begin{table}[]
\scriptsize
\centering
\begin{tabular}{@{}lrrrrrrrrrrrr@{}}
\toprule
\multicolumn{1}{c}{\multirow{2}{*}{CNN Corpus}} & \multicolumn{3}{c}{ROUGE-1}                                           & \multicolumn{3}{c}{ROUGE-2}                                           & \multicolumn{3}{c}{ROUGE-3}                                           & \multicolumn{3}{c}{ROUGE-4}                                           \\ 
\cmidrule(lr){2-4} 
\cmidrule(lr){5-7} 
\cmidrule(lr){8-10} 
\cmidrule(lr){11-13} 
\multicolumn{1}{c}{}                                & \multicolumn{1}{c}{P} & \multicolumn{1}{c}{R} & \multicolumn{1}{c}{F} & \multicolumn{1}{c}{P} & \multicolumn{1}{c}{R} & \multicolumn{1}{c}{F} & \multicolumn{1}{c}{P} & \multicolumn{1}{c}{R} & \multicolumn{1}{c}{F} & \multicolumn{1}{c}{P} & \multicolumn{1}{c}{R} & \multicolumn{1}{c}{F} \\ \midrule
SumyKL                                              & 34.20                 & 32.59                 & 33.37                 & 16.95                 & 16.28                 & 16.61                 & 14.12                 & 14.09                 & 13.89                 & 13.08                 & 13.09                 & 12.88                 \\
SumyLexRank                                         & 38.46                 & 43.83                 & 40.97                 & 23.00                 & 26.21                 & 24.50                 & 21.35                 & 23.18                 & 21.97                 & 20.29                 & 22.04                 & 20.89                 \\
SumyLsa                                             & 32.76                 & 33.74                 & 33.24                 & 16.21                 & 16.90                 & 16.55                 & 14.05                 & 14.97                 & 14.32                 & 13.21                 & 14.09                 & 13.47                 \\
SumyLuhn                                            & 35.86                 & 52.83                 & 42.72                 & 24.14                 & 34.37                 & 28.36                 & 25.50                 & 28.95                 & 26.89                 & 24.61                 & 27.94                 & 25.95                 \\
SumyEdmundson & 35.99 & 41.39 & 38.50 & 21.36 & 24.64 & 22.89 & 20.33 & 22.44 & 21.07 & 19.43 & 21.48 & 20.15 \\
SumyRandom                                          & 33.45                 & 30.77                 & 32.05                 & 15.33                 & 14.52                 & 14.91                 & 12.82                 & 12.54                 & 12.42                 & 11.94                 & 11.72                 & 11.59                 \\
SumyReduction                                       & 33.47                 & 49.65                 & 39.98                 & 21.61                 & 30.81                 & 25.40                 & 22.36                 & 25.20                 & 23.50                 & 21.51                 & 24.23                 & 22.60                 \\
SumySumBasic                                        & 37.72                 & 25.50                 & 30.43                 & 14.72                 & 10.53                 & 12.28                 & 11.51                 & 8.78                  & 9.62                  & 10.30                 & 7.93                  & 8.65                  \\
SumyTextRank                                        & 33.31                 & 50.26                 & 40.07                 & 21.63                 & 31.34                 & 25.60                 & 22.66                 & 25.54                 & 23.82                 & 21.83                 & 24.58                 & 22.93                 \\
bart-large-cnn                                      & 33.29                 & 33.54                 & 33.41                 & 15.87                 & 15.88                 & 15.87                 & 12.82                 & 12.97                 & 12.74                 & 9.38                  & 9.42                  & 9.28                  \\
mT5-multilingual-XLSum                            & 24.43                 & 18.25                 & 20.90                 & 4.98                  & 3.65                  & 4.21                  & 2.90                  & 2.13                  & 2.42                  & 1.38                  & 1.00                  & 1.14                  \\
pegasus-xsum                                        & 22.24                 & 23.24                 & 22.73                 & 6.26                  & 6.62                  & 6.43                  & 4.38                  & 4.68                  & 4.46                  & 2.60                  & 2.76                  & 2.63                  \\ \bottomrule

\end{tabular}

\vspace*{1 cm}

\begin{tabular}{@{}lrrrrrrrrr@{}}
\toprule
\multicolumn{1}{c}{\multirow{2}{*}{CNN Corpus}} & \multicolumn{3}{c}{ROUGE-L}                                           & \multicolumn{3}{c}{ROUGE-SU$_4$}                                          & \multicolumn{3}{c}{ROUGE-W}                                           \\ 
\cmidrule(lr){2-4} 
\cmidrule(lr){5-7} 
\cmidrule(lr){8-10} 
\multicolumn{1}{c}{}                                & \multicolumn{1}{c}{P} & \multicolumn{1}{c}{R} & \multicolumn{1}{c}{F} & \multicolumn{1}{c}{P} & \multicolumn{1}{c}{R} & \multicolumn{1}{c}{F} & \multicolumn{1}{c}{P} & \multicolumn{1}{c}{R} & \multicolumn{1}{c}{F} \\ \midrule
SumyKL                                              & 25.15                 & 24.19                 & 24.66                 & 19.14                 & 18.33                 & 18.73                 & 21.73                 & 8.99                  & 12.47                 \\
SumyLexRank                                         & 29.65                 & 33.85                 & 31.61                 & 24.70                 & 28.24                 & 26.36                 & 27.54                 & 12.41                 & 16.83                 \\
SumyLsa                                             & 23.76                 & 24.62                 & 24.18                 & 18.20                 & 18.95                 & 18.57                 & 20.94                 & 9.23                  & 12.59                 \\
SumyLuhn                                            & 29.28                 & 42.62                 & 34.71                 & 25.37                 & 36.66                 & 29.99                 & 30.48                 & 14.53                 & 19.40                 \\
SumyEdmundson & 27.99 & 32.30 & 29.99 & 23.04 & 26.66 & 24.72 & 26.30 & 12.05 & 16.26 \\
SumyRandom                                          & 23.89                 & 22.11                 & 22.96                 & 17.75                 & 16.64                 & 17.18                 & 20.88                 & 8.25                  & 11.55                 \\
SumyReduction                                       & 26.93                 & 39.42                 & 32.00                 & 22.97                 & 33.31                 & 27.19                 & 27.85                 & 13.21                 & 17.66                 \\
SumySumBasic                                        & 25.77                 & 17.52                 & 20.86                 & 17.80                 & 12.37                 & 14.60                 & 22.55                 & 6.69                  & 10.00                 \\
SumyTextRank                                        & 26.86                 & 39.99                 & 32.13                 & 22.97                 & 33.84                 & 27.36                 & 28.11                 & 13.34                 & 17.83                 \\
bart-large-cnn                                      & 25.29                 & 25.53                 & 25.41                 & 15.89                 & 15.93                 & 15.91                 & 24.33                 & 11.77                 & 15.61                 \\
mT5-multilingual-XLSum                            & 16.32                 & 12.25                 & 14.00                 & 7.29                  & 5.31                  & 6.14                  & 16.50                 & 5.93                  & 8.58                  \\
pegasus-xsum                                        & 15.81                 & 16.65                 & 16.22                 & 7.65                  & 8.03                  & 7.84                  & 15.70                 & 7.94                  & 10.35                 \\ \bottomrule
\end{tabular}

\vspace*{1 cm}

\begin{tabular}{@{}llllllll@{}}
\toprule
CNN Corpus           & \multicolumn{1}{c}{P} & \multicolumn{1}{c}{R} & \multicolumn{1}{c}{F} & \multicolumn{1}{c}{H} & \multicolumn{1}{c}{J} & \multicolumn{1}{c}{K-L} & \multicolumn{1}{c}{C} \\ \midrule
SumyKL                   & 31.92                 & 28.99                 & 30.38                 & 0.24 & 0.84 & 0.21                    & 0.43                  \\
SumyLexRank              & 37.89                 & 41.34                 & 39.54                 & 0.23 & 0.81 & 0.19                    & 0.49                  \\
SumyLsa                  & 30.97                 & 33.67                 & 32.26                 & 0.23 & 0.85 & 0.20                    & 0.37                  \\
SumyLuhn                 & 37.03                 & 49.10                 & 42.22                 & 0.24 & 0.80 & 0.18                    & 0.54                  \\
SumyEdmundson & 35.04 & 38.99 & 36.91 & 0.23 & 0.82 & 0.19 & 0.46 \\
SumyRandom               & 30.89                 & 29.19                 & 30.02                 & 0.23 & 0.85 & 0.21                    & 0.39                  \\
SumyReduction            & 34.14                 & 45.55                 & 39.03                 & 0.24 & 0.81 & 0.18                    & 0.54                  \\
SumySumBasic             & 34.90                 & 25.03                 & 29.15                 & 0.22 & 0.86 & 0.20                    & 0.39                  \\
SumyTextRank             & 33.95                 & 46.11                 & 39.11                 & 0.24 & 0.81 & 0.18                    & 0.54                  \\
bart-large-cnn           & 33.80                 & 33.91                 & 33.85                 & 0.22 & 0.83 & 0.19                    & 0.35                  \\
mT5-multilingual-XLSum & 24.14                 & 18.73                 & 21.09                 & 0.23 & 0.90 & 0.22                    & 0.22                  \\
pegasus-xsum             & 25.75                 & 22.45                 & 23.99                 & 0.25 & 0.89 & 0.26                    & 0.25                  \\ \bottomrule

\end{tabular}

\caption{CNN Corpus results. The evaluators used in this table are ROUGE-$n$ (1 to 4), ROUGE-L, ROUGE-SU$_4$, ROUGE-W, Precision (P), Recall (R), the F-measure, Hellinger (H), Jaccard distance (J), KL divergence (KL) and Cosine similarity (C).}
    \label{tab:CNNCorpus_exercise}
\end{table}

\clearpage

\begin{landscape}
\begin{table}[t]
    \centering
    \scriptsize
        \begin{tabular}{p{2cm}p{22cm}}
Method	&	Summary	\\\hline
SumyRandom	&	A fourth infant has been discovered to have been infected with a rare, sometimes fatal form of bacteria that can come from baby formula, but there is no evidence the cases are related, federal health authorities said Friday. In the Missouri case, Cronobacter bacteria were found in an opened bottle of nursery water and prepared infant formula, but it was not clear how they became contaminated, the update said. “We're pleased with the FDA and CDC testing, which should reassure consumers, health care professionals and retailers everywhere about the safety and quality of our products”, Tim Brown, senior vice president and general manager for North America, said in a statement. Symptoms can start with fever, poor feeding, crying or listlessness. 	\\\hline
SumyLuhn	&	A fourth infant has been discovered to have been infected with a rare, sometimes fatal form of bacteria that can come from baby formula, but there is no evidence the cases are related, federal health authorities said Friday. “Based on test results to date, there is no need for a recall of infant formula and parents may continue to use powdered infant formula, following the manufacturer's directions on the printed label”, the Centers for Disease Control and Prevention and the Food and Drug Administration said in a joint update. In the Missouri case, Cronobacter bacteria were found in an opened bottle of nursery water and prepared infant formula, but it was not clear how they became contaminated, the update said. Tests on factory-sealed containers of powdered infant formula and nursery water with the same lot numbers turned up no Cronobacter bacteria, it said, adding, “There is currently no evidence to conclude that the infant formula or nursery water was contaminated during manufacturing or shipping.” 	\\\hline
SumyEdmundson & A fourth infant has been discovered to have been infected with a rare, sometimes fatal form of bacteria that can come from baby formula, but there is no evidence the cases are related, federal health authorities said Friday. "Based on test results to date, there is no need for a recall of infant formula and parents may continue to use powdered infant formula, following the manufacturer's directions on the printed label", the Centers for Disease Control and Prevention and the Food and Drug Administration said in a joint update. The latest case of Cronobacter infection occurred in Florida, the update said. In a typical year, the CDC said, it learns of four to six such cases. 
\\\hline
SumyLsa	&	A fourth infant has been discovered to have been infected with a rare, sometimes fatal form of bacteria that can come from baby formula, but there is no evidence the cases are related, federal health authorities said Friday. Bacteria from the other two cases were not available for testing, they said. Tests on factory-sealed containers of powdered infant formula and nursery water with the same lot numbers turned up no Cronobacter bacteria, it said, adding, “There is currently no evidence to conclude that the infant formula or nursery water was contaminated during manufacturing or shipping.” In a typical year, the CDC said, it learns of four to six such cases. 	\\\hline
SumyLexRank	&	“Based on test results to date, there is no need for a recall of infant formula and parents may continue to use powdered infant formula, following the manufacturer's directions on the printed label”, the Centers for Disease Control and Prevention and the Food and Drug Administration said in a joint update. After cases occurred in Missouri and Illinois, authorities looked for other cases, and found an Oklahoma case and the one in Florida. Tests on factory-sealed containers of powdered infant formula and nursery water with the same lot numbers turned up no Cronobacter bacteria, it said, adding,” There is currently no evidence to conclude that the infant formula or nursery water was contaminated during manufacturing or shipping.” The bacteria can be found in the environment and can multiply in powdered infant formula once it is mixed with water, said the CDC, which recommends breastfeeding whenever possible.	\\\hline
SumyTextRank	&	“Based on test results to date, there is no need for a recall of infant formula and parents may continue to use powdered infant formula, following the manufacturer's directions on the printed label”, the Centers for Disease Control and Prevention and the Food and Drug Administration said in a joint update. But DNA fingerprinting of the bacteria from the Missouri and Illinois cases found the bacteria differed, suggesting they were not related, the agencies said. Tests on factory-sealed containers of powdered infant formula and nursery water with the same lot numbers turned up no Cronobacter bacteria, it said, adding, “There is currently no evidence to conclude that the infant formula or nursery water was contaminated during manufacturing or shipping.” The bacteria can be found in the environment and can multiply in powdered infant formula once it is mixed with water, said the CDC, which recommends breastfeeding whenever possible.	\\\hline
SumySumBasic	&	The latest case of Cronobacter infection occurred in Florida, the update said. After cases occurred in Missouri and Illinois, authorities looked for other cases, and found an Oklahoma case and the one in Florida. Tests on factory-sealed containers of powdered infant formula and nursery water with the same lot numbers turned up no Cronobacter bacteria, it said, adding, “There is currently no evidence to conclude that the infant formula or nursery water was contaminated during manufacturing or shipping.” Infection survivors can be left with severe neurological problems. 	\\\hline
SumyKL	&   The latest case of Cronobacter infection occurred in Florida, the update said. After cases occurred in Missouri and Illinois, authorities looked for other cases, and found an Oklahoma case and the one in Florida. But DNA fingerprinting of the bacteria from the Missouri and Illinois cases found the bacteria differed, suggesting they were not related, the agencies said. The bacteria can cause severe infection or inflammation of the membranes that cover the brain. 	\\\hline
SumyReduction	&	“Based on test results to date, there is no need for a recall of infant formula and parents may continue to use powdered infant formula, following the manufacturer's directions on the printed label”, the Centers for Disease Control and Prevention and the Food and Drug Administration said in a joint update. But DNA fingerprinting of the bacteria from the Missouri and Illinois cases found the bacteria differed, suggesting they were not related, the agencies said. Tests on factory-sealed containers of powdered infant formula and nursery water with the same lot numbers turned up no Cronobacter bacteria, it said, adding, “There is currently no evidence to conclude that the infant formula or nursery water was contaminated during manufacturing or shipping.” The bacteria can be found in the environment and can multiply in powdered infant formula once it is mixed with water, said the CDC, which recommends breastfeeding whenever possible. 	\\\hline

        \end{tabular}

\caption{Examples of extractive summaries.}
\label{tab:typical_Summaries_extractive}
\end{table}
\end{landscape}

\begin{table}[t]
    \centering
        \begin{tabular}{p{3cm}p{12cm}}
Method	&	Summary	\\\hline

Transformers- facebook/bart-large-cnn	&	A fourth infant has been infected with a rare, sometimes fatal form of bacteria. There is no evidence the cases are related, federal health authorities say. The latest case of Cronobacter infection occurred in Florida. In a typical year, the CDC says, it learns of four to six such cases.	\\\hline
Transformers- google/pegasus-xsum	&	There is no need for a recall of infant formula, US health officials say, after three babies died from an infection linked to the product in Missouri, Florida and Illinois. Tests on factory-sealed containers of powdered infant formula and nursery water with the same lot numbers turned up no Cronobacter.	\\\hline 
Transformers- csebuetnlp/ mT5\_multilingual \_XLSum	&	The US government has said there is no need for a recall of powdered infant formula and nursery water, following tests that show no evidence they were contaminated during manufacturing or shipment of the product.	\\\hline
        \end{tabular}
\caption{Examples of abstractive summaries.}
\label{tab:typical_Summaries_abstractive}
\end{table}

\clearpage

\section{Final remarks}
\label{sec:final_remarks}

In this work, we have provided a literature review of ATS systems. As we have previously mentioned, this is not an easy task. Since work by \cite{Luhn1958}, thousands of papers were introduced about this subject. In order to present a comprehensive review, we have divided the contributions by the type of output summary, as described in Section \ref{sec:classification}, namely {\it extractive}, {\it abstractive} and {\it hybrid}, and the type of model used to generate the summary. Several models have been used to generate summaries including the {\it classical frequency-based models} and the state of art {\it deep neural network sequence-to-sequence models}. In general, we may see in this work that any model of language can be used to generate a summary.

Besides the presentation of the models, we also introduced the most popular datasets used in the field, the methods used to evaluate the quality of the summaries, the public libraries that can be used to generate summaries and some empirical exercises exploring how the models discussed in the paper can be applied using the public libraries reviewed in Section \ref{sec:libraries} to generate useful summaries.

Although we  may surely assert that the field of ATS systems is a mature field, many difficulties still remain and these field difficulties  will undoubtedly be explored in the future:

\begin{enumerate}
    \item Text quality: The quality of a summary may strongly depend on the approach used to implement the ATS system. In extractive approaches, they still fail to provide solutions that avoid a lack of coherence between the sentences. In particular, one of the biggest difficulties is the ``dangling" anaphoras, i.e., sentences in the extracted summaries referring to other sentences that are not in the summary or to the wrong previous sentence.  Although there are some solutions that try to minimize this problem such as extracting entire paragraphs \citep{wu2003ontology} or using heuristics to avoid extracting sentences that make references to other sentences \citep{rush1971automatic}, this is still an ongoing problem. In abstractive summarization systems, the quality of the summary is strongly related to the capacity of text generation.

    \item Completeness: Most choices about what should be included in the summary are based on assumptions made about the text. A large set of models have been made available.  Due to the complexity of human language, all models have their own limitations. Frequency-based models strongly depend on the frequency of the tokens. With these models, important pieces of information presented using ``rare" words may not be considered. One solution to deal with this is to use semantically based approaches such as the ones presented in Section \ref{sec:extractive_embedding} based on word embeddings. Linguistic models usually depend on specific structures, rules or ontologies that may not be correctly applied in many situations. Sequence-to-sequence deep learning models are also kinds of frequency-based models with the same merits and problems. Furthermore, they need to be trained with large annotated datasets. If the dataset is not available, we have to transfer the learning from one dataset to the other. This transference may work or may not work depending on the differences and similarities between the two pieces of text. 

    \item Size of the input text: In extractive summarization, although the size of input text usually is not directly a problem, the quality of the summary may deteriorate when we concatenate sentences that come from very different parts of the text. The size of the input text may not be a problem for the linguistic-based abstractive approaches either. However, if the size of the input text is positively correlated with the complexity of the text, this might be a problem, since the structures used for linguistic-based abstractive summarization may not be able to accommodate this richness of details. Due to the computational complexity, deep learning sequence-to-sequence models suffer directly from the size of input since most models are available to texts of moderate size.
    
    \item Size of the abstract: Considering abstracts that convey the same pieces of information, abstracts generated by extractive summarization approaches are usually larger than the ones generated by abstractive methods. The point is that in the case of the extractive approaches, the relevant pieces of information are distributed in several sentences, and in many situations, irrelevant pieces of information remain in the extracted sentences.   

    \item Datasets: Although there are many datasets available today as we presented in Section \ref{sec:datasets}, many of these datasets are related to a specific field (news or scientific papers) since in general, it is very costly to create datasets. This problem increases when we need datasets to train extractive systems, since human summaries are abstractive in nature, as we mentioned in Section \ref{sec:extractive_ml}.
    
    \item Evaluation: The biggest issue with the activity of evaluating the task of ATS is that there is not a perfect summary that can be used as a model such as in other machine learning tasks. Different human experts may generate great but different summaries. Besides the fact that different human experts may choose different pieces of text to include in the summary, they may include the same pieces of text in semantically equivalent ways, making the task of comparing machine summaries to human summaries very difficult.
    
\end{enumerate}

\section{Acknowledgment}

The authors acknowledge and thank the partial financial support from the Ministry of Science and Technology of Brazil (MCTI). DOC (grant number 302629/2019-0) and LW (309545/2021-8) also thank CNPQ for the partial financial support.




\end{document}